\pdfoutput=1
\documentclass{article}%
\usepackage{amsmath}
\usepackage{amsfonts}
\usepackage{cite}
\usepackage{float}
\usepackage{amssymb}
\usepackage{graphicx}
\usepackage{minitoc}%
\providecommand{\U}[1]{\protect\rule{.1in}{.1in}}

\makeatletter
\def\@crosshairs{\vbox to0pt{}}
\makeatother
\begin{document}

\title{\textbf{Deep Convolutional Spiking Neural Networks}\\\textbf{for Image Classification}\\}
\author{Ruthvik Vaila and John Chiasson\\Boise State University\\\emph{\{ruthvikvaila,johnchiasson\}@boisestate.edu}
\and Vishal Saxena\\University of Idaho\\\emph{vsaxena@uidaho.edu}}
\date{March 23, 2019}
\maketitle

\begin{abstract}
Spiking neural networks are biologically plausible counterparts of the
artificial neural networks, artificial neural networks are usually trained
with stochastic gradient descent and spiking neural networks are trained with
spike timing dependant plasticity. Training deep convolutional neural networks
is a memory and power intensive job. Spiking networks could potentially help
in reducing the power usage. In this work we focus on implementing a spiking CNN using
Tensorflow to examine behaviour of the network and empirically study the
effect of various parameters on learning capabilities and also study
catastrophic forgetting in the spiking CNN and weight initialization problem
in R-STDP using MNIST and N-MNIST data sets.\newpage

\end{abstract}

\section{\noindent Introduction}

Deep learning, i.e., the use of deep convolutional neural networks (DCNN ), is
a powerful tool for pattern recognition (image classification) and natural
language (speech) processing \cite{SCHMIDHUBER2015}\cite{Nielsen}. Deep
convolutional networks use multiple convolution layers to learn the input data
\cite{LeCun2015} \cite{Long} \cite{Girshick}. They have been used to classify
the large data set Imagenet \cite{Krizhevsky} with an accuracy of 96.6\%
\cite{Chollet}. In this work deep spiking networks are
considered\cite{Pfeiffer2018}. This is new paradigm for implementing
artificial neural networks using mechanisms that incorporate spike-timing
dependent plasticity which is a learning algorithm discovered by
neuroscientists \cite{Hassabis} \cite{Masquelier2017}. Advances in deep
learning has opened up multitude of new avenues that once were limited to
science fiction \cite{VanRullen}. The promise of spiking networks is that they
are less computationally intensive and much more energy efficient as the
spiking algorithms can be implemented on a neuromorphic chip such as Intel's
\textsc{LOIHI}\ chip \cite{lohi} (operates at low power because it runs
asynchronously using spikes \cite{wu} \cite{wu2018} \cite{wu2015} \cite{vishal2018} \cite{Dahl2018}). Our work is based on the work of Masquelier and
Thorpe \cite{Masquelier2007} \cite{Masquelier2008}, and Kheradpisheh et al.
\cite{Kheradpisheh_2016} \cite{Kheradpisheh_2016b}. In particular a study is
done of how such networks classify MNIST image data \cite{mnist} and N-MNIST
spiking data \cite{nmnist}. The networks used in \cite{Kheradpisheh_2016}
\cite{Kheradpisheh_2016b} consist of multiple convolution/pooling layers of
spiking neurons trained using spike timing dependent plasticity (STDP
\cite{Sjostrom:2010}) and a final classification layer done using a support
vector machine (SVM) \cite{svm}.

\subsection{Spike Timing Dependant Plasticity (STDP)}

\begin{figure}[H]%
\centering
\includegraphics[
height=1.1761in,
width=7.0076in
]%
{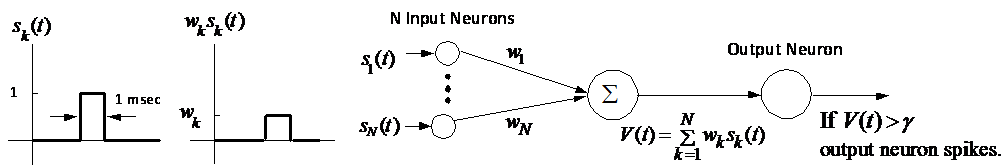}%
\caption{The neurons $s_{i},i=1,...,N$ are the pre-synaptic neurons and the
output neuron is the post-synaptic neuron.}%
\label{stdp_fig1b}%
\end{figure}

Spike timing dependant plasticity (STDP) \cite{Markram} has been shown to be
able to detect hidden (in noise) patterns in spiking data
\cite{Masquelier2008}. Figure \ref{stdp_fig1b} shows a simple 2 layer fully
connected network with $N$ input (pre-synaptic) neurons and 1 output neuron.
The spike signals $s_{i}(t)$ are modelled as being either 0 or 1 in one
millisecond increments. That is, 1 msec pulse of unit amplitude represents a
spike while a value of 0 represents no spike present. See the left side of the
Figure \ref{stdp_fig1b}. Each spike signal has a weight (synapse) associated
with it which multiplies the signal to obtain $w_{i}s_{i}(t)$ which is called
the \emph{post synaptic potential} due to the $i^{th}$ input neuron. These
potentials are then summed as%
\[
V(t)=\sum_{i=1}^{N}w_{k}s_{k}(t).
\]
$V(t)$ is called the \emph{membrane potential} of the output neuron. At any
time $t$ if the membrane potential $V(t)$ is greater than a specified
threshold $\gamma$, i.e., if
\[
V(t)>\gamma
\]
then the output neuron spikes. By this we mean that the output neuron produces
a 1 msec pulse of unit amplitude. See the right side of Figure
\ref{stdp_fig1b}.%

Denote the input spike pattern $s(t)$ as
\[
s(t)=\left[
\begin{array}
[c]{c}%
s_{1}(t)\\
s_{2}(t)\\
\vdots\\
s_{N}(t)
\end{array}
\right]
\]
Let $t_{1}<t_{2}<t_{3}<\cdots$ be a sequence of times for which the spike
pattern is fixed, that is, $s_{fixed}=s(t_{1})=s(t_{2})=s(t_{3})=\cdots$ while
at all other times the values $s_{i}(t)$\ are \emph{random} (E.g.,
$P(s_{i}(t)=1)=0.01$ and $P(s_{i}(t)=0)=0.99$). The idea here is that the
weights can be updated according to an unsupervised learning rule that results
in the output spiking if and only if the fixed pattern is present. The
learning rule used here is called spike timing dependent plasticity or STDP.
Specifically, we used a simplified STDP model as in given as
\cite{Kheradpisheh_2016}%
\[
w_{i}\leftarrow w_{i}+\Delta w_{i},\text{ \ }\Delta w_{i}=%
\begin{cases}
-a^{-}w_{i}(1-w_{i}),\ \ \text{if}\ \ t_{out}-t_{i}<0\\
+a^{+}w_{i}(1-w_{i}),\ \ \text{if}\ \ t_{out}-t_{i}\geq 0.
\end{cases}
\]
Here $t_{i}$ and $t_{out}$ are the spike times of the pre-synaptic (input) and
the post-synaptic (output) neuron, respectively. That is, if the $i^{th}$
input neuron spikes before the output neuron spikes then the weight $w_{i}$ is
increased otherwise the weight is decreased.\footnote{The input neuron is
assumed to have spiked \emph{after} the output neuron spiked.} Learning refers
to the change $\Delta w_{i}$ in the (synaptic) weights with $a^{+}$ and
$a^{-}$ denoting the learning rate constants. These rate constants are
initialized with low values $(0.004,0.003)$ and are typically increased as
learning progresses. This STDP rule is considered simplified because the
amount of weight change doesn't depend on the time duration between
pre-synaptic and post-synaptic spikes.

To summarize, if the pre-synaptic (input) neuron spikes before post-synaptic
(output) neuron, then the synapse is increased. If the pre-synaptic neuron
doesn't spike before the post-synaptic neuron then it is assumed that the
pre-synaptic neuron will spike later and the synapse is decreased.

\begin{figure}[H]%
\centering
\includegraphics[
height=2.9291in,
width=5.3048in
]%
{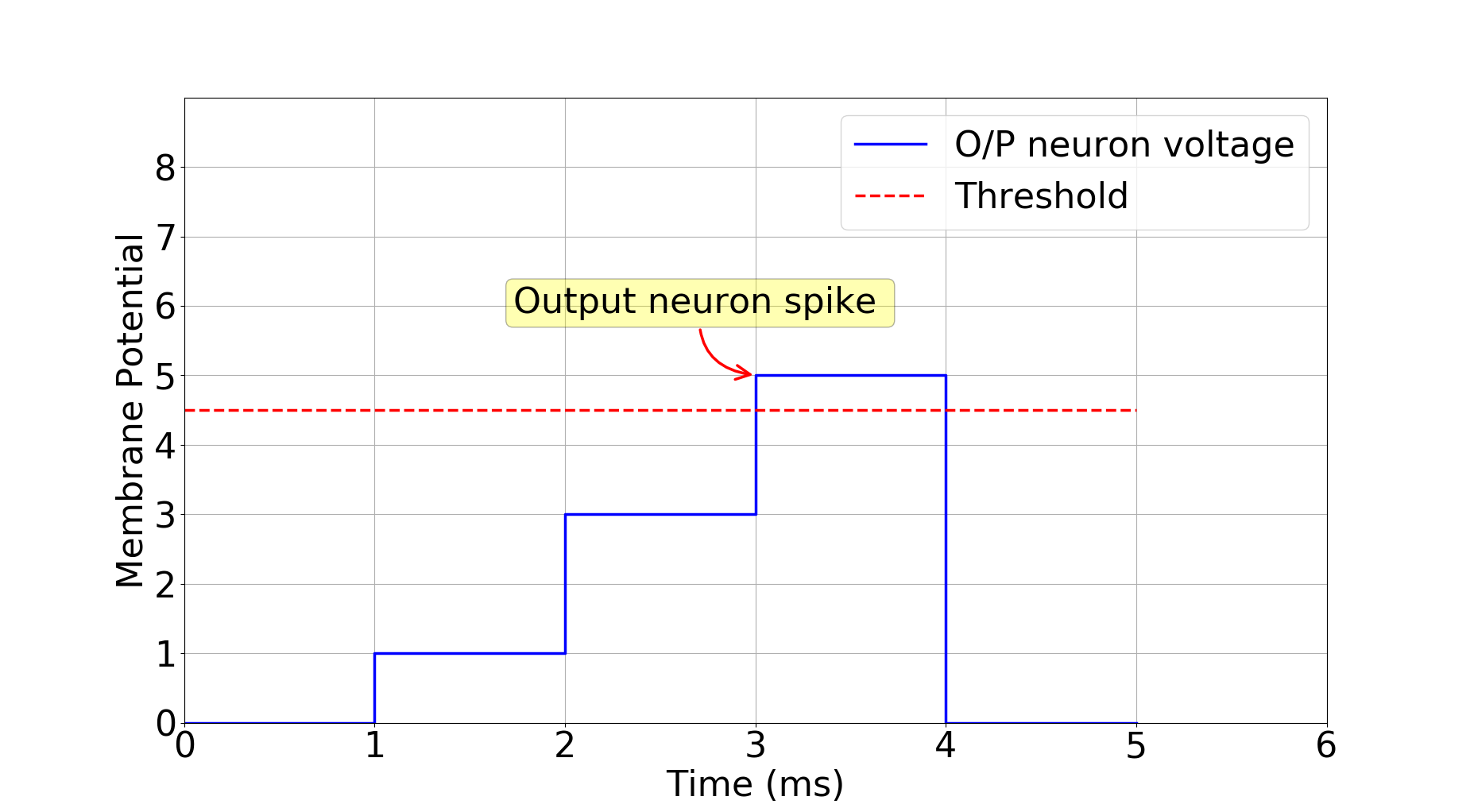}%
\caption{Spike generation by an output neuron.}%
\label{outputneuron}%
\end{figure}

The membrane potential profile of the type of output neuron considered here
looks as shown in the Figure \ref{outputneuron}. In Figure \ref{outputneuron}
the output neuron is shown to receive a spike at 1 msec, two spikes at 2 msec
and another two spikes at 3 msec. The output neuron spikes at time 3 msec as
its membrane potential exceeded the threshold ($\gamma=4.5$).%

\begin{figure}[H]%
\centering
\includegraphics[
height=2.9992in,
width=5.4293in
]%
{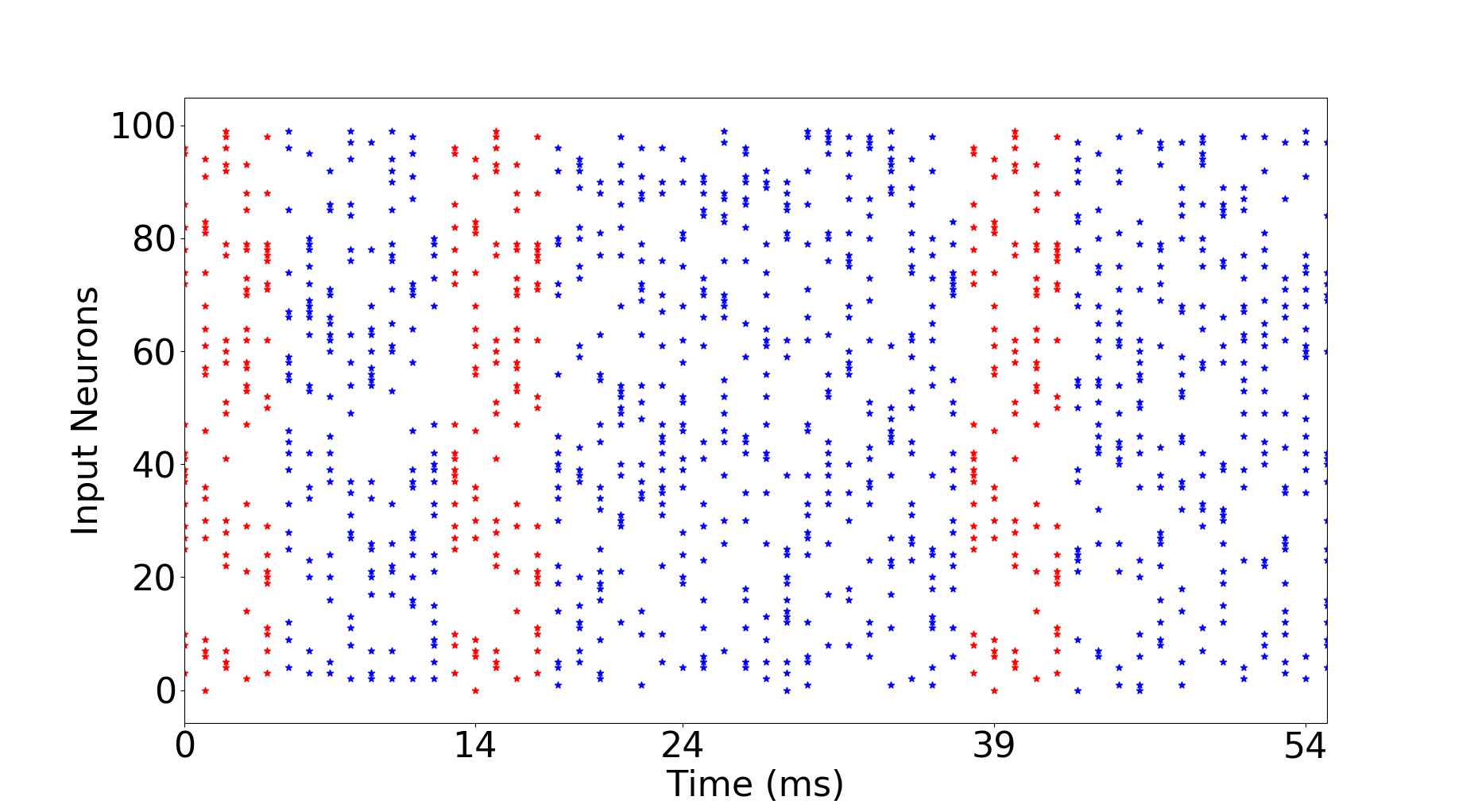}%
\caption{The pattern is red and has a duration of 5 miliseconds. This pattern
is presented recurrently to the network at random times. The random noisy
spikes are represented in blue.}%
\label{rasterplots}%
\end{figure}

Figure \ref{rasterplots} shows a raster plot of an input neuron versus its
spike times for the first 54 msecs. Figure \ref{rasterplots} shows $N=100$
input neurons and at time $t$ a dot $\ast$ denotes a spike while an empty
space denotes no spike. Red dots in the plot indicates a spike as part of a
fixed pattern of spikes. In Figure \ref{rasterplots} the pattern presented to
the output neuron is 5 msec long in duration. The blue part of Figure
\ref{rasterplots} denotes random spikes being produced by the input neurons
(noise). On close observation of Figure \ref{rasterplots} one can see that
fixed spike pattern in red is presented at time 0, time 13, and time 38.

\begin{figure}[H]%
\centering
\includegraphics[
height=3.2344in,
width=5.8565in
]%
{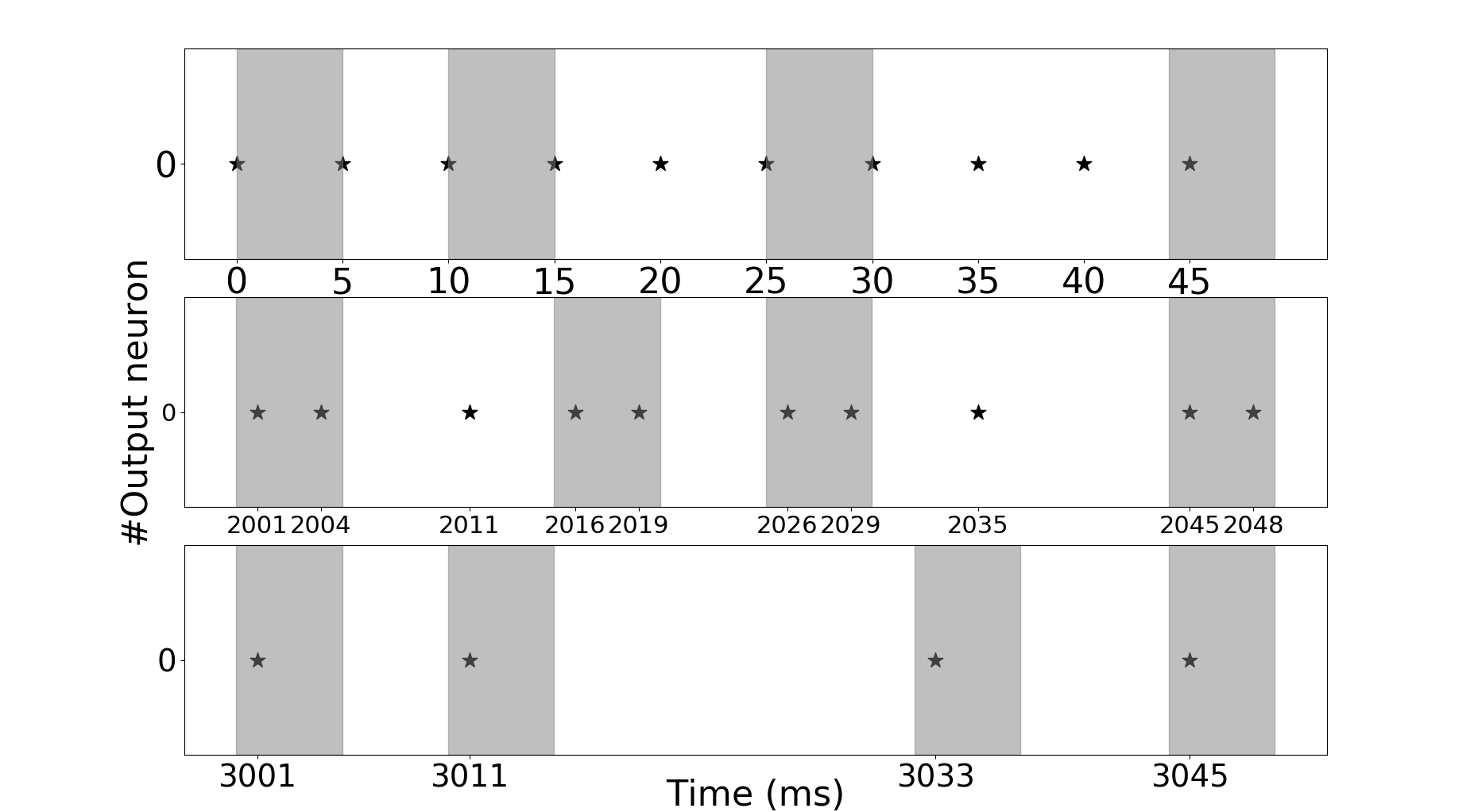}%
\caption{The grey box indicates the fixed pattern being produced by the input
neurons $s_{i}.$}%
\label{allpatterns}%
\end{figure}

Using only the above STDP learning rule, the output neuron learns to spike
only when the fixed pattern $s_{fixed}$ is produced by the input neurons. With
the weights $w_{i}$ set randomly from normal distribution, i.e., $w_{i}%
\sim\mathcal{N}(0.5,0.05)$ Figure \ref{allpatterns} (top plot) shows the output spiking randomly
for the first 50 msecs. However after about 2000 msec, Figure \ref{allpatterns} (middle plot) shows
the output neuron starts to spike selectively, though it incorrectly spikes at
times when the pattern is not present. Finally, after about 3000 msec, Figure
\ref{allpatterns} (bottom plot) shows that the output neuron spikes only when the pattern is
present.

%


%


\subsection{Convolution operation}

\begin{figure}[H]%
\centering
\includegraphics[
height=2.0332in,
width=3.7723in
]%
{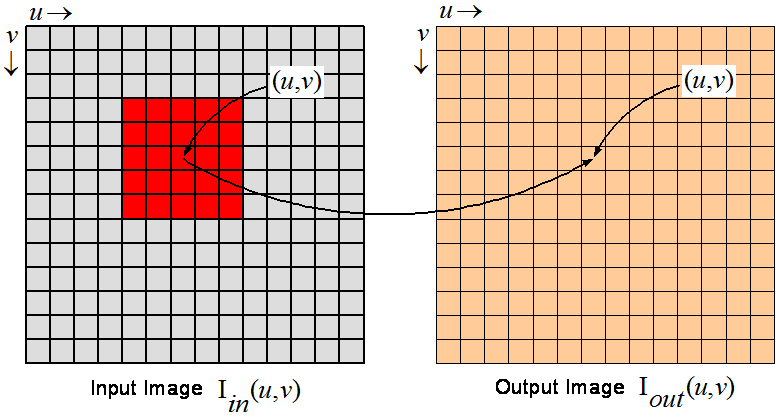}%
\caption{Convolution operation.}%
\label{stdp_fig10}%
\end{figure}

In this work spiking convolutional neural networks (SCNN) are used for feature
extraction. A short explanation of convolution is now presented. Figure
\ref{stdp_fig10} shows a convolution operation on an input image.

Let
\[
W_{C}(i,j),\text{ \ \ }0\leq i,j\leq4
\]
denote a $5\times5$ convolution weight kernel (filter) indicated by the red
square Figure \ref{stdp_fig10} above. With the kernel centered on the location
$(u,v)$ of the input image $\mathbf{I}_{in}(u,v)$ ($0\leq u,v\leq14 $) the
value $\mathbf{I}_{out}(u,v)$ ($0\leq u,v\leq14$) of the output image at
$(u,v)$ is given by
\[
\mathbf{I}_{out}(u,v)=\sum_{j=-2}^{j=2}\sum_{i=-2}^{i=2}\mathbf{I}%
_{in}(u+i,v+j)W_{C}(i,j).
\]
Note that the shape of the output image is same as the input image, such
convolutions are called same mode convolutions.

\begin{figure}[H]%
\centering
\includegraphics[
height=2.0081in,
width=3.3408in
]%
{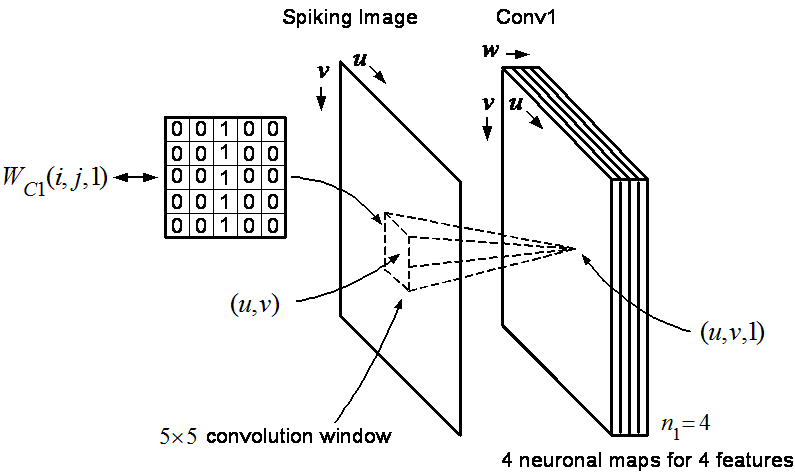}%
\caption{Feature detection.}%
\label{stdp_fig13b}%
\end{figure}

Convolution networks are used to detect features in images. To explain,
consider the convolution kernel $W_{C1}(i,j,1)$ as shown in Figure
\ref{stdp_fig13b}. This kernel is used to find vertical lines of spikes at any
location of the spiking input image. For example, at the location $(u,v)$ at
time $\tau$ the kernel is convolved with the spiking image to give
\[
\sum_{j=-2}^{2}\sum_{i=-2}^{2}s_{in}(u+i,v+j,\tau)W_{C1}(i,j,1).
\]
If there is a vertical line of spikes in the spiking image that matches up
with the kernel, then this result will be a maximum (maximum correlation of
the kernel with the image). The accumulated membrane potential for the neuron
at $(u,v)$ of map1 of the Conv1 layer is given by
\[
V_{m}(u,v,t,1)=\sum\limits_{\tau=0}^{t}\left(  \sum_{j=-2}^{2}\sum_{i=-2}%
^{2}s_{in}(u+i,v+j,\tau)W_{C1}(i,j,1)\right)  .
\]
The neuron at $(u,v)$ of map 1 of the Conv1 layer then spikes at time $t$ if
\[
V_{m}^{(1)}(u,v,t)\geq\gamma_{C1}%
\]
where $\gamma_{C1}$ is the threshold. If the neuron at $(u,v)$ in map 1 of
Conv1 spikes then a vertical line of spikes have been detected in the spiking
image centered at $(u,v)$.%

\begin{figure}[H]%
\centering
\includegraphics[
height=1.8429in,
width=3.3927in
]%
{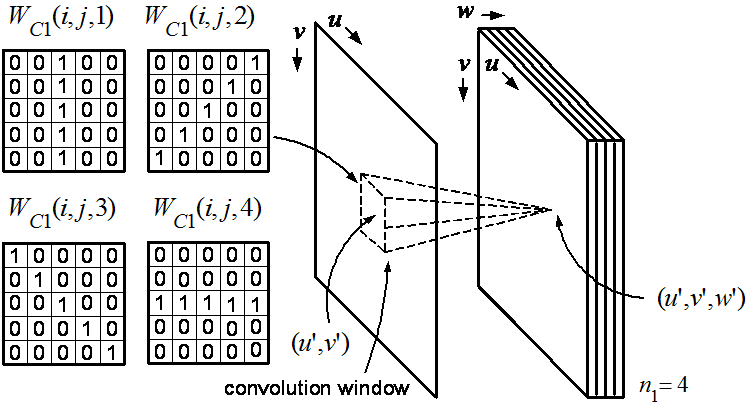}%
\caption{Feature detection.}%
\label{stdp_fig13d}%
\end{figure}

Figure \ref{stdp_fig13d} shows that map 2 (second feature map) of Conv1 is
used to detect a line of spikes at 45 degrees. The third feature map (map 3)
is used to detect a line of spikes at 135 degrees and the fourth feature map
(map 4) is used to detect a horizontal line of spikes.%

A typical SCNN has multiple layers. Each layer will have multiple feature maps
(simply, maps).

\section{Literature survey}

In $1951$ Hubel and Wiesel \cite{Hubel} showed that a cat's neurons in primary
visual cortex are tuned to simple features and the inner regions of the cortex
combined these simple features to represent complex features. The neocognitron
model was proposed in $1980$ by Fukushima to explain this behavior
\cite{Fukushima1980}. This model didn't require a "teacher" (unsupervised) to
learn the inherent features in the input, akin to the brain. The neocognitron
model is a forerunner to the spiking convolutional neural networks considered
in this work. These convolutional layers are arranged in layers to extract
features in the input data. The terminology "deep" CNNs refers to a network
with many such layers. However, the deep CNNs used in industry (Google,
Facebook, etc.) are fundamentally different in that they are trained using
supervision (back propagation of a cost function). Here our interest is to
return to the neocognitron model using spiking convolutional layers in which
all but the output layer is trained without supervision.

\subsection{Unsupervised networks}

A network equipped with STDP \cite{Markram} and lateral inhibition was shown
to develop orientation selectivity similar to the visual frontal cortex in a
cat's brain \cite{delorme2001} \cite{Zylberberg11asparse}. STDP was shown to
facilitate approximate Bayesian computation in the visual cortex using
expectation-maximization \cite{nessler}. STDP is used for feature extraction
in multi-layer spiking CNNs. It has been shown that deeper layers combine the
features learned in the earlier layers in order to represent advanced
features, but at the same time sparsity of the network spiking activity is
maintained \cite{Paulun} \cite{Kheradpisheh_2016} \cite{Kheradpisheh_2016b}
\cite{paul} \cite{Masquelier2007} \cite{Tavanaei} \cite{Tavanaei2017}
\cite{Zambrano} \cite{Ghodrati}. In \cite{Diehl} a fully connected networks
trained using unsupervised STDP and homeostasis achieved a 95.6\%
classification accuracy on the MNIST data set.

\subsection{Reward modulated STDP}

\label{rstdp} Mozafari et al. \cite{mozafari2} \cite{mozafari1} proposed
reward modulated STDP (R-STDP) to avoid using a support vector machine (SVM)
as a classifier. It has been shown that the STDP learning rule can find
spiking patterns embedded in noise \cite{Masquelier2008}. That is, after
unsupervised training, the output neuron spikes if the spiking pattern is
input to it. A problem with this unsupervised STDP approach is that as this
training proceeds the output neuron will spike when just the first few
milliseconds of the pattern have been presented. (For example, the pattern in
Figure \ref{rasterplots} is 5 msecs long and the output starts to spike when
only (say) the first 2 msecs of the pattern have been presented to it though
it should only spike after the full 5 msec pattern has been presented.
Mozafari et al showed in \cite{mozafari1} that R-STDP helps to alleviate this problem.

When unsupervised training methods are used, the features learned in the last
layer are used as input to an SVM classifier \cite{Kheradpisheh_2016b}%
\cite{Kheradpisheh_2016} or a simple two or three layer back propagation
classifier \cite{stromatias}. In contrast, R-STDP uses a reward or punishment
signal (depending upon if the prediction is correct or not) to update the
weights in the final layer of a multi-layer (deep) network. Spiking
convolutional networks are successful in extracting features \cite{mozafari1}%
\cite{Kheradpisheh_2016b}\cite{Kheradpisheh_2016}. Because R-STDP is a
supervised learning rule, the extracted features (reconstructed weights) more
closely resemble the object they detect and thus can (e.g.,) more easily
differentiate between a digit \textquotedblleft1\textquotedblright\ and a
digit "7" compared to STDP. That is, reward modulated STDP seems to compensate
for the inability of the STDP to differentiate between features that closely
resemble each other \cite{Florian} \cite{Legenstein} \cite{mozafari2}
\cite{Maxim}. It is also reported in \cite{mozafari2} that R-STDP is more
computationally efficient. However, R-STDP is prone to over fitting, which is
alleviated to some degree by scaling the rewards and punishments (e.g.,
receiving higher punishment for a false positive and a lower reward for a true
positive) \cite{mozafari2} \cite{mozafari1}. In more detail, the reward
modulated STDP learning rule is:

If a reward signal is generated then the weights are updated according to
\[%
\begin{cases}
\Delta w_{ij}=+\frac{N_{miss}}{N}a_{r}^{+}w_{ij}(1-w_{ij}) & \text{ if}%
\ t_{j}-t_{i}\leq0\\
\Delta w_{ij}=-\frac{N_{miss}}{N}a_{r}^{-}w_{ij}(1-w_{ij}) & \text{ if}%
\ t_{j}-t_{i}>0.
\end{cases}
\]

If a punishment signal is generated then the weights are updated according to
\[%
\begin{cases}
\Delta w_{ij}=-\frac{N_{hit}}{N}a_{p}^{+}w_{ij}(1-w_{ij}) & \text{ if}%
\ t_{j}-t_{i}\leq0\\
\Delta w_{ij}=+\frac{N_{hit}}{N}a_{p}^{-}w_{ij}(1-w_{ij}) & \text{ if}%
\ t_{j}-t_{i}>0.
\end{cases}
\]

Here $t_{j}$ and $t_{i}$ are the pre- and post-synaptic times, respectively.
For every $N$ input images, $N_{miss}$ and $N_{hit}$ are number of
misclassified and correctly classified samples. Note that $N_{miss}+N_{hit}=N
$, if the decision of the network is based on the maximum potential of the
network, if the decision of the network is based on the early spike
$N_{miss}+N_{hit}\leq N$ because there might be no spikes for some inputs.

\subsection{Spiking networks with back propagation}

\cite{Panda} used two unsupervised spiking CNNs for feature extraction. Then
initializing with these weights, they used a type of softmax cost function for
classification with the error back propagated through all layers. They were
able to obtain a classification accuracy 99.1\% on the MNIST data set. A
similar approach with comparable accuracy was carried by \cite{Tavanaei2018}.
Other methods such as computing the weights on conventional (non spiking) CNNs
trained using the back propagation algorithm and then converting them to work
on spiking networks have been shown to achieve an accuracy of 99.4\% on MNIST
data set and 91.35\% on CIFAR10 data set \cite{Lungu}. An approximate back
propagation algorithm for spiking neural networks was proposed in
\cite{Anwani} \cite{Lee}. In \cite{jin} a spiking\ CNN\ with
15C5-P2-40C5-P2-300-10 layers using error back propagation through all the
layers reported an accuracy of 99.49\% on the\ MNIST data set. The authors in
\cite{jin} also classified the N-MNIST data set using a fully connected
three-layer network with 800 neurons in the hidden layer and reported an
accuracy of 98.84\%.

Another approach to back propagation in spiking networks is the \emph{random
back} propagation approach. First the standard back propagation equations in
(non-spiking) neural networks is now summarized \cite{Nielsen}. The gradient
of a quadratic cost $C=\sum_{i=1}^{n_{0ut}}(y-a^{L})^{2}$ gives the error from
the last layer as
\begin{equation}
\delta^{L}=\frac{\partial C}{\partial a^{L}}\sigma^{\prime}(z^{L}).\label{eq1}%
\end{equation}
$a^{L}$ is the activation of the neurons in the output layer, $\sigma$ is the
activation function and $z$ is the net input to the output layer. This error
on the last layer is back propagated according to
\begin{equation}
\delta^{l}=((W^{l+1})^{T}\delta^{l+1})\odot\sigma^{\prime}(z^{l})\label{eq2}%
\end{equation}
where $W^{l+1}$ are the weights connecting the $l^{th}$ and ($l+1)^{th}$
layer. The weights and biases are updated as follows:%
\begin{equation}
\frac{\partial C}{\partial b_{j}^{l}}=\delta_{j}^{l}\label{eq3}%
\end{equation}

\begin{equation}
\frac{\partial C}{\partial W_{jk}^{l}}=a_{k}^{l-1}\delta_{j}^{l}\label{eq4}%
\end{equation}
In equation (\ref{eq2}), the weight matrix $W^{l+1}$ connecting the $l^{th}$
and ($l+1)^{th}$ layer is the same as the weight matrix used in forward
propagation to calculate the activations $a^{l+1}$ of $(l+1)^{th}$ layer. This
is bothersome to the neuroscience community as this is not biologically
plausible \cite{Liao} \cite{GROSSBERG198723} \cite{Aidan}. This is referred to
as the \emph{weight transport problem}. Lillicrap et al. \cite{Lillicrap}
showed that the back propagation algorithm works well even if $W^{l+1}$ in
equation (\ref{eq2}) is replaced with another fixed \emph{random} matrix
$(W^{\prime})^{l+1}$. This eliminates the requirement of weight symmetry,
i.e., the same weights for forward and backward propagations. A neuromorphic
hardware specific adaptation of random error back propagation that solves the
weight transport problem was introduced by \cite{Neftci} and was shown to
achieve an error rate of 1.96\% for the MNIST data set. The cost function in
\cite{Neftci} is defined as%
\begin{equation}
L_{sp}=0.5%
{\displaystyle\sum\limits_{i}}
(v_{i}^{p}(t)-v_{i}^{l}(t))^{2}\label{eq5}%
\end{equation}
where $e_{i}(t)$ is the error of the $i^{th}$ output neuron and $v^{p}$ and
$v^{l\text{ }}$are the firing rates of the prediction neuron and the label
neuron.%
\begin{equation}
\dfrac{\partial L_{sp}}{\partial W_{ij}}=-%
{\displaystyle\sum\limits_{i}}
e_{i}(t)\dfrac{\partial v_{i}^{p}(t)}{\partial W_{ij}}\label{eq6}%
\end{equation}
In equation (\ref{eq6}), $\dfrac{\partial v_{i}^{p}(t)}{\partial W_{ij}}$ was
approximated as%
\begin{equation}
\dfrac{\partial v_{i}^{p}(t)}{\partial W_{ij}}\propto%
\begin{cases}
1 & \text{ if}\ s_{j}^{h}(t)=1\text{ and }b_{\min}<I_{i}(t)<b_{\max}\\
0 & \text{ otherwise}%
\end{cases}
\label{eq7}%
\end{equation}
Where $I_{i}(t)$ is the current entering into $i^{th}$ post-synaptic neuron
and $s_{j}^{h}(t)=1$ indicates the presence of a pre-synaptic spike. For more
details see \cite{Neftci}. The weight update for the last layer is then%
\begin{equation}
\Delta W_{ij}^{E}\propto%
\begin{cases}
-e_{i}(t) & \text{ if}\ s_{j}^{h}(t)=1\text{ and }b_{\min}<\text{\ }%
I_{i}(t)<b_{\max}\\
0 & \text{ otherwise}%
\end{cases}
\label{eq8}%
\end{equation}
The weight update for hidden layers is
\begin{equation}
\Delta w_{ij}^{C}\propto%
\begin{cases}
-\sum_{k}g_{ik}e_{k}^{E}(t) & \text{ if}\ s_{j}^{C}(t)=1\text{ and }b_{\min
}<\text{\ }I_{i}(t)<b_{\max}\\
0 & \text{ otherwise}%
\end{cases}
\label{eq9}%
\end{equation}
where $e_{k}^{E}(t)$ denotes the error term of the $k^{th}$ neuron in the
output layer and $g_{ik}$ is a fixed random number as suggested by the random
back propagation algorithm. In the work to be reported below, random back
propagation is not used. Specifically, when back propagation is used below, it
is only between the penultimate and output layer making random back
propagation unnecessary.

\subsection{Spike encoding}

Spikes are either rate coded or latency coded \cite{Gollisch2008}
\cite{Kiselev} \cite{Reinagel} \cite{Butts}. Rate coding refers to the
information encoded by the number of spikes per second (more spikes per time
carries more information) In this case the spike rate is determined by the
mean rate of a Poisson process. Latency encoding refers to the information
encoded in the time of arrival of a spike (earlier spikes carry more
information). The raster plot of Figure \ref{rasterplots} shows that
spatiotemporal information is provided by the input spikes to the output
neuron. That is, which input neuron is spiking (spatio) and the time a neuron
spikes (temporal) is received by the output neuron. The spiking networks use
this spatiotemporal information to extract features (e.g., detect the pattern
in Figure \ref{rasterplots}) in the input data \cite{Gupta} \cite{Meftah2010}.

\subsection{Realtime spikes}

Image sensors (silicon retinas) such as ATIS \cite{atis} and eDVS \cite{edvs}
provide (latency encoded) spikes as their output. These sensors detect changes
in pixel intensities. If the pixel value at location $(u,v)$ increases then an
ON-center spike is produced while if the pixel value decreased an OFF-center
spike is produced. Finally, if the pixel value does not change, no spike is
produced. The spike data from an image sensor is packed using an address event
representation (AER \cite{Avila}) protocol and can be accessed using serial
communication ports. A recorded version of spikes from eDVS data set was
introduced in \cite{Pineda} and a similar data set of MNIST images recorded
with ATIS data set was introduced in \cite{nmnist}.

\section{Background}

\subsection{Spiking Images}

We have considered the standard $27\times27$ grey-scale
MNIST\ images\footnote{We removed the outer most pixels in the data set
\cite{mnist} giving $27\times27$ images.} \cite{mnist} and the spiking N-MNIST
data files \cite{nmnist} for our experiments. In the case of the MNIST\ images
we needed to convert them to spikes. This was done by first using both an
on-center and an off-center Difference of Gaussian (DoG) convolution filter
$\Gamma_{\sigma_{1},\sigma_{2}}(i,j)$\ for edge detection given by
\[
K_{\sigma_{1},\sigma_{2}}(i,j)=\left\{
\begin{array}
[c]{cl}%
\dfrac{1}{2\pi\sigma_{1}^{2}}e^{-\dfrac{i^{2}+j^{2}}{2\sigma_{1}^{2}}}%
-\dfrac{1}{2\pi\sigma_{2}^{2}}e^{-\dfrac{i^{2}+j^{2}}{2\sigma_{2}^{2}}} &
\text{for }-3\leq i\leq3,-3\leq j\leq3\\
0 & \text{otherwise}%
\end{array}
\right.
\]
where $\sigma_{1}=1,\sigma_{2}=2$ for the on-center and $\sigma_{1}%
=2,\sigma_{2}=1$ for the off-center.%
\begin{figure}[H]%
\centering
\includegraphics[
height=1.772in,
width=4.3284in
]%
{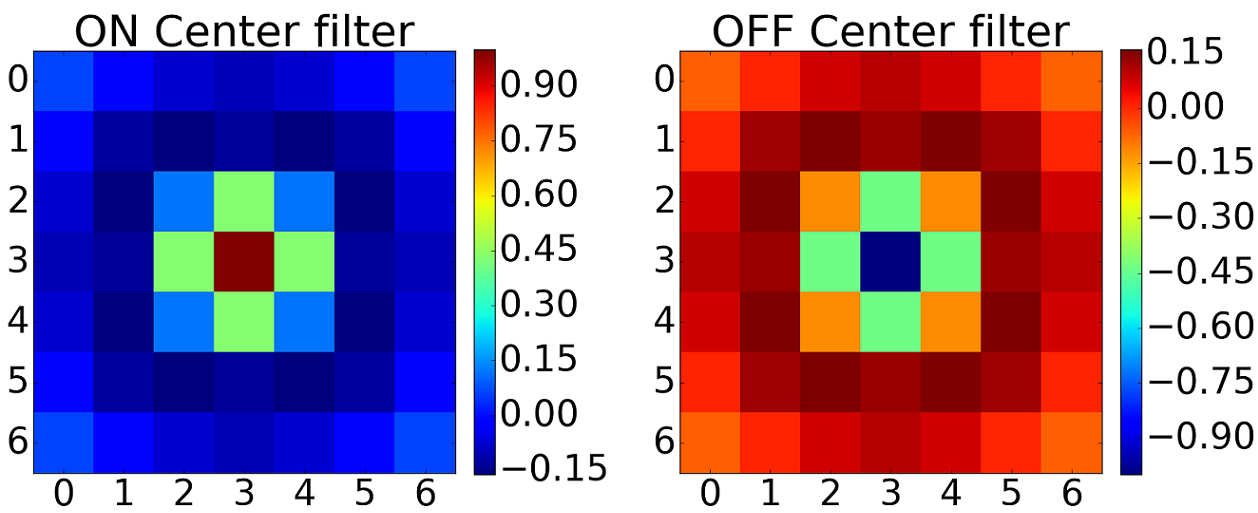}%
\caption{On center filter has higher values in the center whereas the off
center filter has lower values in the center. Colour code indicates the filter
values.}%
\label{onofffilters}%
\end{figure}
With the input image $\mathbf{I}_{in}(u,v)\in%
\mathbb{R}
^{27\times27}$, the output of each of the two DoG filters is computed using
the \emph{same} mode convolution
\[
\Gamma_{\sigma_{1},\sigma_{2}}(u,v)=\sum_{j=-3}^{j=3}\sum_{i=-3}%
^{i=3}\mathbf{I}_{in}(u+i,v+j)K_{\sigma_{1},\sigma_{2}}(i,j)\text{ \ for
}0\leq u\leq26,0\leq v\leq26.
\]%
\begin{figure}[H]%
\centering
\includegraphics[
height=1.75in,
width=5.0in
]%
{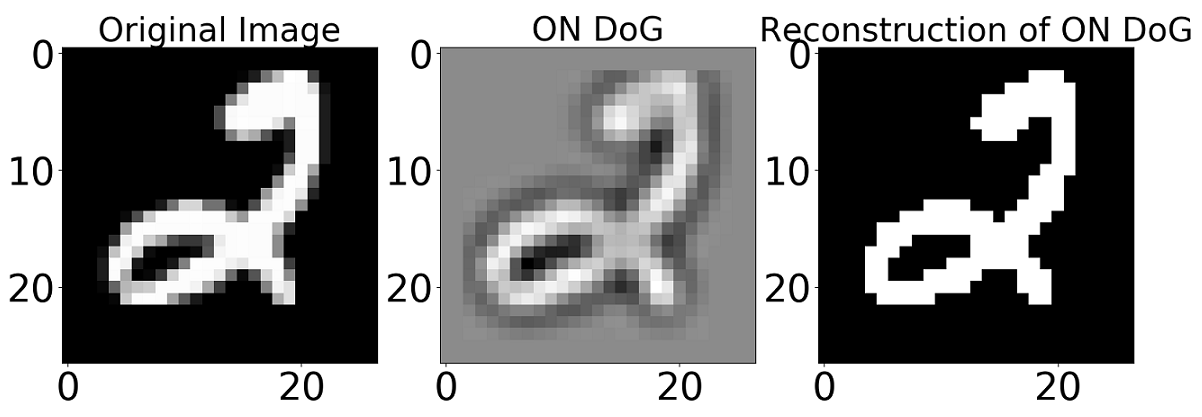}%
\caption{Left: Original grey-scale image. Center: Output of the ON DoG filter.
Right:\ Accumulation of spikes (white indicates a spike black indicates no
spike).}%
\label{new_on_centre}%
\end{figure}
\begin{figure}[H]%
\centering
\includegraphics[
height=1.75in,
width=5.01in
]%
{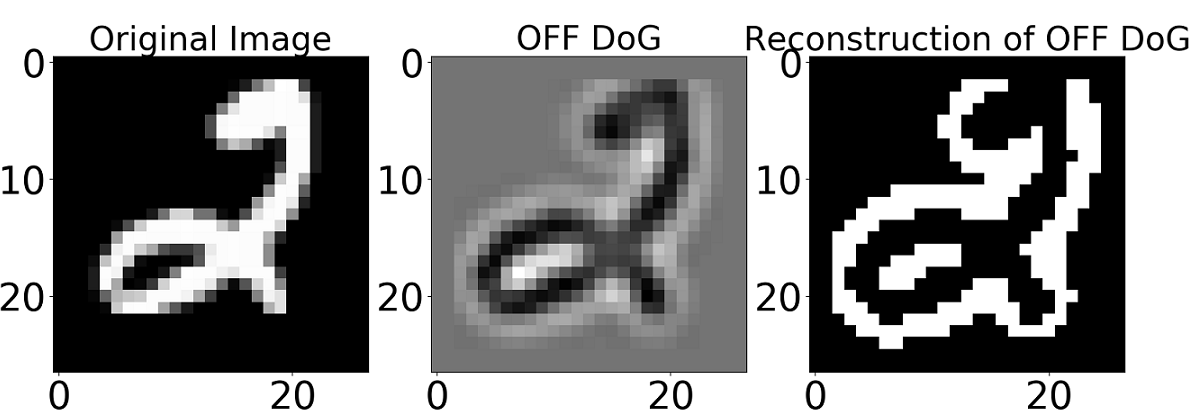}%
\caption{Left: Original grey-scale image. Center: Output of the OFF DoG
filter. Right:\ Accumulation of spikes (white indicates a spike black
indicates no spike).}%
\label{new_off_centre}%
\end{figure}

Then these two resulting \textquotedblleft images\textquotedblright\ were then
converted to an on and an off spiking image by \ At each location $(u,v) $ of
the output image $\Gamma_{\sigma_{1},\sigma_{2}}(u,v)$ a unit spike
$s_{(u,v)}$ is produced if and only if (\cite{Kheradpisheh})
\[
\Gamma_{\sigma_{1},\sigma_{2}}(u,v)>\gamma_{DoG}=50.
\]

The spike signal $s_{(u,v)}(t)$ is temporally coded (rank order
coding\cite{delorme2001}) by having it delayed \textquotedblleft
leaving\textquotedblright\ the Difference of Gaussian image $\Gamma
_{\sigma_{1},\sigma_{2}}(u,v)$ by the amount
\[
\tau_{(u,v)}=\frac{1}{\Gamma_{\sigma_{1},\sigma_{2}}(u,v)}\text{ in
milliseconds}.
\]%
\begin{figure}[H]%
\centering
\includegraphics[
height=0.8302in,
width=2.316in
]%
{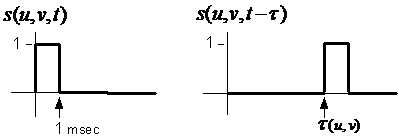}%
\caption{Spike signal }%
\label{stdp_fig11b}%
\end{figure}
That is, the more $\Gamma_{\sigma_{1},\sigma_{2}}(u,v)$ exceeds the threshold
$\gamma_{DoG}$ the sooner it leaves $\Gamma_{\sigma_{1},\sigma_{2}}(u,v)$ or
equivalently, the value of $\Gamma_{\sigma_{1},\sigma_{2}}(u,v)$ is encoded in
the value $\tau_{(u,v)}.$

\begin{figure}[H]%
\centering
\includegraphics[
height=2.706in,
width=5.0695in
]%
{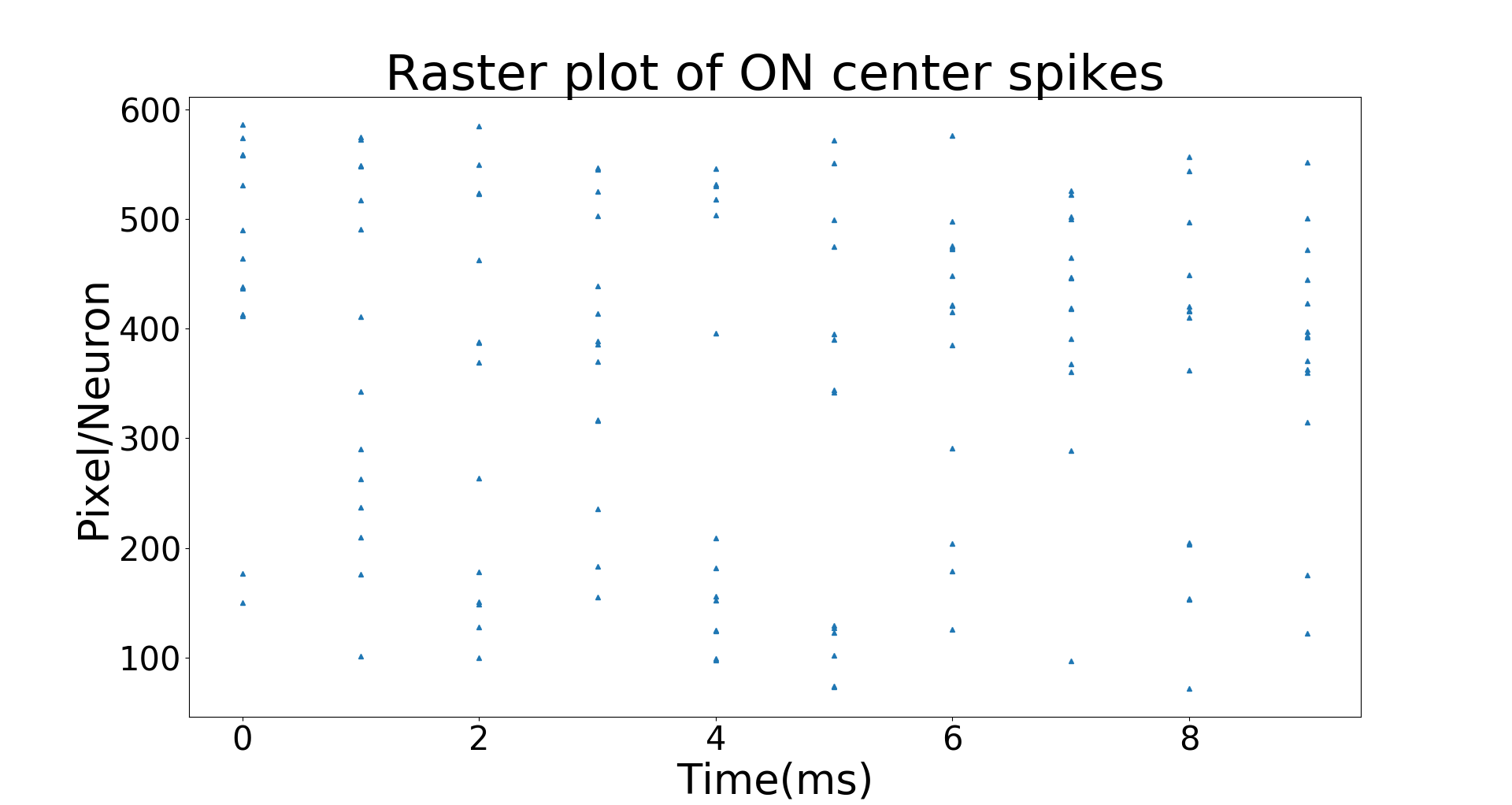}%
\caption{Rasterplot of spikes for an on centre cell. Blue dots in the plot
indicates the presence of a spike for a particular neuron and bin (timestep).}%
\label{new_on_centre_raster_plot}%
\end{figure}

For all experiments the arrival times of the spikes were sorted in ascending
order and then (approximately) equally divided into 10 bins (10 times in
Figure \ref{new_on_centre_raster_plot}). The raster plot shows which neurons
(pixels of $\Gamma_{\sigma_{1},\sigma_{2}}(u,v)$) spiked to make up bin 1
(time 0), bin 2 (time 1), etc. Figure \ref{new_on_centre_raster_plot} shows an
example for ON center cell spikes. In all the experiments each image is
encoded into 10 msec (10 bins) and there is a 2 msec silent period between
every image.%

\subsection{Network Description}

\begin{figure}[H]%
\centering
\includegraphics[
height=2.5901in,
width=3.9902in
]%
{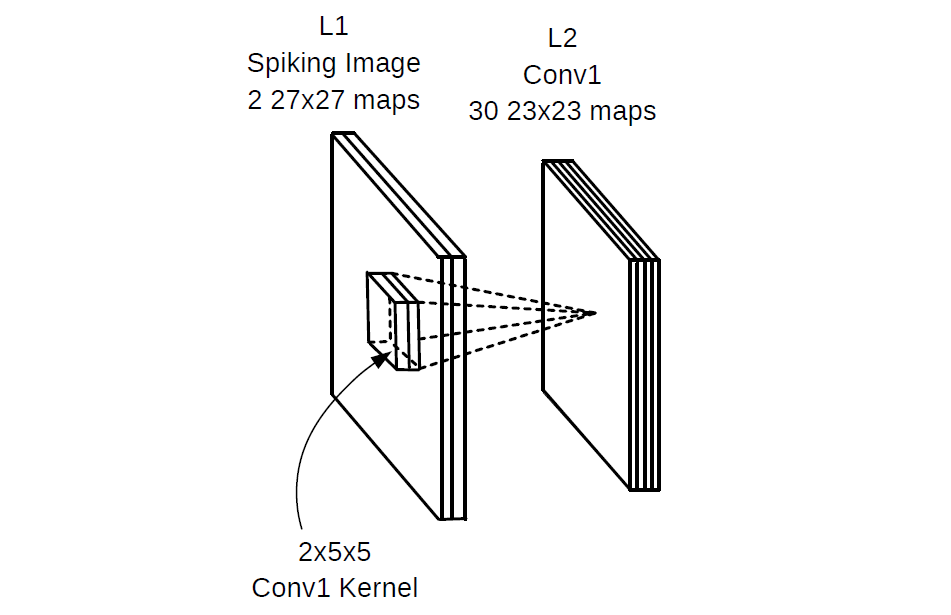}%
\caption{Demonstration of convolution with a 3D kernel.}%
\label{3dconv_intro}%
\end{figure}

We have a similar network as in \cite{Kheradpisheh_2016}%
\cite{Kheradpisheh_2016b} as illustrated in Figure \ref{3dconv_intro}. We let
$s_{L1}(t,k,u,v)$ denote the spike signal at time $t$ emanating from the
$(u,v)$ neuron of spiking image $k$ where $k=0$ (ON center) or $k=1$ (OFF
center). The L2 layers consists of 30 maps with each map having its own
convolution kernel (weights) of the form
\[
W_{C1}(w,k,i,j)\in%
\mathbb{R}
^{2\times5\times5}\text{ \ for \ }w=0,1,2,...,29
\]
The \textquotedblleft membrane potential\textquotedblright\ of the $(u,v)$
neuron of map $w$ ($w=0,1,2,...,29$) of L2 at time $t$ is given by the
\emph{valid} mode convolution%
\[
V_{L2}(t,w,u,v)=\sum_{\tau=0}^{t}\left(  \sum_{k=0}^{1}\sum_{i=0}^{4}%
\sum_{i=0}^{4}s_{L1}(\tau,k,u+i,v+j)W_{C1}(w,k,i,j)\right)  \text{\ \ for
}(0,0)\leq(u,v)\leq(22,22)
\]
If at time $t$ the potential%
\[
V_{L2}(t,w,u,v)>\gamma=15
\]
then the neuron at $(w,u,v)$ emits a unit spike.

\subsubsection{Convolution Layers and STDP}

At any time $t,$ \emph{all} of the potentials $V_{L2}(t,w,u,v)$ for
$(0,0)\leq(u,v)\leq(22,22)$ and $w=0,1,2,...,29$ are computed (in theory this
can all be done in parallel) with the result that neurons in different
locations within a map and in different maps may have spiked. In particular,
at the location $(u,v)$ there can be multiple spikes (up to 30) produced by
different maps. The desire is to have different maps learn different features
of an image. To enforce this learning, \emph{lateral inhibition} and
\emph{STDP\ competition} are used \cite{Kheradpisheh_2016}.

\paragraph{\textbf{Lateral Inhibition}}

\begin{figure}[H]%
\centering
\includegraphics[
height=2.3065in,
width=4.3223in
]%
{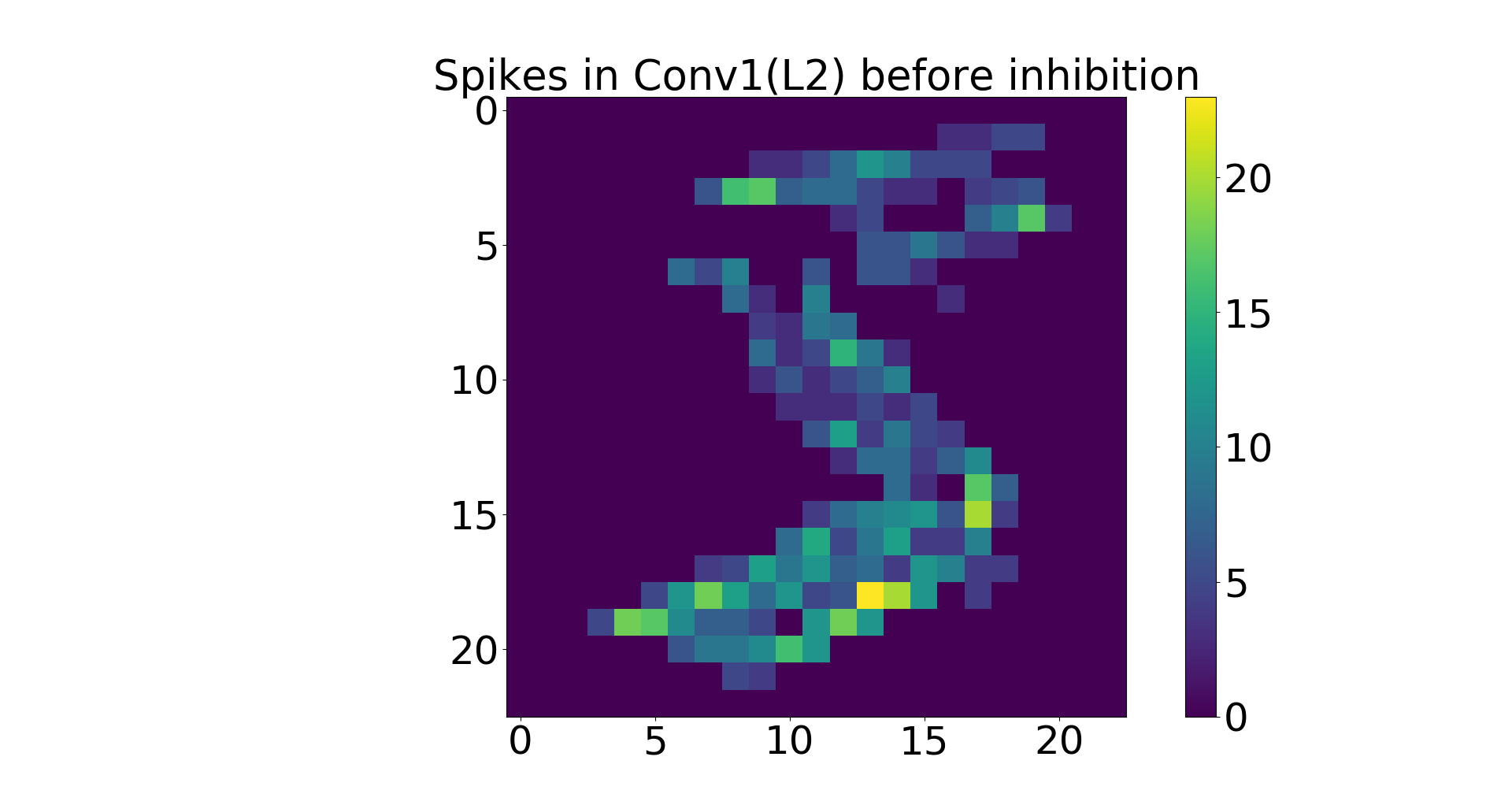}%
\caption{Accumulation of spikes in L2 \emph{without} lateral inhibition.}%
\label{l2ip2b4inhibition}%
\end{figure}

To explain lateral inhibition, suppose at the location $(u,v)$ there were
potentials $V_{L2}(t,w,u,v)$ in different maps $w$ at time $t$ that exceeded
the threshold $\gamma$ Then the neuron in the map with the highest potential
$V_{L2}(t,w,u,v)$ at $(u,v)$ inhibits the neurons in all the other maps at the
location $(u,v)$ from spiking till the end of the present image (even if their
potential exceeded the threshold). Figure \ref{l2ip2b4inhibition} shows the
accumulated spikes (from an MNIST\ image of \textquotedblleft%
5\textquotedblright) from all 30 maps at each location $(u,v)$ with lateral
inhibition \emph{not} being imposed. For example, at location (19,14) in
Figure \ref{l2ip2b4inhibition} the color code is yellow indicating in excess
of 20 spikes, i.e., more than 20 of the maps produced a spike at that
location.

\begin{figure}[H]%
\centering
\includegraphics[
height=2.3385in,
width=4.382in
]%
{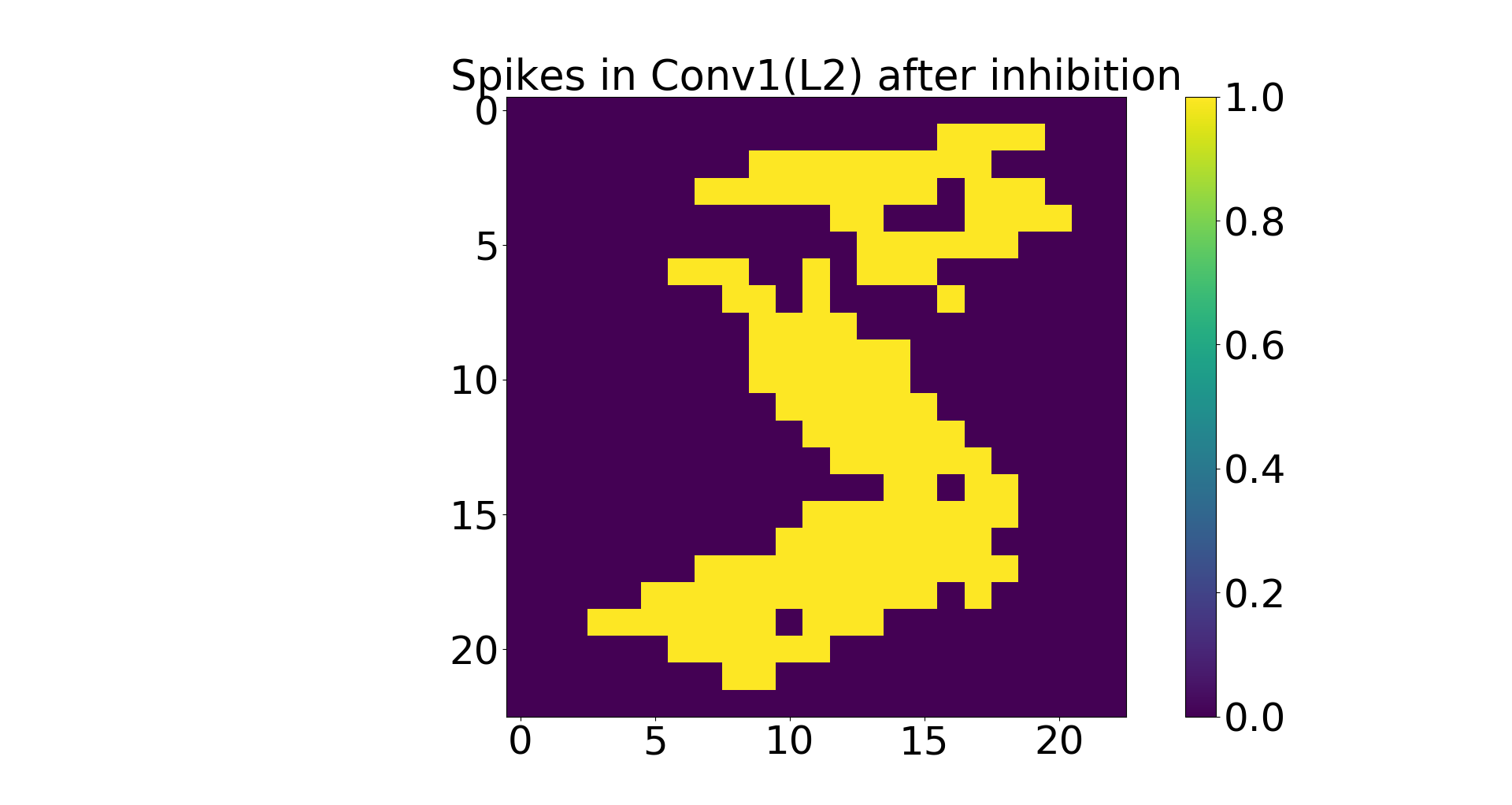}%
\caption{Accumulation of spikes from the MNIST digit \textquotedblleft%
5\textquotedblright\ with lateral inhibition imposed. }%
\label{l2ip2aftrinhibition}%
\end{figure}

Figure \ref{l2ip2aftrinhibition} shows the accumulation of spikes from all 30
maps, but now with lateral inhibition imposed. Note that at each location
there is either 1 spike or no spike as indicated by the color code.

\paragraph{STDP\ Competition}

After lateral inhibition, we consider each map that had one or more neurons
whose potential $V$ exceeded $\gamma.$ Let these maps be $w_{k1}%
,w_{k2},...,w_{km}$ where\footnote{The other maps did not have any neurons
whose membrane potential crossed the threshold and therefore cannot spike.}
$0\leq k_{1}<k_{2}<\cdots<k_{m}\leq29$. Then in each map $w_{ki}$ we locate
the neuron in that map that has the maximum potential value. Let
$(u_{k1},v_{k1}),(u_{k2},v_{k2}),...,(u_{km},v_{km})$ be the location of these
maximum potential neurons in each map. Then neuron $(u_{ki},v_{ki})$ inhibits
all other neurons in map $w_{ki}$ from spiking for the remainder of the time
steps of that spiking image. Further, these $m$ neurons can inhibit each other
depending on their relative location as we now explain. Suppose neuron
$(u_{ki},v_{ki})$ of map $w_{ki}$ has the highest potential of these $m$
neurons. Then, in an $11\times11$ area centered about $(u_{ki},v_{ki}),$ this
neuron inhibits all neurons of all the other maps in the same $11\times11$
area. Next, suppose neuron $(u_{kj},v_{kj})$ of map $w_{kj}$ has the second
highest potential of the remaining $m-1$ neurons. If the location
$(u_{kj},v_{kj})$ of this neuron was within the $11\times11$ area centered on
neuron $(u_{ki},v_{ki})$ of map $w_{ki},$ then it is inhibited. Otherwise,
this neuron at $(u_{kj},v_{kj})$ inhibits all neurons of all the other maps in
a $11\times11$ area centered on it. This process is continued for the
remaining $m-2$ neurons. In summary, there can be no more than one neuron that
spikes in the same $11\times11$ area of all the maps.

\begin{figure}[H]%
\centering
\includegraphics[
height=2.3765in,
width=4.4547in
]%
{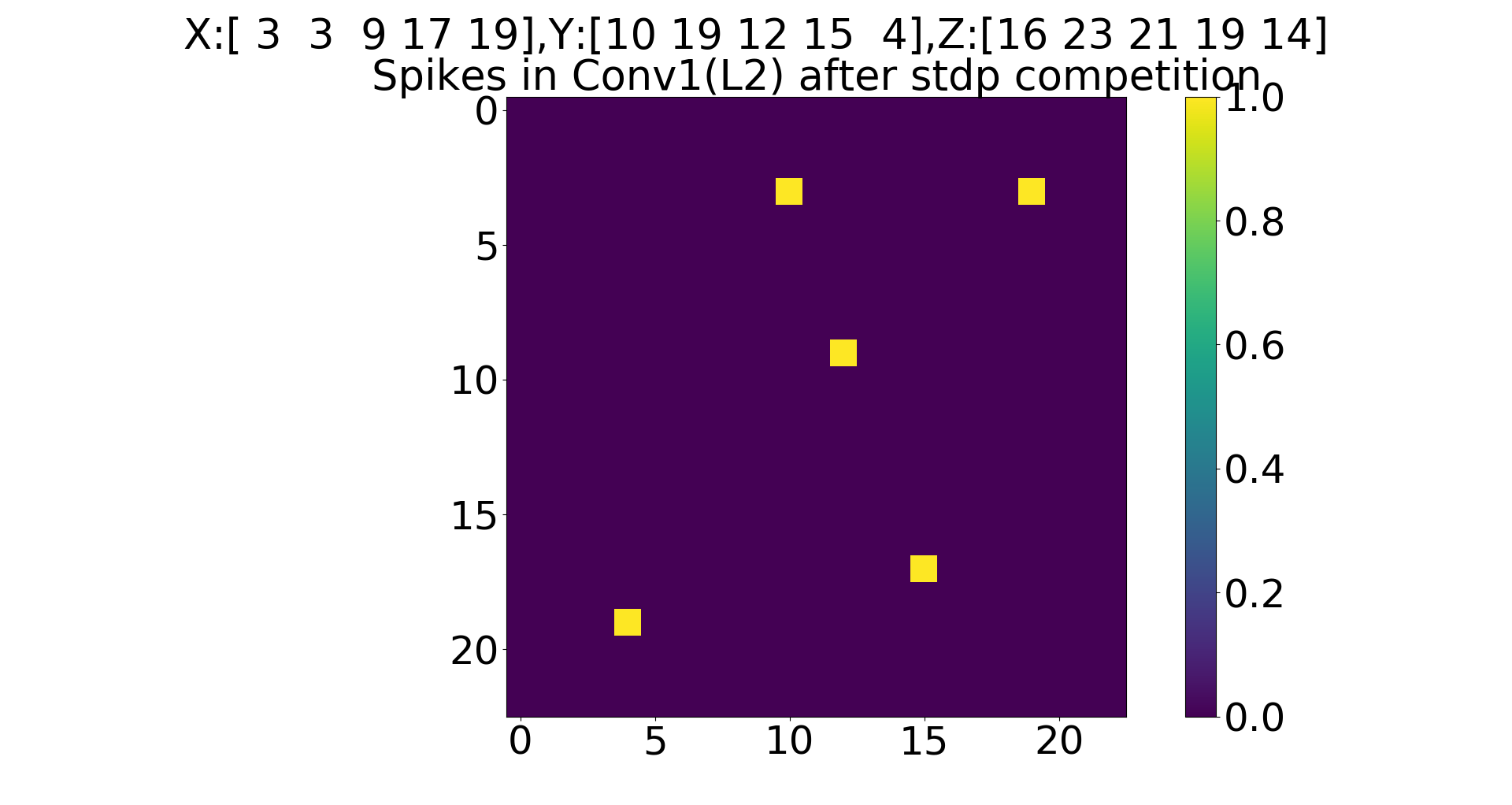}%
\caption{Accumulation of spikes with both lateral inhibition and
STDP\ competition imposed. }%
\label{l2ip2aftrstdpcomptn}%
\end{figure}

Figure \ref{l2ip2aftrstdpcomptn} shows the spike accumulation after both
lateral inhibition and STDP competition have been imposed. The figure shows
that there is at most one spike from all the maps in any $11\times11$ area.%

Lateral Inhibition and STDP inhibition enforce sparse spike activity and, as a
consequence, the network tends to spike sparsely

\paragraph{\textbf{Spike Timing Dependent Plasticity (STDP)}}

Only those maps that produced a spike (with lateral inhibition and STDP
competition imposed) have their weights (convolution kernels) updated using
spike timing dependent plasticity. Let $w_{ij}$ be the weight connecting the
$j^{th}$ pre-synaptic neuron in the L1 layer to $i^{th}$ post-synaptic neuron
in the L2 layer. If the $i^{th}$ post-synaptic neuron spikes at time $t_{i}$
with the pre-synaptic neuron spiking at time $t_{j}$ then the weight $w_{ij}$
is updated according to the simplified STDP\ rule \cite{delorme2001}%
\[
w_{ij}\longleftarrow w_{ij}+\Delta w_{ij},\text{ \ where \ }\Delta w_{ij}=%
\begin{cases}
+a^{+}w_{ij}(1-w_{ij})\ \ \text{if }\ t_{i}>t_{j}\\
-a^{-}w_{ij}(1-w_{ij})\ \ \text{otherwise.}%
\end{cases}
\]
The parameters $a^{+}>0$ and $a^{-}>0$ are referred to as learning rate
constants. $a^{+}$ is initialized to $0.004$ and $a^{-}$ is initialized to
$0.003$ and are increased by a factor of 2 after every 1000 spiking images.
STDP is shown to detect a hidden pattern in the incoming spike data
\cite{Masquelier2008}. In all of our experiments we used the above simplified
STDP model as in \cite{Kheradpisheh_2016} (simplified STDP refers to the
weight update not depending on the exact time difference between pre-synaptic
and post-synaptic spikes). If the pre-synaptic neuron spikes before
post-synaptic neuron then the synapse is strengthened, if the pre-synaptic
neuron doesn't spike before post-synaptic neuron then it is assumed that the
pre-synaptic neuron will spike later and the synapse is weakened.

\begin{figure}[H]%
\centering
\includegraphics[
height=3.4999in,
width=6.5587in
]%
{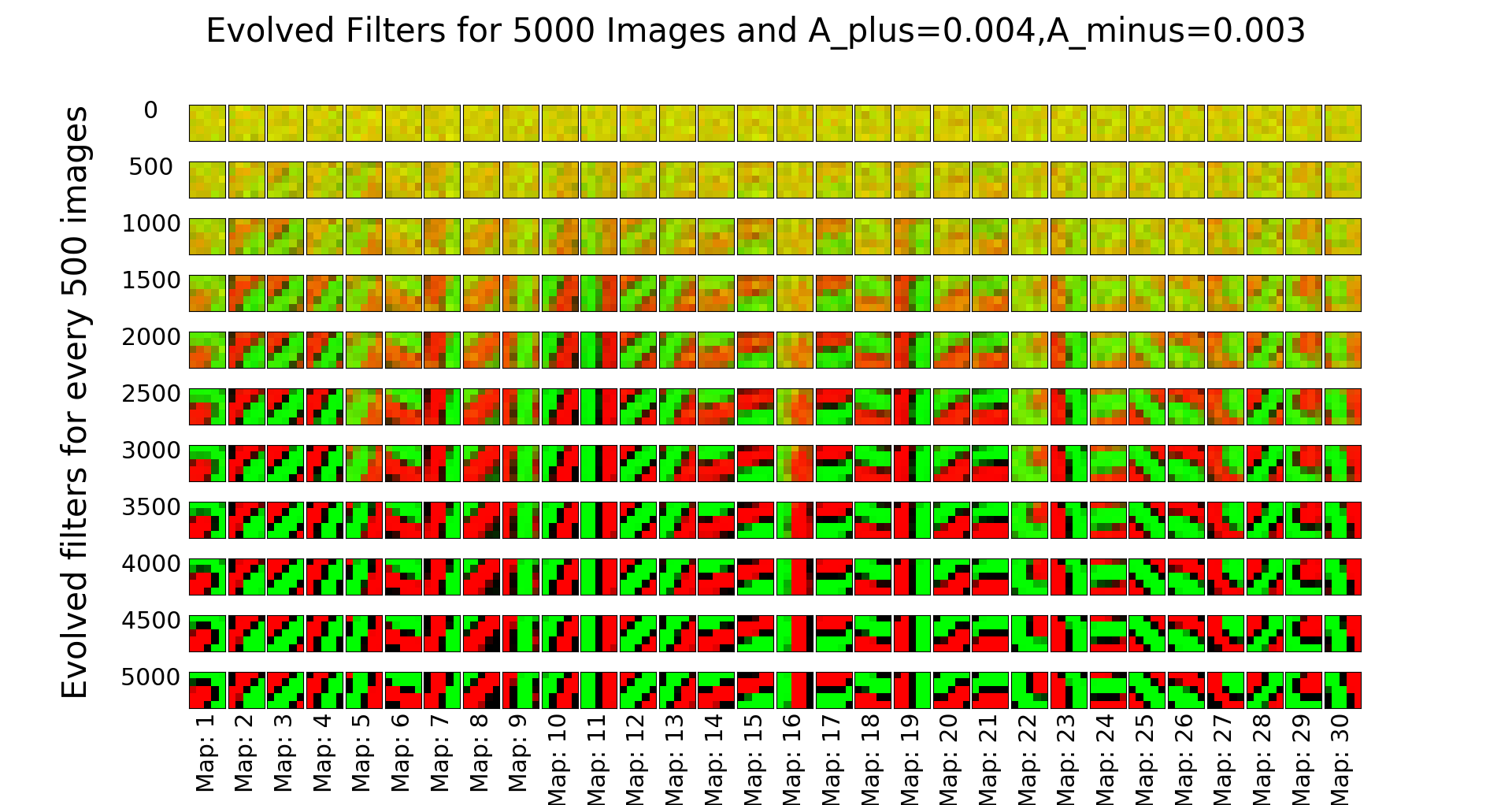}%
\caption{Plot of the weights of 30 maps of L2. The ON (green) $5\times5$
filter and the OFF (red) $5\times5$ filter are superimposed on top of each
other. }%
\label{l1l2evolvedfilters}%
\end{figure}

Figure \ref{l1l2evolvedfilters} is a plot of the weights (convolution kernels)
for each of the 30 maps. Following \cite{Kheradpisheh_2016}, each column
corresponds to a map and each row presents the weights after every 500 images.
For example, $W_{C1}(29,k,i,j)$ for $k=0,1$ and $(0,0)\leq(i,j)\leq(26,26)$
are the weights for the ON (green) and OFF (red) filters\footnote{That is, the
ON (green) and Off (red) weight are superimposed on the same plot.} for the
$30^{th}$ map (right-most column of Figure \ref{l1l2evolvedfilters}). It
turned out that there were approximately 17 spikes per image in this layer
(L2). At the end of the training most of the synapses will be saturated either
at 0 or 1.

\paragraph{\textbf{Homeostasis}}

Homeostasis refers to the convolution kernels (weights) for all maps being
updated approximately the same number of times during training. With
homeostasis each kernel gets approximately the same number of opportunities to
learn its unique feature. Some maps tend to update their weights more than
others and, if this continues, these maps can take over the learning. That is,
only the features (weights of the convolution filter) of those maps that get
updated often will be of value with the rest of the maps not learning any
useful feature (as their weights are not updated). Homeostasis was enforced
by\ simply decreasing the weights of a map by $w_{ij}\rightarrow w_{ij}%
-a^{-}w_{ij}(1-w_{ij})$ if it tries to update more than twice for every 5 of
input images.

\subsubsection{Pooling Layers}

A pooling layer is a way to down sample the spikes from the previous
convolution layer to reduce the computational effort.

\paragraph{Max Pooling}

After the synapses (convolution kernels or weights) from L1 to L2 have been
learned (unsupervised STDP\ learning is over\footnote{And therefore
STDP\ competition is no longer enforced.}), they are fixed, but lateral
inhibition continues to be enforced in L2. Spikes from the maps of the
convolution layer L2 are now passed on to layer L3 using max pooling. First of
all, we ignored the last row and last column of each of the $23\times23$ maps
of L2 so that they may be considered to be $22\times22.$ \ Next, consider the
first map of the convolution layer L2. This map is divided into
non-overlapping $2\times2$ area of neurons. In each of these $2\times2$ sets
of neurons, at most one spike is allowed through. If there is more than one
spike coming from the $2\times2$ area, then one compares the membrane
potentials of the spikes and passes the one with the highest membrane
potential. Each $2\times2$ set of neurons in the first map is then a single
neuron in the first map of the L3 layer. Thus each map of L3 has $11\times11$
(down sampled) neurons. This process is repeated for all the maps of L2 to
obtain the corresponding maps of L3. Lateral inhibition is not applied in a
pooling layer. There is no learning done in the pooling layer, it is just way
to decrease the amount of data to reduce the computational effort.

\begin{figure}[H]%
\centering
\includegraphics[
height=4.0in,
width=7.25in
]%
{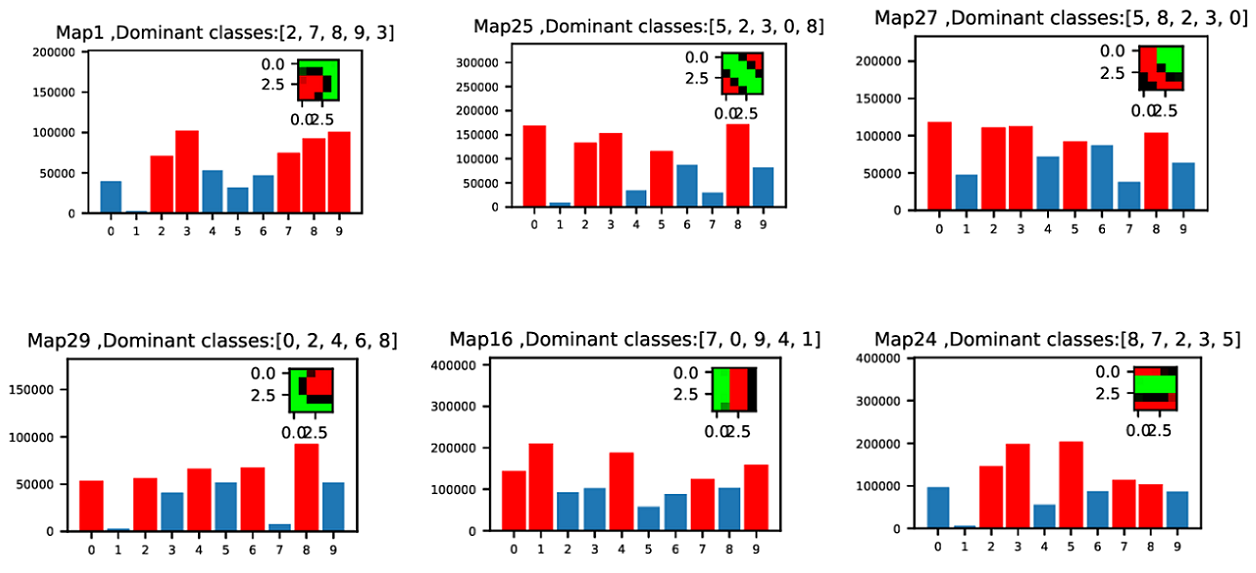}%
\caption{Spikes per map per digit. Headings for each of the sub-plots indicate
the dominant (most spiking) digit for respective features.}%
\label{l3spikespermapperdigit}%
\end{figure}

After training the L2 convolution layer, we then passed 60,000 MNIST digits
through the network and recorded the spikes from the L3 pooling layer. This is
shown in Figure \ref{l3spikespermapperdigit}. For example, in the upper
left-hand corner of Figure \ref{l3spikespermapperdigit} is shown the number
spikes coming out of the first map of the pooling layer L3 for each of the 10
MNIST\ digits. It shows that the digit \textquotedblleft3\textquotedblright%
\ produced over 100, 000 spikes when the 60,000 MNIST digits were passed
through the network while the digit \textquotedblleft1\textquotedblright%
\ produced almost no spikes. That is, the spikes coming from digit
\textquotedblleft1\textquotedblright\ do not correlate with the convolution
kernel (see the inset) to produce a spike. On the other hand, the digit "3"
almost certainly causes a spike in the first map of the L3 pooling layer. In
the bar graphs of Figure \ref{l3spikespermapperdigit} the red bars are the
five MNIST digits that produced the most spikes in the L3 pooling layer while
the blue bars are the five MNIST digits that produced the least.

\begin{figure}[H]%
\centering
\includegraphics[
height=2.975in,
width=5.1828in
]%
{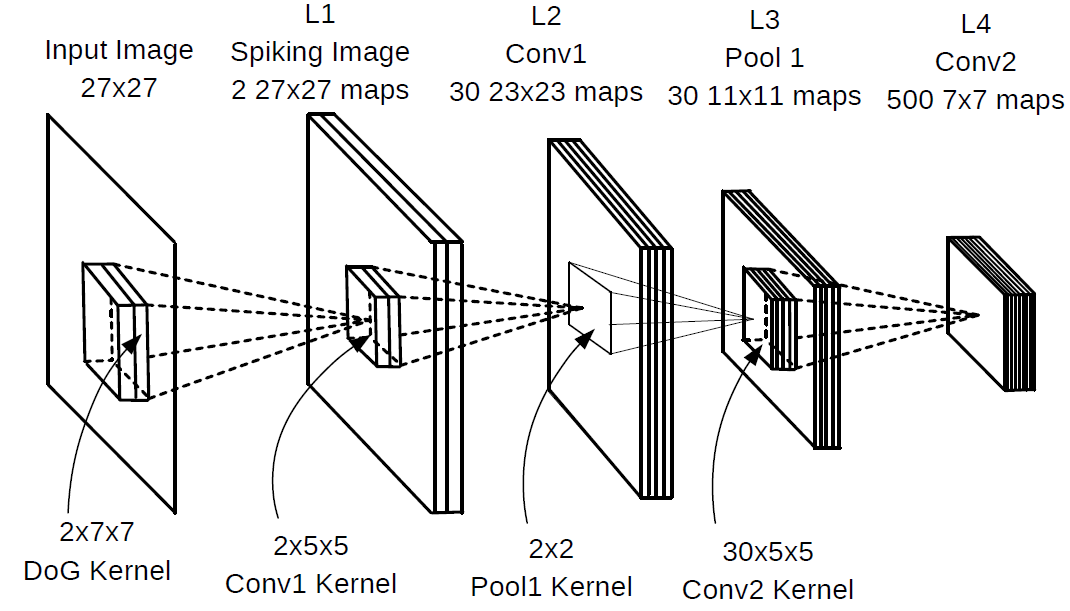}%
\caption{Network showing two convolution layers.}%
\label{2c2p_mnistnetwork}%
\end{figure}

Figure \ref{2c2p_mnistnetwork} shows convolution kernel between the L3 pooling
layer and the L4 convolution layer. We chose to have 500 maps in L4 which
means that for $w=0,1,2,...,499$ we have
\[
W_{C2}(w,k,i,j)\in%
\mathbb{R}
^{30\times5\times5}\text{ \ for \ }0\leq k\leq29\text{ \ and }(0,0)\leq
(i,j)\leq(4,4).
\]

The spikes from the L3 pooling layer are then used to train the weights
(convolutional kernels) $W_{C2}$ in the same manner as $W_{C1}.$

In some of our experiments we simply did a type of global pooling to go to the
output layer L5. Specifically, at each time step, we convolve the spikes from
L3 to compute the potential for each of the $500\times7\times7$ neurons of L4.
The maximum potential for each map in L4 was then found and stored (This is a
vector in $%
\mathbb{R}
^{500}$). The potentials in L4 were then reset to 0 and the process repeated
for each of the remaining time steps of the current image. This procedure
results in ten $%
\mathbb{R}
^{500}$ vectors for each image. The sum of these vectors then encodes the
current image in L5, i.e., as a single vector in $%
\mathbb{R}
^{500}.$ The motivation to take the maximum potential of each map at each time
step is because all the neurons in a given map of L4 are looking for the
\emph{same} feature in the current image.%

\begin{figure}[H]%
\centering
\includegraphics[
height=2.5443in,
width=5.2382in
]%
{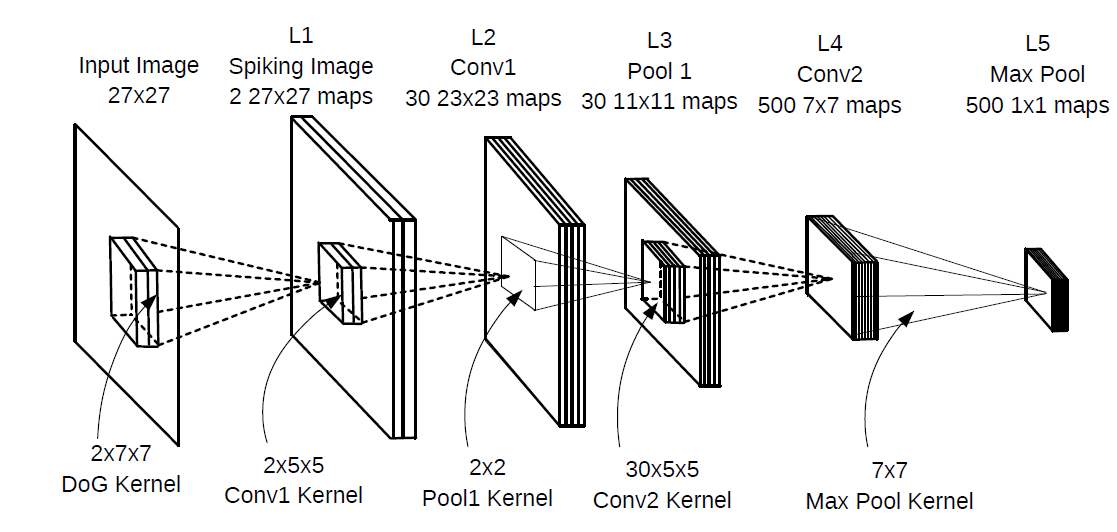}%
\caption{Network with two convolution and pooling layers, global pooling is
also shown here.}%
\label{mnistnetwork_maxpots2}%
\end{figure}

Unsupervised STDP training is done in the convolution layers with both STDP
competition and lateral inhibition applied to the maps of the convolution
layer doing training. Once a convolution layer is trained, it's weights are
fixed and the spikes are passed through it with only lateral inhibition imposed.

\section{Classification of MNIST data set}

In the following subsections we considered two different network architectures
along with different classifiers for the MNIST data set.

\subsection{Classification with Two Convolution/Pool Layers}

In this first experiment the architecture shown in Figure
\ref{mnistnetwork_maxpots2} was used. Max pooled "membrane potentials", i.e.,
the L5 layer of Figure \ref{mnistnetwork_maxpots2}, was used to transform each
$27\times27$ ($=729)$ training image into a new "image" in $R^{500}$. Using
these images along with their labels, a support vector machine \cite{svm} was
then used to find the hyperplanes that optimally\footnote{In is optimal in the
sense that a Lagrangian was minimized.} separate the training digits into 10
classes. With $W\in%
\mathbb{R}
^{45\times500}$ the SVM weights, the quantity $\lambda W^{T}W$ was added to
the SVM Lagrangian to for regularization. Both linear and radial basis
function (RBF) kernels were used in the SVM. We used 20,000 MNIST\ images for
the (unsupervised) training of the two convolution/pool layers (Layers L2-L5).
Then we used 50,000 images to train the SVM with another 10,000 images used
for validation (to determine the choice of $\lambda$). The SVM gives the
hyperplanes that optimally separate the 10 classes of digit. Table 1 shows
classification accuracies when $500$ maps were used in L4. The first two rows
of Table 2 give the test accuracy on 10,000 MNIST test images. In particular,
note a 98.01 \% accuracy for the RBF\ SVM and a 97.8 \% accuracy for a Linear
SVM. Using a similar network with linear SVM, Kheradpisheh et al.
\cite{Kheradpisheh_2016} reported an accuracy of 98.3\%.

\begin{table}[th]%
\begin{tabular}
[c]{|c|c|c|c|c|c|c|}\hline
\textbf{Classifier} & \textbf{Test Acc } & \textbf{Valid Acc } &
\textbf{Training Time} & $\lambda$ & $\eta$ & \textbf{Epochs}\\\hline
RBF SVM & 97.92 \% & 97.98 \% & 8 minutes & 1/3.6 & - & -\\\hline
Linear SVM & 97.27 \% & 97.30 \% & 4 minutes & 1/0.012 & - & -\\\hline
2 Layer FCN (backprop) & 96.90 \% & 97.02 \% & 15 minutes & 1.0 & $\frac
{0.1}{(1.007)^{\#Epoch}}$ & 30\\\hline
3 layer FCN (backprop) & 97.8 \% & 97.91 \% & 50 minutes & 6.0 & $\frac
{0.1}{(1.007)^{\#Epoch}}$ & 30\\\hline
\end{tabular}
\caption{Classification accuracies on MNIST data set with various classifiers
when number of maps in L4 is 500.}%
\end{table}

\begin{figure}[H]%
\centering
\includegraphics[
height=2.9164in,
width=5.8612in
]%
{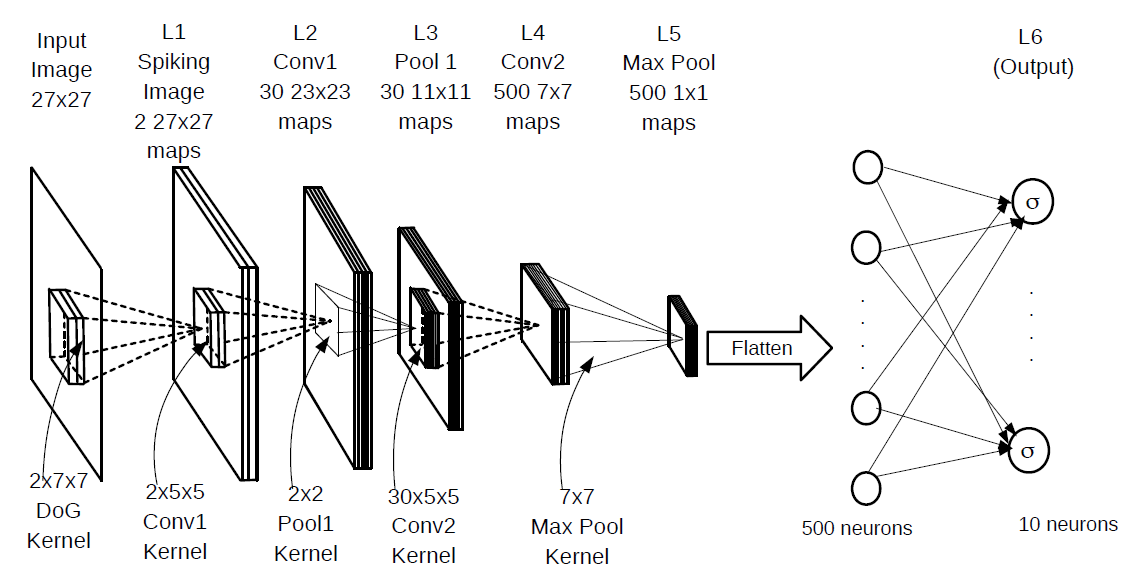}%
\caption{Network showing with 2 fully connected layers as a classifier.}%
\label{2c2p2fc_mnistnetwork}%
\end{figure}

\begin{figure}[H]%
\centering
\includegraphics[
height=3.1506in,
width=5.9339in
]%
{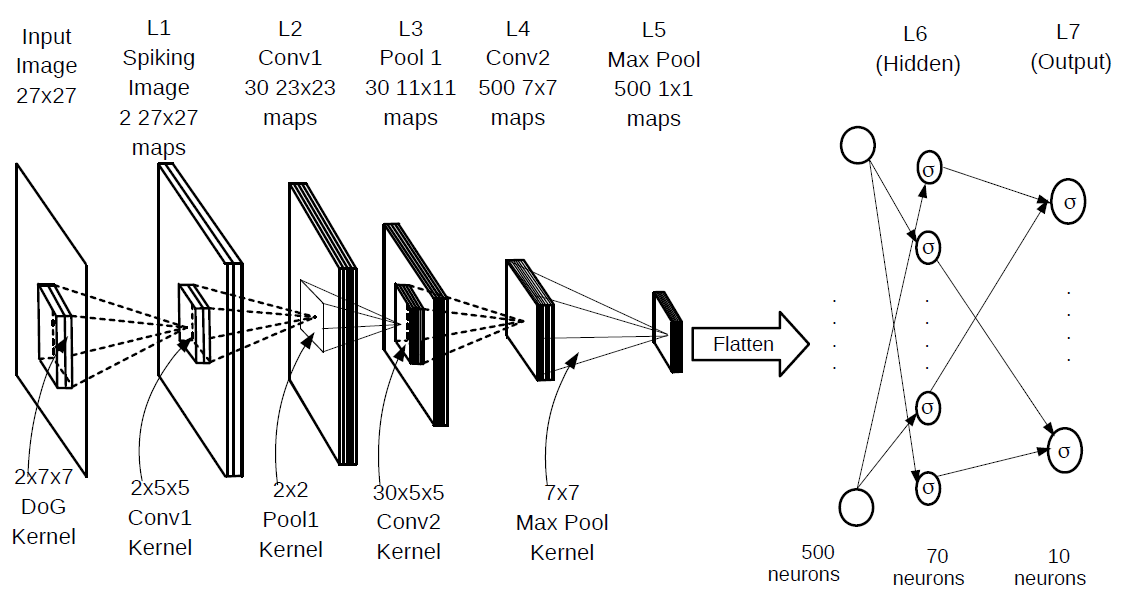}%
\caption{Network showing with 2 fully connected layers as a classifier. }%
\label{2c2p3fc_mnistnetwork}%
\end{figure}

For comparison purposes with SVM, we also considered putting the L5 neurons
(i.e., vectors in $%
\mathbb{R}
^{500}$) into both a conventional two and three layer fully connected network
(FCN). Using a two layer FCN (see Figure \ref{2c2p2fc_mnistnetwork}) with
sigmoidal outputs, a cross-entropy cost function, and a learning rate
$\eta=0.1/(1.001)^{\#Epoch}$ we obtained 97.97 \% classification accuracy.
Similarly with a three layer FCN (see Figure \ref{2c2p3fc_mnistnetwork}) with
the same conditions an accuracy of 98.01 \% was obtained.%

%

\noindent\textbf{Separability of the MNIST\ Set}\ 

If $\lambda=1/1000$ then the 50,000 training and 10,000 validation images
converted to $%
\mathbb{R}
^{500}$ \textquotedblleft images\textquotedblright\ turn out to be completely
separable into the 10 digit classes! However, the accuracy on the remaining
10,000 test images drops to 97.01\%. The original 60,000 MNIST (training \&
validation) images in $R^{784}$ are not separable by a linear SVM (The SVM
code was run for 16 hours with $\lambda=1/1000$ without achieving separability).

\noindent\textbf{Increasing the Number of Output Maps}

If the number of maps in the L4 layer are increased to 1000 with the L5
$1\times1$ maps correspondingly increased to 1000, then there is a slight
increase in test accuracy as shown in Table 2. With $\lambda=1$ the 50,000
training and 10,000 validation images converted to $%
\mathbb{R}
^{1000}$ \textquotedblleft images\textquotedblright\ also turn out to be
completely separable into the 10 digit classes. However, with $\lambda=1$ the
test accuracy decreases to 97.61.

\begin{table}[th]%
\begin{tabular}
[c]{|c|c|c|c|c|c|c|}\hline
\textbf{Classifier} & \textbf{Test Acc } & \textbf{Valid Acc } &
\textbf{Training Time} & $\lambda$ & $\eta$ & \textbf{Epochs}\\\hline
RBF SVM & 98.01 \% & 98.20 \% & 8 minutes & 1/3.6 & - & -\\\hline
Linear SVM & 97.80 \% & 98.02 \% & 4 minutes & 1/0.012 & - & -\\\hline
2 Layer FCN (backprop) & 97.71 \% & 98.74 \% & 15 minutes & 1.0 & $\frac
{0.1}{(1.007)^{\#Epoch}}$ & 30\\\hline
3 layer FCN (backprop) & 98.01 \% & 98.10 \% & 50 minutes & 6.0 & $\frac
{0.1}{(1.007)^{\#Epoch}}$ & 30\\\hline
\end{tabular}
\caption{Classification accuracies on MNIST data set with various classifiers
when number of maps in L4 is 1000.}%
\end{table}

\subsection{Classification with a Single Convolution/Pool Layer
\label{Section3.2}}

\begin{figure}[H]%
\centering
\includegraphics[
height=2.5573in,
width=5.9352in
]%
{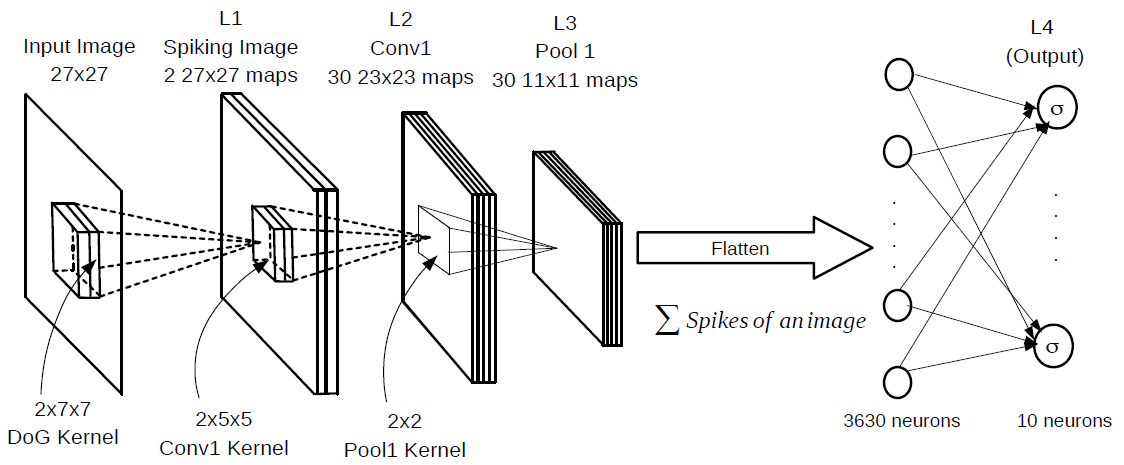}%
\caption{Deep spiking convolutional network architecture for classification of
the MNIST\ data set.}%
\label{1c1p1fcn_mnistnetwork2}%
\end{figure}

The architecture shown in Figure \ref{1c1p1fcn_mnistnetwork2} has a single
convolutional/pooling layer with $30\times11\times11=3630$ pooled neurons in
L3. These neurons are fully connected to L4 layer of $3630$ neurons. However,
the neurons in L4 are in 1-1 correspondence with the L3 neurons (flatten).
Further, each neuron in L4 simply sums the spikes coming into it from its
corresponding neuron in L3. The L4 neurons are fully connected (with trainable
weights) to 10 output neurons. This final layer of weights are then trained
using backprop only on this output layer, i.e., only backprop to L4. (See Lee
at al. \cite{Panda} where the error is back propagated through all the layers
and reported an accuracy of 99.3\%). Inhibition settings are same as in the
above experiment.%
The first row of Table 3 shows a 98.4\% test accuracy using back propagation
on the output layer (2 Layer FCN). The second and third rows give the
classification accuracy using an SVM trained on the L4 neurons (their spike
counts). The feature extraction that takes place in the L2 layer (and passed
through the pooling layer) results in greater than 98\% accuracy with a two
layer conventional FCNN output classifier. A conventional FC two layer NN
(i.e., no hidden layer) with the $28\times28$ images of the MNIST data set as
input has only been reported to achieve 88\% accuracy and 91.6\% with
preprocessed data \cite{Lecun98}. This result strengthens our view that the
unsupervised STDP\ appears to convert the MNIST classes into classes in a
higher space that are separable.

We also counted the spikes in network with two convolution/pool layers (see
Figure \ref{mnistnetwork_maxpots2}) but found that the accuracy decreased (see
Table 2) This decrease may be due to that reduced number of spikes in the
output neurons compared to have only one convolution/pool layer. 

\begin{table}[th]%
\begin{tabular}
[c]{|c|c|c|c|c|c|c|}\hline
\textbf{Classifier} & \textbf{Test Acc} & \textbf{Valid Acc} &
\textbf{Training Time} & $\mathbf{\lambda}$ & $\eta$ & \textbf{Epochs}\\\hline
\multicolumn{1}{|l|}{2 Layer FCN} & 98.4\% & 98.5\% \  & 10mins & $1/10$ &
$0.1/(1.007)^{\#Epoch}$ & 20\\\hline
\multicolumn{1}{|l|}{RBF SVM} & 98.8\% & 98.87\% & 150 minutes & $1/3.6$ & - &
-\\\hline
\multicolumn{1}{|l|}{Linear SVM} & 98.41\% & 98.31\% & 100 minutes & $1/0.012
$ & - & -\\\hline
\end{tabular}
\caption{Classification accuracies on MNIST data set with various classifiers
when a single convolution/pool layer is used.}%
\end{table}

\section{Reward Modulated STDP}

\begin{figure}[H]%
\centering
\includegraphics[
height=2.5495in,
width=6.4048in
]%
{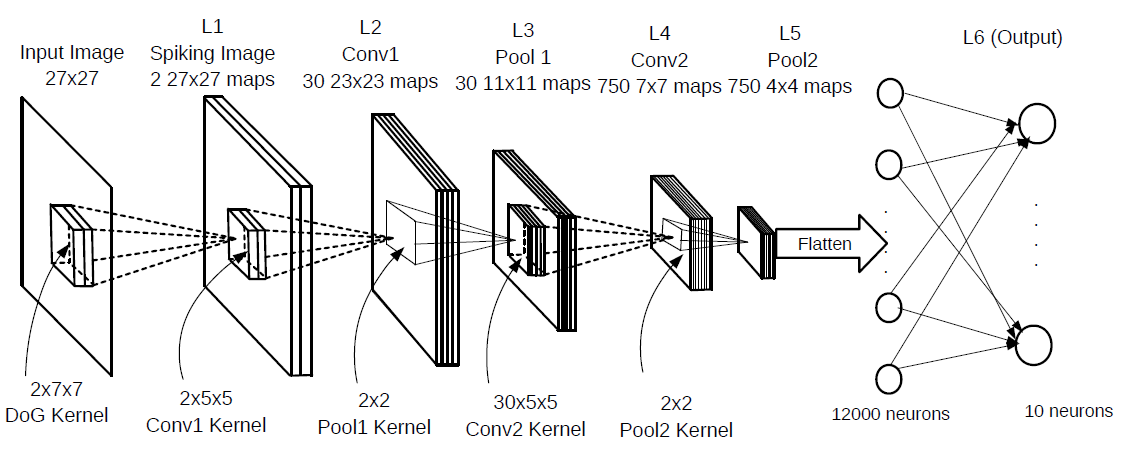}%
\caption{Network with 750 maps in L4.}%
\label{2c2p1fc_mnistnetwork_rstdp}%
\end{figure}

Reward modulated STDP is a way to use the accumulated spikes at the output to
do the final classification (in contrast to SVM and a two layer backprop
mentioned above). Figure \ref{2c2p1fc_mnistnetwork_rstdp} shows the network
architecture where the reward modulated STDP is carried out between the
(flattened) L5 layer and the ten output neurons of the L6 layer. The weights
between the fully connected neurons of Layer 5 and Layer 6 are then trained as
follows: For any input image the spikes through the network arrive between
$t=0$ and $t=11$ time steps. The final ($t=11$) membrane potential of the
$k^{th}$ output neuron for $k=1,2,...,10$ is then%
\[
V_{k}=\sum_{t=0}^{11}\sum_{j=1}^{12000}w_{kj}s_{L5}(t,j).
\]
Denote by $N_{hit}$ and $N_{miss}$ the number of correctly classified and
incorrectly classified images for every $N$ (e.g., $N=100,500,1500,$ etc.)
input images so $N_{miss}+N_{hit}=N$. If the $k^{th}$ output potential $V_{k}
$ is maximum (i.e., $V_{k}>V_{j}$ for $j\neq k$) and the input image has label
$k$ then the weights going into the $k^{th}$ output neuron are rewarded in the
sense that
\begin{equation}
w_{kj}\longleftarrow w_{kj}+\Delta w_{kj},\text{ \ where \ }\Delta w_{kj}=%
\begin{cases}
+\dfrac{N_{miss}}{N}a_{r}^{+}w_{kj}(1-w_{kj})\text{ \ if at least one
pre-synaptic spike from j to k.}\\
-\dfrac{N_{miss}}{N}a_{r}^{-}w_{kj}(1-w_{kj})\ \ \text{otherwise.}%
\end{cases}
\label{eq11}%
\end{equation}
If $V_{k}$ is the maximum potential, but the label of the image is $j\neq k,$
then the weights going into output neuron $k$ are punished in the sense that%
\begin{equation}
w_{kj}\longleftarrow w_{kj}+\Delta w_{kj},\text{ \ where \ }\Delta w_{kj}=%
\begin{cases}
-\dfrac{N_{hit}}{N}a_{p}^{+}w_{kj}(1-w_{kj})\text{ \ if at least one
pre-synaptic spike from j to k.}\\
+\dfrac{N_{hit}}{N}a_{p}^{-}w_{kj}(1-w_{kj})\ \ \text{otherwise.}%
\end{cases}
\label{eq12}%
\end{equation}
Note that only the weights of those neurons connected to the output neuron
with the maximum potential are updated. The term \textquotedblleft
modulated\textquotedblright\ in reward modulated STDP refers to the factors
$\dfrac{N_{miss}}{N}$ and $\dfrac{N_{hit}}{N}$ which multiply (modulate) the
learning rule. Equation (\ref{eq11}) refers to the case where the k$^{th}$
output neuron also has the high membrane potential of the ten outputs. If
$N_{miss}/N$ is small then the network accuracy is performing well in terms of
accuracy and the change is weights is small (as the weights are thought to
already have learned to correctly classify). On the other hand, equation
(\ref{eq12}) refers to the case where the k$^{th}$ output has the highest
membrane potential, but the label is $j\neq k.$ Then, if $N_{miss}/N$ is
small, it follows that $N_{hit}/N$ is large the weights of the neurons going
into the k$^{th}$ neuron have their values changed by a relatively large
amount to (hopefully) correct the misclassification.
In this experiment with R-STDP, only 20,000 MNIST digits were used for
training, 10,000 digits for validation (used to choose the number of training
epochs), and the 40,000 remaining digits were used for testing. The R-STDP
synaptic weights between L5 and L6 were initialized from the normal
distribution $\mathcal{N}(0.8,0.01)$. Table 4 shows that a test accuracy of
only 90.1\% was obtained.

\begin{table}[th]
\centering
\begin{tabular}
[c]{|c|c|c|c|}\hline
M\textbf{aps in L4} & \textbf{Valid acc \%} & \textbf{Test Acc \%} &
\textbf{Epochs}\\\hline
750 & 91.2 & 90.1 & 150\\\hline
\end{tabular}
\caption{Classification accuracy on MNIST data set with R-STDP when one neuron
per class is used.}%
\end{table}

For comparison, we replaced the R-STDP classifier\ (from L5 to L6) with a
simple 2 layer neural network (from L5 to L6) which used error back
propagation. These weights for back propagation were initialized from the
normal distribution $\mathcal{N}(0,1/\sqrt{12000})$ as in \cite{Nielsen}.
Table 6 shows that R-STDP performed poorly compared to the simple two layer
backprop which ran for only 20 epochs.

\begin{table}[th]
\centering
\begin{tabular}
[c]{|c|c|c|c|c|c|}\hline
\textbf{Classifier} & \textbf{Test Acc} & \textbf{Valid Acc} &
$\mathbf{\lambda}$ & $\eta$ & \textbf{Epochs}\\\hline
2 Layer FCN & 97.5\% & 97.6\% & $1.0$ & $0.1/(1.007)^{\#Epoch}$ & 20\\\hline
\end{tabular}
\caption{Classification accuracy on MNIST data set with single layer
backprop.}%
\end{table}

Mozafari et al. \cite{mozafari1}\cite{mozafari2} got around this poor
performance by having 250 neurons in the output layer and assigning 25 output
neurons per class. They reported 97.2 \% test accuracy while training on
60,000 images and testing on 10,000 images. We also considered multiple
neurons per class in the output layer. As Table 6 shows, we considered 300
output neurons (30 per class) and we also consider dropout. $P_{drop}=0.4$
means that $0.4(300)=120$ output neurons were prevented from updating their
weights for the particular training image. For each input image a different
set of 120 randomly neurons were chosen to not have their weights updated.
Table 6 shows that the best performance of 95.91 \% test accuracy was obtained
with $P_{drop}=0.4.$

\begin{table}[th]
\centering
\begin{tabular}
[c]{|c|c|c|c|c|c|}\hline
\textbf{Maps in L4} & \textbf{\#Output Neurons} & \textbf{P}$_{drop}$ &
\textbf{Valid acc \%} & \textbf{Test acc \%} & \textbf{Epochs}\\\hline
750 & 300 & 0.3 & 95.81 & 95.84 & 400\\\hline
750 & 300 & 0.4 & 96.01 & 95.91 & 400\\\hline
750 & 300 & 0.5 & 95.76 & 95.63 & 400\\\hline
\end{tabular}
\caption{Classification accuracy on MNIST data set with R-STDP when more than
one neuron per class is used.}%
\end{table}

\subsection{\noindent\textbf{R-STDP as a Classification Criteria}}

We experimented with R-STDP learning rule applied to L5-L6 synapses of the
network in the Figure \ref{2c2p1fc_mnistnetwork_rstdp} by two different kinds
of weight initialization and also varying initialization of parameters like
$\dfrac{N_{miss}}{N},$ $\dfrac{N_{hit}}{N}$ and $N$.

\subsubsection{Backprop Initialized Weights for R-STDP}

We were concerned with the poor performance using an R-STDP as a classifier as
given in Table 6. In particular, perhaps the weight initialization plays a
role in that the R-STDP rule can get stuck in a local minimum. To study this
in more detail the network in Figure \ref{2c2p1fc_mnistnetwork_rstdp} was
initialized with a set of weight that are known to give a high accuracy. To
explain, the final weights used in the 2 Layer FCN reported in Table 5 were
used as a starting point. As these weights are both positive and negative,
they were shifted to be all positive. This was done by first finding the
minimum value $w_{\min}$ $(<0)$ of these weights and simply adding $-w_{\min
}>0$ to them so that they are all positive. Then this new set of weights were
re-scaled to be between 0 and 1 by dividing them all by their maximum value
(positive). These shifted and scaled weights were then used to initialize the
weights of the R-STDP classifier. The parameters $a_{r}^{+},a_{r}^{-}%
,a_{p}^{+},a_{p}^{-}$ were initialized to be 0.004, 0.003, 0.0005, 0.004
respectively. With the network in Figure \ref{2c2p1fc_mnistnetwork_rstdp}
initialized by these weights, the validation images were fed through the
network and the neuron number with the maximum potential is the predicted
output. The validation accuracy was found to be 97.1\%.

With weights of the fully connected layer of Figure
\ref{2c2p1fc_mnistnetwork_rstdp} initialized as just described, the R-STDP
rule was used to train the network further for various number of epochs and
two different ways of updating $\dfrac{N_{miss}}{N}$ and $\dfrac{N_{hit}}{N}.$

\paragraph{Batch Update of $\dfrac{N_{miss}}{N}$ and $\dfrac{N_{hit}}{N}$}

The first set of experiments were done with the $\dfrac{N_{miss}}{N}$ and
$\dfrac{N_{hit}}{N}$ ratios updated after every \emph{batch} of $N$ images for
$N=100,500,1500,2500.$ As the weights of the fully connected layer of Figure
\ref{2c2p1fc_mnistnetwork_rstdp} with the backprop trained values, we expect
$\dfrac{N_{miss}}{N}$ to be a low fraction or equivalently $\dfrac{N_{hit}}%
{N}$ to be high. Consequently, they were initialized as $\dfrac{N_{miss}}%
{N}=0.1,$ $\dfrac{N_{hit}}{N}=0.9.$ With these initialization, Table 7 shows
that accuracy on the validation set did not decrease significantly for $N$ not
too large. However, using larger values of $N$ (value of N depends on the
initialization of $N_{miss}/N$ and $N_{hit}/N$) the accuracy goes down
significantly. For example, for the cases where $N_{miss}/N=0.035$ and
$N_{hit}/N=0.965$ the accuracy didn't significantly decrease until the batch
size was $N=3500.$ In the case with $N_{miss}/N=0.0$ and $N_{hit}/N=1.0$ the
accuracy didn't decrease at all. This is because the best performing weights
for validation accuracy were used, but these same weights also gave 100\%
accuracy on the training data.

\begin{table}[th]
\centering
\begin{tabular}
[c]{|c|c|c|c|c|}\hline
$\dfrac{N_{miss}}{N}$ & $\dfrac{N_{hit}}{N}$ & $N$ & Acc. at start & Acc. at
end\\\hline
0.1 & 0.9 & 100 & 97.1\% & 96.91\%\\\hline
0.1 & 0.9 & 500 & 97.1\% & 96.96\%\\\hline
0.1 & 0.9 & 1500 & 97.1\% & 96.82\%\\\hline
0.1 & 0.9 & 2500 & 97.1\% & 90.76\%\\\hline
0.035 & 0.965 & 2500 & 97.1\% & 96.69\%\\\hline
0.035 & 0.965 & 3000 & 97.1\% & 96.58\%\\\hline
0.035 & 0.965 & 3500 & 97.1\% & 91.05\%\\\hline
0.035 & 0.965 & 4000 & 97.1\% & 90.98\%\\\hline
0.0 & 1.0 & 100 & 97.1\% & 96.93\%\\\hline
0.0 & 1.0 & 500 & 97.1\% & 96.93\%\\\hline
0.0 & 1.0 & 1500 & 97.1\% & 96.94\%\\\hline
0.0 & 1.0 & 2500 & 97.1\% & 96.94\%\\\hline
0.0 & 1.0 & 3000 & 97.1\% & 96.94\%\\\hline
0.0 & 1.0 & 3500 & 97.1\% & 96.94\%\\\hline
0.0 & 1.0 & 4000 & 97.1\% & 96.93\%\\\hline
\end{tabular}
\caption{Demonstration of sensitivity of R-STDP to N value with correct
initialization of hit and miss ratios.}%
\end{table}

Table 8 shows the classification accuracy with "poor" initialization
$N_{miss}/N=0.9$ and $N_{hit}/N=0.1.$ If the weights had been randomly
initialized then the initialization $N_{miss}/N=0.9$ and $N_{hit}/N=0.1$ would
be appropriate. However, Table 8 shows that R-STDP isn't able to recover from
this poor initialization.

\begin{table}[th]
\centering
\begin{tabular}
[c]{|c|c|c|c|c|}\hline
$\dfrac{N_{miss}}{N}$ & $\dfrac{N_{hit}}{N}$ & $N$ & Acc. at start & Acc. at
end\\\hline
0.9 & 0.1 & 100 & 97.1\% & 91.52\%\\\hline
0.9 & 0.1 & 500 & 97.1\% & 90.67\%\\\hline
0.9 & 0.1 & 1500 & 97.1\% & 90.47\%\\\hline
0.9 & 0.1 & 2500 & 97.1\% & 90.45\%\\\hline
\end{tabular}
\caption{Demonstration of sensitivity of R-STDP to N value with incorrect
initialization of hit and miss ratios.}%
\end{table}

\newpage

\paragraph{Update of $\dfrac{N_{miss}}{N}$ and $\dfrac{N_{hit}}{N}$ after each
image}

Next, $N_{miss}/N$ and $N_{hit}/N$ were updated after every image using the
most recent $N$ images. Even with $N_{miss}/N$ and $N_{hit}/N$ initialized
incorrectly, the validation accuracies in Table 9 did not decrease
significantly. Though the accuracy still goes down slightly, the table
indicates that updating $N_{miss}/N$ and $N_{hit}/N$ after every image
mitigates this problem. 

\begin{table}[th]
\centering
\begin{tabular}
[c]{|c|c|c|c|c|}\hline
$\dfrac{N_{miss}}{N}$ & $\dfrac{N_{hit}}{N}$ & $N$ & Acc. at start & Acc. at
end\\\hline
0.9 & 0.1 & 100 & 97.1\% & 96.93\%\\\hline
0.9 & 0.1 & 500 & 97.1\% & 96.94\%\\\hline
0.9 & 0.1 & 1500 & 97.1\% & 96.93\%\\\hline
0.9 & 0.1 & 2500 & 97.1\% & 96.94\%\\\hline
\end{tabular}
\caption{Demonstration of sensitivity of R-STDP.}%
\end{table}

Still updating $N_{miss}/N$ and $N_{hit}/N$ after each image, it was found
that R-STDP accuracy was very sensitive to the initialized weights.
Specifically the L5-L6 R-STDP weights were initialized using the backprop
trained weights (as explained above) by doing the backprop for just 10 epochs
(instead of 20) and $\lambda=10.0$ (regularization parameter) which gave
99.6\% training and 96.8\% validation accuracies. Table 10 gives the
validation accuracies using R-STDP for 100 epochs. Surprisingly, even with a
good initialization of the weights and the ratios $N_{miss}/N$ and $N_{hit}/N
$, the validation accuracy suffers. 

\begin{table}[th]
\centering
\begin{tabular}
[c]{|c|c|c|c|c|}\hline
$\dfrac{N_{miss}}{N}$ & $\dfrac{N_{hit}}{N}$ & $N$ & Acc. at start & Acc. at
end\\\hline
0.0 & 1.0 & 100 & 96.8\% & 90.75\%\\\hline
0.0 & 1.0 & 4000 & 96.8\% & 90.67\%\\\hline
\end{tabular}
\caption{Demonstration of sensitivity of R-STDP for weight initialization.}%
\end{table}

\begin{figure}[H]%
\centering
\includegraphics[
height=2.9058in,
width=5.4457in
]%
{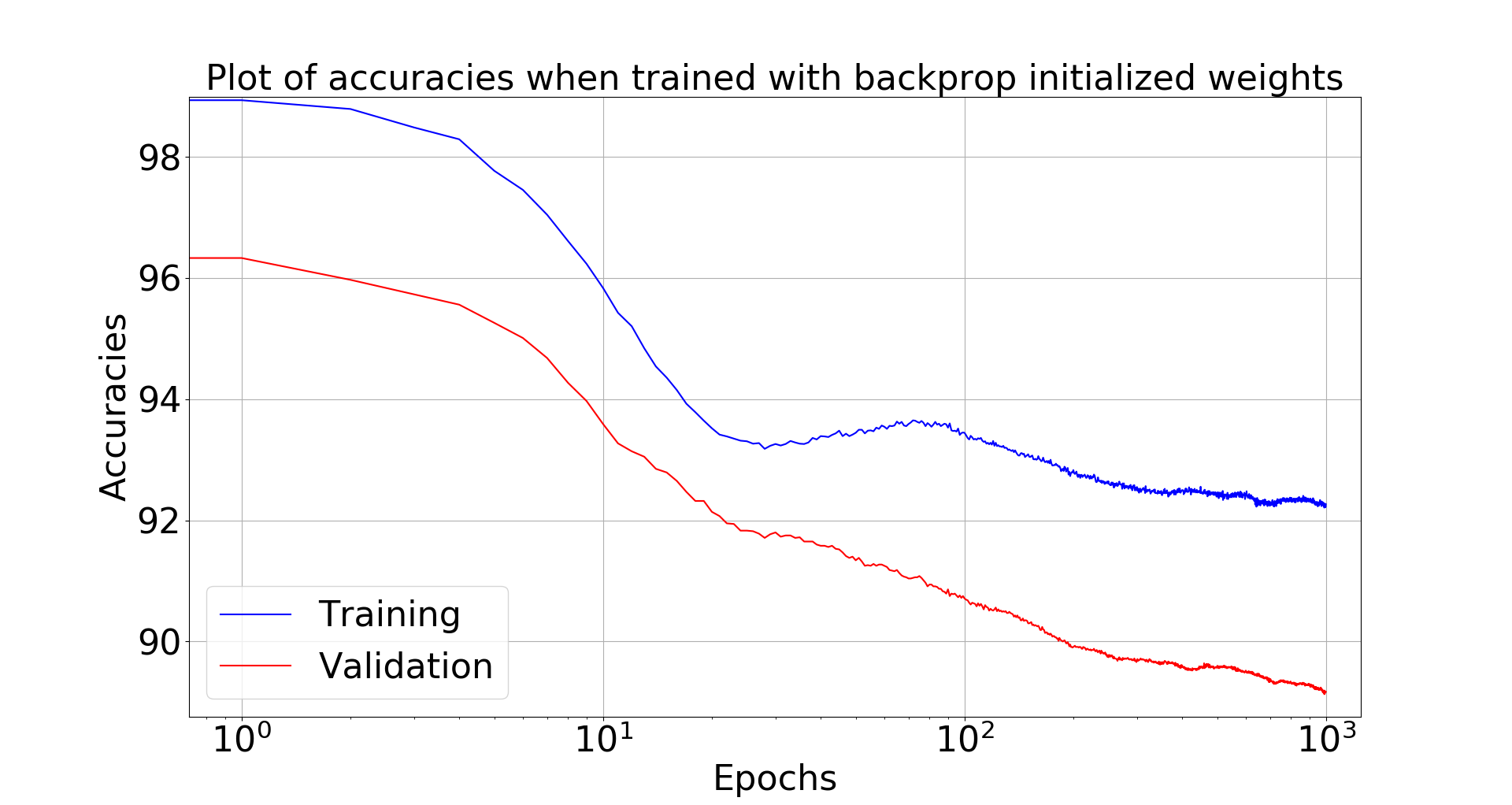}%
\caption{Plot of accuracies versus epochs when the weights were initialized
with backprop trained weights.}%
\label{acc_bacprop_weights_3}%
\end{figure}

For this same cases as Table 10, the R-STDP algorithm was run for 1000 epochs
with the training and validation accuracies versus epoch plotted in Figure
\ref{acc_bacprop_weights_3}. Notice that the validation accuracy drops to
\symbol{126}90\%. It seems that R-STDP is not a valid cost function as far as
accuracy is concerned\footnote{At least using one output neuron per class.}.
Interestingly, as shown next, training with R-STDP with randomly initialized
weights, the validation accuracy only goes up to \symbol{126}90\% (see Figure
\ref{acc_rndm_weights_3}).
\bigskip

\subsubsection{Randomly Initialized Weights for R-STDP}

In the set of experiments, the weights trained with R-STDP were
\emph{randomly} initialized from the normal distribution $\mathcal{N}%
(0.8,0.01)$ and the $N_{miss}/N,N_{hit}/N,N$ parameters initialized with the
values given in Table 13. Validation accuracies are shown at the end of 100
epochs $N_{miss}/N$ and $N_{hit}/N,N$ were updated after every
image.

\begin{table}[th]
\centering
\begin{tabular}
[c]{|c|c|c|c|c|}\hline
$\dfrac{N_{miss}}{N}$ & $\dfrac{N_{hit}}{N}$ & $N$ & Acc. at start & Acc. at
end\\\hline
0.9 & 0.1 & 100 & 10.3 & 90.22\\\hline
0.9 & 0.1 & 500 & 10.1 & 90.13\\\hline
0.9 & 0.1 & 1500 & 10.2 & 90.12\\\hline
0.9 & 0.1 & 2500 & 10.6 & 90.16\\\hline
\end{tabular}
\caption{Demonstration of sensitivity of R-STDP.}%
\end{table}

\begin{center}%
\begin{figure}[H]%
\centering
\includegraphics[
height=2.8106in,
width=5.2676in
]%
{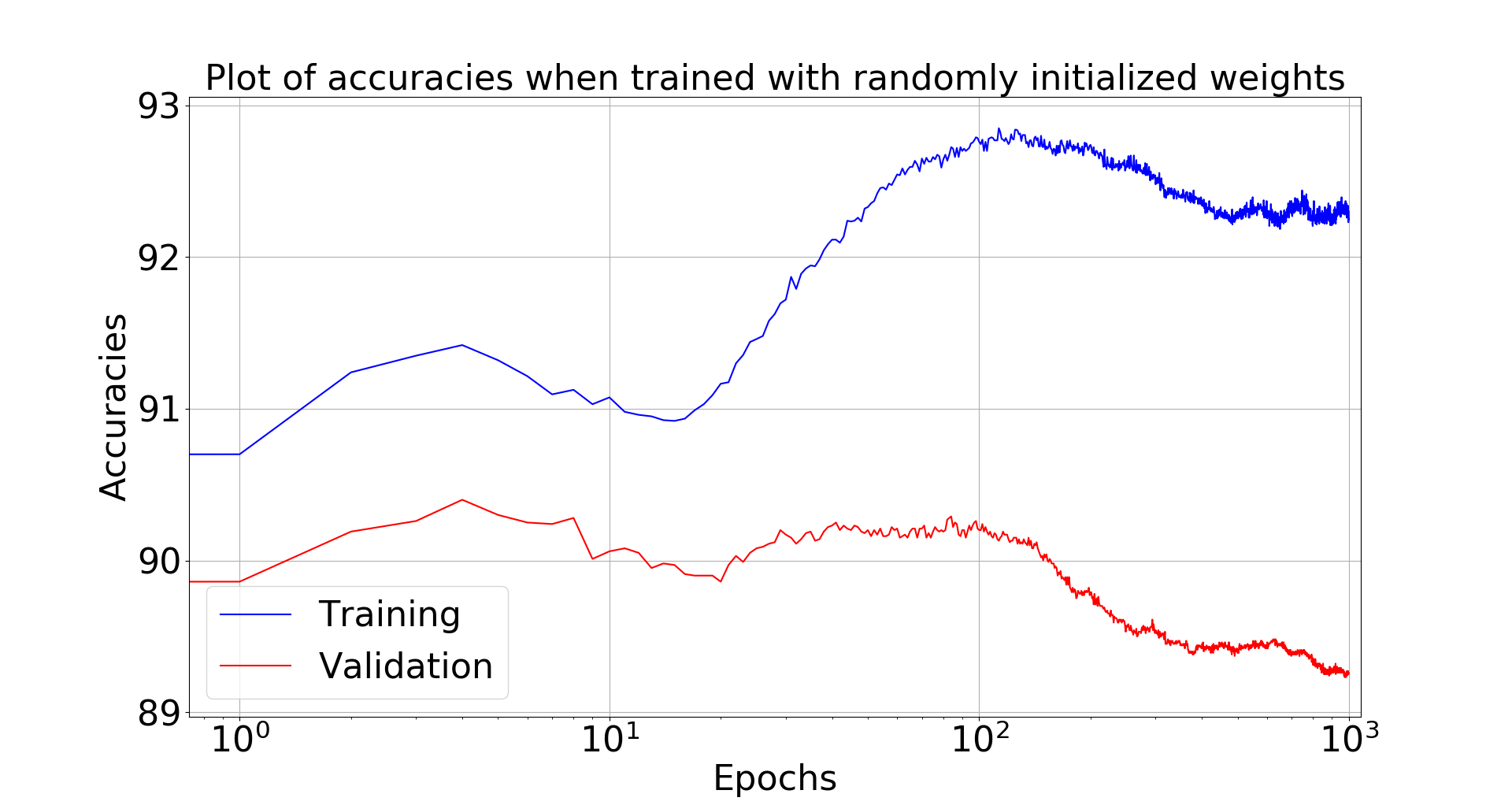}%
\caption{Plot of accuracies versus epochs when the weights were randomly
initialized.}%
\label{acc_rndm_weights_3}%
\end{figure}

\end{center}

For this same cases as Table 11, the R-STDP algorithm was run for 1000 epochs
with the training and validation accuracies versus epoch plotted in Figure
\ref{acc_rndm_weights_3}. The validation accuracy only goes up to \symbol{126}90\%.

\section{Classification of N-MNIST data set}%

\begin{figure}[H]%
\centering
\includegraphics[
height=3.179in,
width=6.3477in
]%
{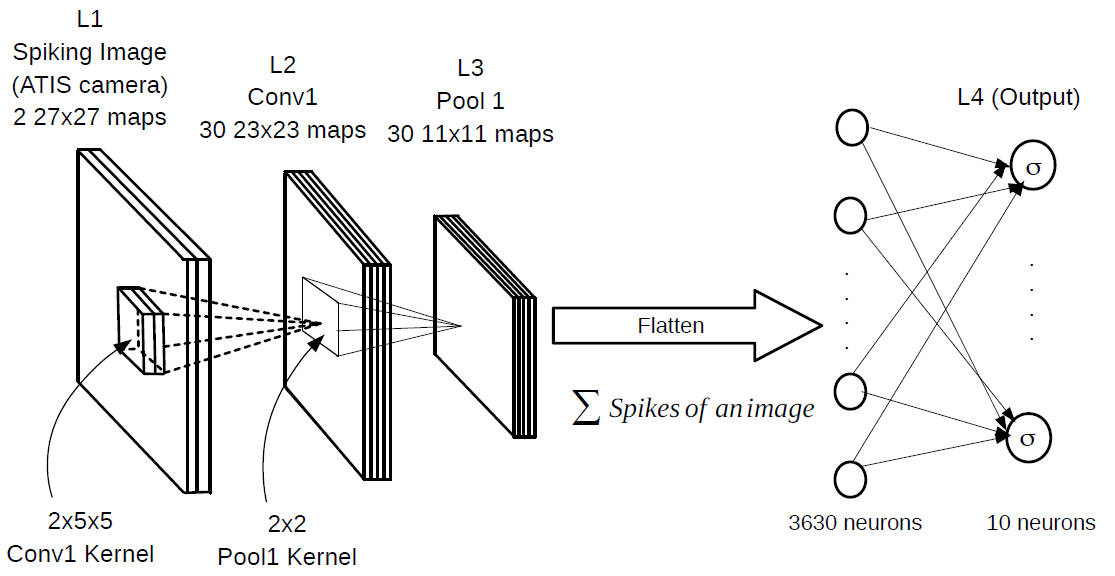}%
\caption{Network for N-MNIST classification.}%
\label{1c1p1fc_atismnistnetwork2}%
\end{figure}

\begin{figure}[H]%
\centering
\includegraphics[
height=1.8991in,
width=6.0044in
]%
{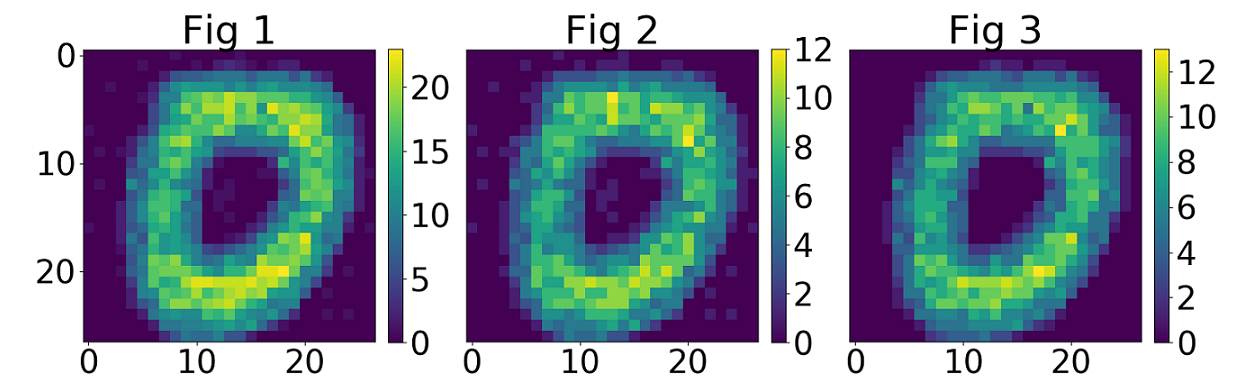}%
\caption{Left: Accumulated ON and OFF center spikes. Center: Accumulate ON
center spikes. Right: Accumulated OFF center spikes.}%
\label{untrackedinput}%
\end{figure}

In the above we artificially constructed spiking images using a DoG filter on
the standard MNIST\ data set as in \cite{Kheradpisheh_2016}%
\cite{Kheradpisheh_2016b}. However the ATIS (silicon retina) camera
\cite{atis} works by producing spikes. We also considered classification
directly on recorded output from the ATIS\ camera given in the N-MNIST\ data
set \cite{nmnist}. A silicon retinal detects change in pixel intensity and
thus the MNIST digits are recorded with camera moving slightly (saccades).
Figure \ref{untrackedinput} shows the raw accumulated spikes of the N-MNIST
data set as given in \cite{nmnist}.

\begin{figure}[H]%
\centering
\includegraphics[
height=1.8732in,
width=5.9629in
]%
{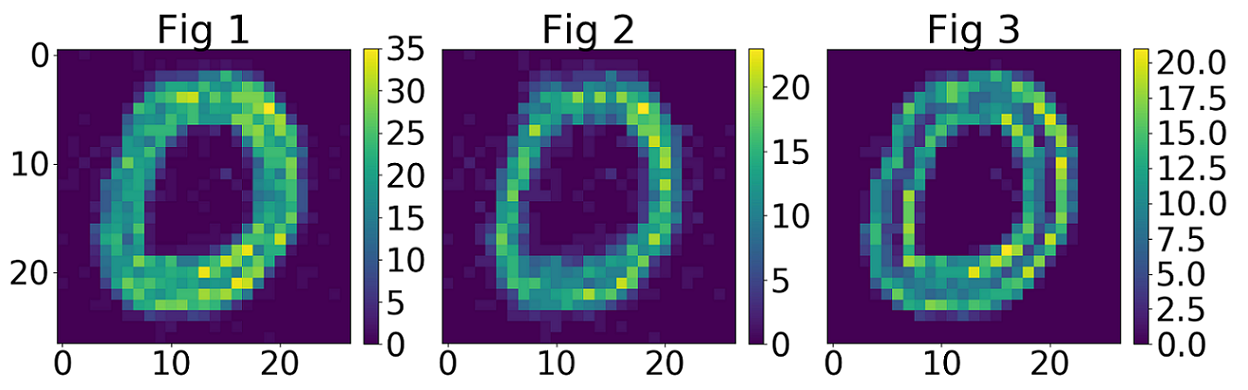}%
\caption{Left: Accumulated ON and OFF center spikes. Center: Accumulate ON
center spikes. Right: Accumulated OFF center spikes.}%
\label{trackedinput}%
\end{figure}

Figure \ref{trackedinput} is the same as Figure \ref{untrackedinput}, but
corrected for saccades (camera motion) using the algorithm given in
\cite{nmnist}.%

Figure \ref{1c1p1fc_atismnistnetwork2} shows the network we used for
classification of the N-MNIST data. We first hard wired the weights $W_{C1}$
of the convolution kernel from L1 to L2 of Figure
\ref{1c1p1fc_atismnistnetwork2} to the values already trained above in
subsection \ref{Section3.2} (see Figure \ref{1c1p1fcn_mnistnetwork2}). Only
the weights from L4 to L5 were trained for classification by simply back
propagating the errors from L5 to L4. This result in given in the first row of
Table 12. We also trained an SVM on the L4 neuron outputs with the results
given in row 2 (RBF) and row 3 (linear) of Table 12. All the results in Table
12 were done on the raw spiking inputs from \cite{nmnist} (i.e., not corrected
for saccade) with training done on 50,000 (spiking) images, validation \&
testing done on 10,000 images each.

\begin{table}[th]
\centering
\begin{tabular}
[c]{|c|c|c|c|c|c|c|}\hline
\textbf{Classifier} & \textbf{Test Acc} & \textbf{Valid Acc} &
\textbf{Training Time} & $\mathbf{\lambda}$ & $\eta$ & \textbf{Epochs}\\\hline
2 Layer FCN & 97.45\% & 97.62\% & 5 minutes & $\frac{1}{10.0}$ & $\frac
{0.1}{1.007^{\#Epoch}}$ & 20\\\hline
RBF SVM & 98.32\% & 98.40\% & 200 minutes & $\frac{1}{3.6}$ & - & -\\\hline
Linear SVM & 97.64\% & 97.71\% & 100 minutes & $\frac{1}{0.012}$ & - &
-\\\hline
\end{tabular}
\caption{Classification accuracies of N-MNIST data set with one
convolution/pool layers for transfer learning.}%
\end{table}

In Table 13 we show the results for the case where the weights $W_{C1}$ of the
convolution kernel from L1 to L2 of Figure \ref{1c1p1fc_atismnistnetwork2}
were trained (unsupervised) using the N-MNIST data set. In this instance we
used N-MNIST\ data corrected for saccades since this gave better result than
the uncorrected data. All the results in Table 8 were produced by training on
50,000 (spiking) images with validation \& testing done using 10,000 images
each. 

\begin{table}[th]
\centering
\begin{tabular}
[c]{|c|c|c|c|c|c|c|}\hline
\textbf{Classifier} & \textbf{Test Acc} & \textbf{Valid Acc} &
\textbf{Training Time} & $\mathbf{\lambda}$ & $\eta$ & \textbf{Epochs}\\\hline
1 Layer FCN & 97.21\% & 97.46\% & 5 minutes & $\frac{1}{10.0}$ & $\frac
{0.1}{1.007^{\#Epoch}}$ & 20\\\hline
RBF SVM & 98.16\% & 98.2\% & 150 minutes & $\frac{1}{3.6}$ & - & -\\\hline
Linear SVM & 97.38\% & 97.44\% & 100 minutes & $\frac{1}{0.012}$ & - &
-\\\hline
\end{tabular}
\caption{Classification accuracies of N-MNIST data set with one
convolution/pool layers when trained with N-MNIST spikes.}%
\end{table}

We also added an extra convolution layer, but found that the classification
accuracy decreased. Jin et al reported an accuracy of 98.84\% by using a
modification of error back propagation (all layers) algorithm \cite{jin}.
Stromatias et al reported an accuracy of 97.23\% accuracy by using
artificially generated features for the kernels of the first convolutional
layer and training a 3 layer fully connected neural network classifier on
spikes collected at the first pooling layer \cite{stromatias}.

\section{Catastrophic Forgetting}

\begin{figure}[H]%
\centering
\includegraphics[
height=2.9162in,
width=5.6325in
]%
{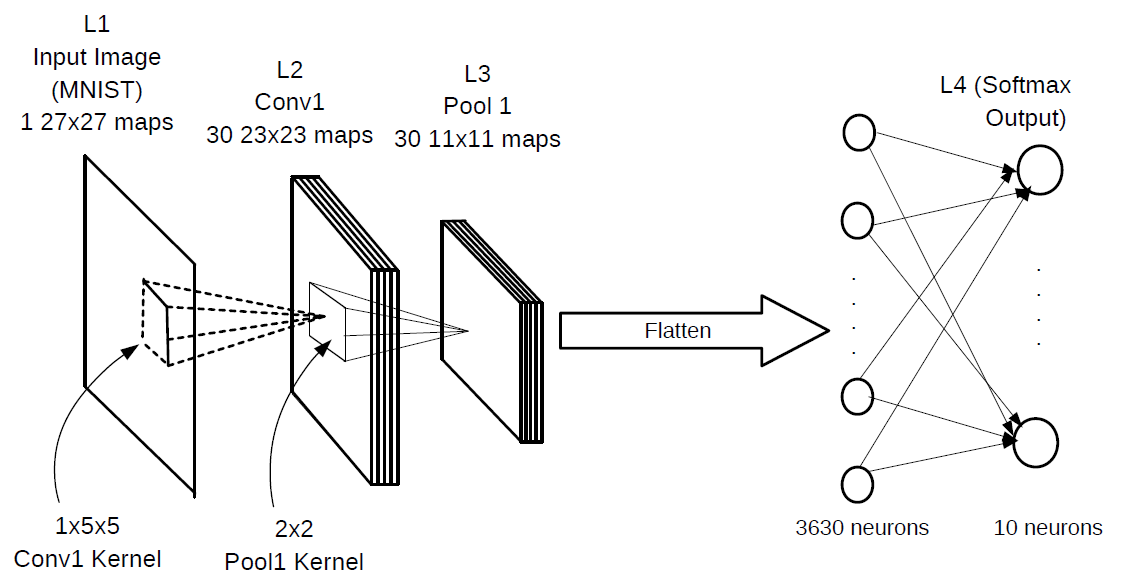}%
\caption{Network architecture for catastrophic forgetting.}%
\label{1c1p1fc_dcnnmnistnetwork}%
\end{figure}

Catastrophic forgetting is a problematic issue in deep convolutional neural
networks. In the context of the MNIST data set this refers to training the
network to learn the digits 0,1,2,3,4 and, after this is done, training on the
digits 5,6,7,8,9 is carried on. The catastrophic part refers to the problem
that the network is no longer able to classify the first set of digits
0,1,2,3,4. In more detail, Figure \ref{1c1p1fc_dcnnmnistnetwork} shows a
conventional (non-spiking) neural network with one convolution layer \& one
pool layer followed by a fully connected softmax output.%

\begin{figure}[H]%
\centering
\includegraphics[
height=3.5in,
width=7.25in
]%
{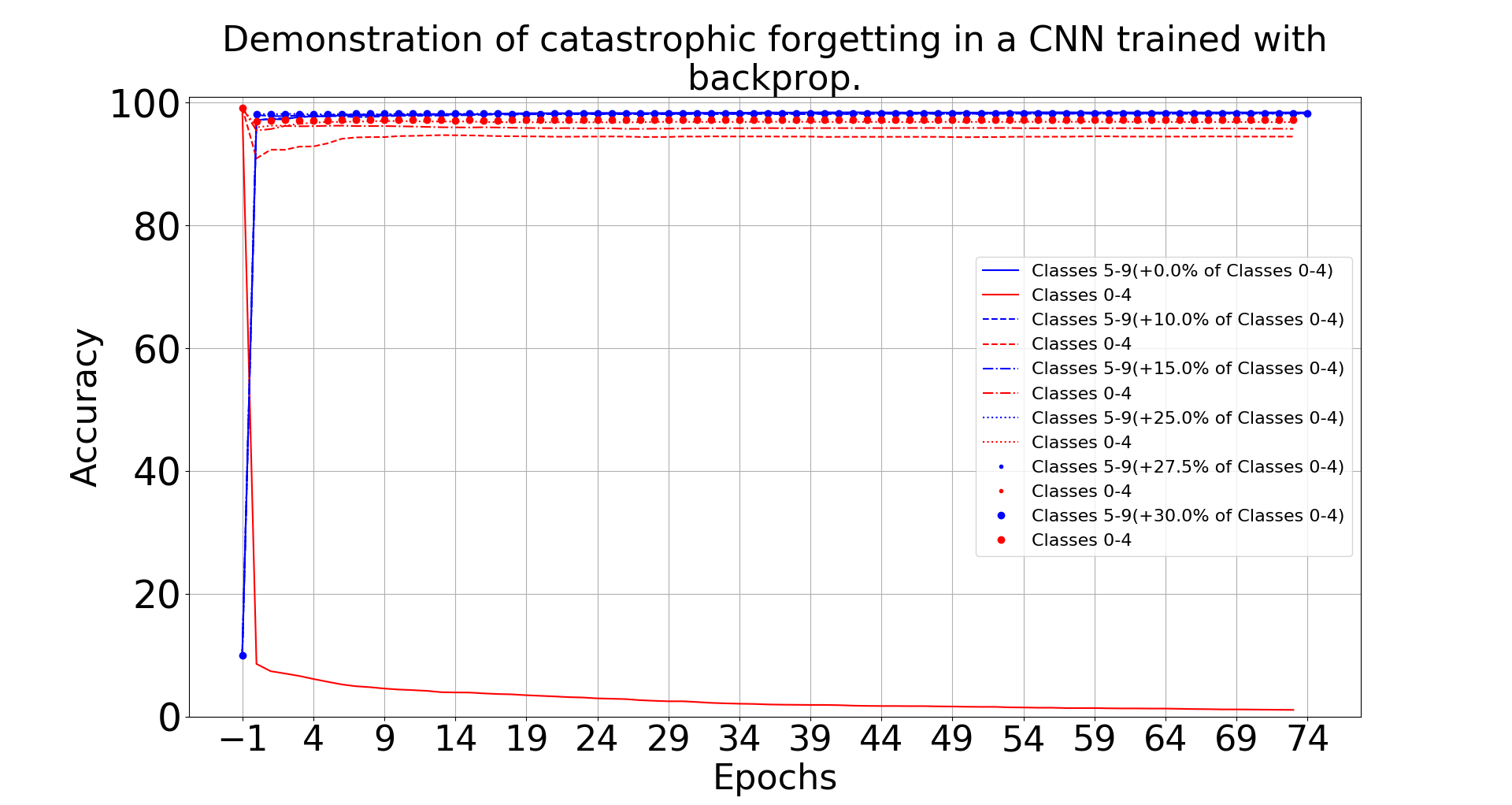}%
\caption{Catastrophic forgetting in a convolutional network while revising a
fraction of the previously trained classes. Note that epoch -1 indicates that
the network was tested for validation accuracy before training of the classes
5-9 started. Brackets in the legend shows the fraction of previously trained
classes that were used to revise the weights from the previous classes.}%
\label{cnn_cata_revision_1c1p1fc_zoom_v2}%
\end{figure}

\begin{figure}[H]%
\centering
\includegraphics[
height=2.7449in,
width=5.1422in
]%
{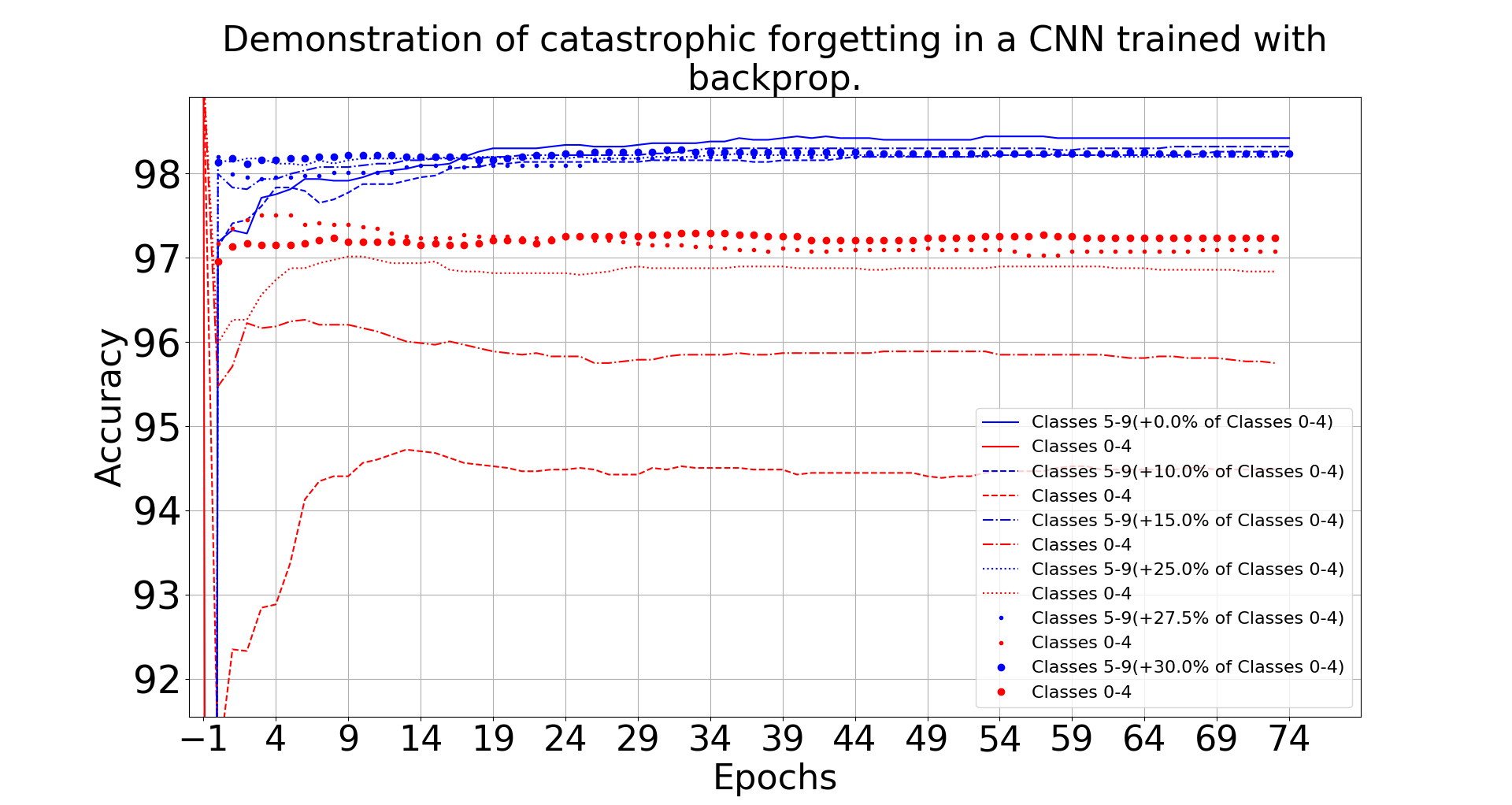}%
\caption{Zoomed upper portion of the Figure
\ref{cnn_cata_revision_1c1p1fc_zoom_v2}}%
\label{cnn_cata_revision_1c1p1fc_v2}%
\end{figure}

This network has 10 outputs but was first trained only on the digits 0,1,2,3,4
back propagating the error (computed from all 10 outputs) to the input
(convolution) layer. This training used approximately 2000 digits per class
and was done for 75 epochs. Before training the network on the digits
5,6,7,8,9 we initialized the weights and biases of the convolution and fully
connected layer with the saved weights of the previous training. For the
training with the digits 5,6,7,8,9 we \emph{fixed} the weights and biases of
the convolution layer with their initial values. The network was then trained,
but only the weights of fully connected layer were updated. (I.e., the error
was only back propagated from the 10 output neurons to the previous layer
(flattened pooled neurons). This training also used approximately 2000 digits
per class and was done for 75 epochs. While the network was being trained on
the second set of digits, we computed the validation accuracy on all 10 digits
at the end of each epochs. We plotted these accuracies in Figure
\ref{cnn_cata_revision_1c1p1fc_zoom_v2}. The solid red line in Figure
\ref{cnn_cata_revision_1c1p1fc_zoom_v2} are the accuracies versus epoch on the
first set of digits \{0,1,2,3,4\} while the solid blue line gives the
accuracies on the second set of digits \{5,6,7,8,9\} versus epoch. Figure
\ref{cnn_cata_revision_1c1p1fc_v2} is a zoomed in picture of Figure
\ref{cnn_cata_revision_1c1p1fc_zoom_v2} for better resolutions of the
accuracies above 90\%. These plots also show the validation accuracy results
when the second set of training data modified to include a fraction of data
from the first set of training digits \{0,1,2,3,4\}. For example, the dashed
red line is the validation accuracy on the first set of digits when the
network was trained with 2000 digits per class of \{5,6,7,8,9\} \emph{along
with} 200 (10\%) digits per class of \{0,1,2,3,4\}. The blue dashed line is
the validation accuracy of the second set of digits after each epoch.
Similarly this was done with 15\%, 25\%, 27.5\%, and 30\% of the first set of
digits included in the training set of the second set of digits. The solid red
line shows that after training with the second set of digits for a single
epoch the validation accuracy on first set goes down to 10\% (random
accuracy). The solid blue line shows a validation accuracy of over 97\% on the
second set of digits after the first epoch. Thus the network has now learned
the second set of digits but has catastrophically forgotten the first set of
digits shown by solid red line.
\bigskip

\subsection{Forgetting In Spiking Networks}

\begin{figure}[H]%
\centering
\includegraphics[
height=2.5226in,
width=4.7249in
]%
{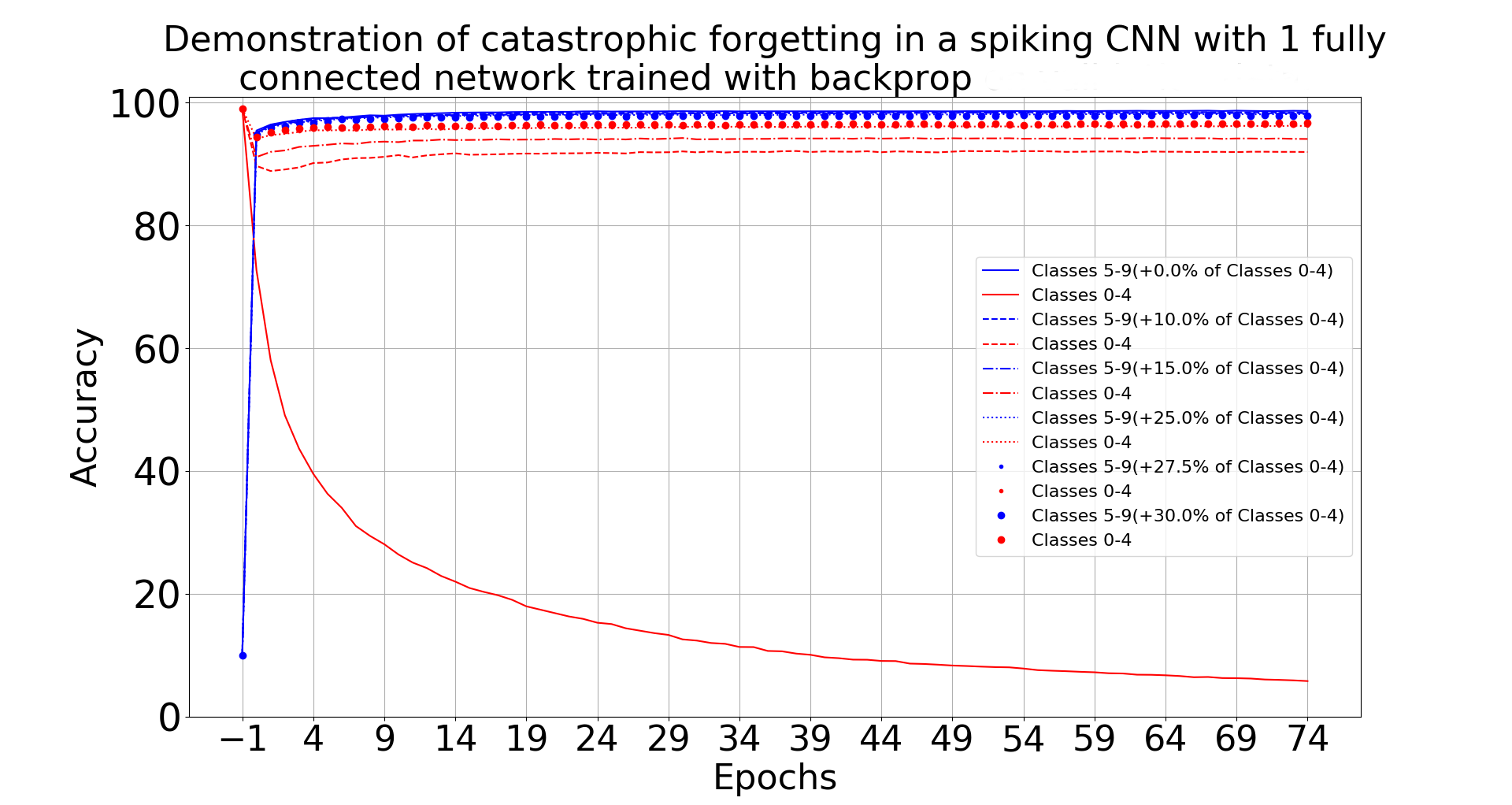}%
\caption{Catastrophic forgetting in a spiking convolutional network while
revising a fraction of the previously trained classes. Note that epoch -1
indicates that the network was tested for validation accuracy before training
of the classes 5-9 started. Brackets in the legend shows the fraction of
previously trained classes that were used to revise the weights from the
previous classes.}%
\label{dscnn_cata_forgetting_revision2_temp}%
\end{figure}

\begin{figure}[H]%
\centering
\includegraphics[
height=2.5607in,
width=4.7988in
]%
{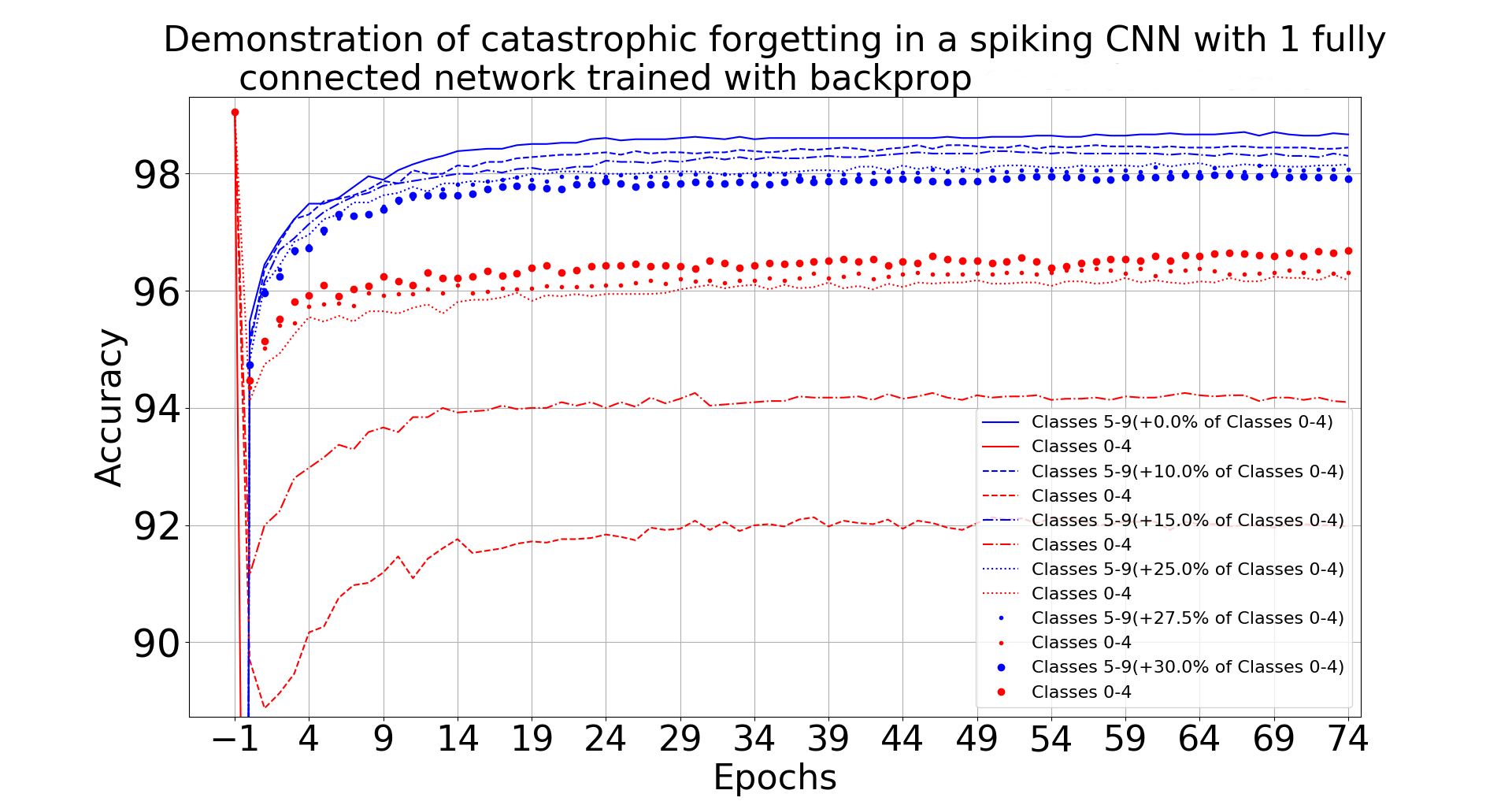}%
\caption{Zoomed upper portion of the Figure
\ref{dscnn_cata_forgetting_revision2_temp}}%
\label{dscnn_cata_forgetting_revision3_temp}%
\end{figure}

For comparison we tested forgetting in our spiking network of Section
\ref{Section3.2} (see Figure \ref{1c1p1fcn_mnistnetwork2}). The network was
first trained only on the digits 0,1,2,3,4 with STDP\ on the convolution layer
and back propagating the error (computed from all 10 outputs) just to the
previous (flattened pool layer) layer. This training used approximately 2000
digits per class and was done for 75 epochs. Then, before training the network
on the set of digits \{5,6,7,8,9\}, we initialized the weights of the
convolution and fully connected layer with the saved weights of the previous
training. For the training with the digits 5,6,7,8,9 we \emph{fixed} the
weights of the convolution layer with their initial values. The network was
then trained, but only the weights of fully connected layer were updated.
(I.e., the error was only back propagated from the 10 output neurons to the
previous layer (flattened pooled neurons). This training also used
approximately 2000 digits per class and was done for 75 epochs. While the
network was being trained on the second set of digits, we computed the
validation accuracy on all 10 digits at the end of each epochs. We plotted
these accuracies in Figure \ref{dscnn_cata_forgetting_revision2_temp}. The
solid red line in Figure \ref{dscnn_cata_forgetting_revision2_temp} are the
accuracies versus epoch on the first set of digits \{0,1,2,3,4\} while the
solid blue line gives the accuracies on the second set of digits \{5,6,7,8,9\}
versus epoch. Figure \ref{dscnn_cata_forgetting_revision3_temp} is a zoomed in
picture of Figure \ref{dscnn_cata_forgetting_revision2_temp} for better
resolutions of the accuracies above 90\%. These plots also show the validation
accuracy results when the second set of training data modified to include a
fraction of data from the first set of training digits \{0,1,2,3,4\}. For
example, the dashed red line is the validation accuracy on the first set of
digits when the network was trained with 2000 digits per class of
\{5,6,7,8,9\} \emph{along with} 200 (10\%) digits per class of \{0,1,2,3,4\}.
The blue dashed line is the validation accuracy of the second set of digits
after each epoch. Similarly this was done with 15\%, 25\%, 27.5\%, and 30\% of
the first set of digits included in the training set of the second set of
digits. The solid red line shows that after training with the second set of
digits for a single epoch the validation accuracy on first set goes down to
77\% (compared to the 10\% accuracy of a non-spiking CNN). The solid blue line
shows a validation accuracy of about 95\% on the second set of digits after
the first epoch. Thus the network has now learned the second set of digits but
has not catastrophically forgotten the first set of digits shown by solid red
line.
%

\begin{figure}[H]%
\centering
\includegraphics[
height=2.6273in,
width=4.9234in
]%
{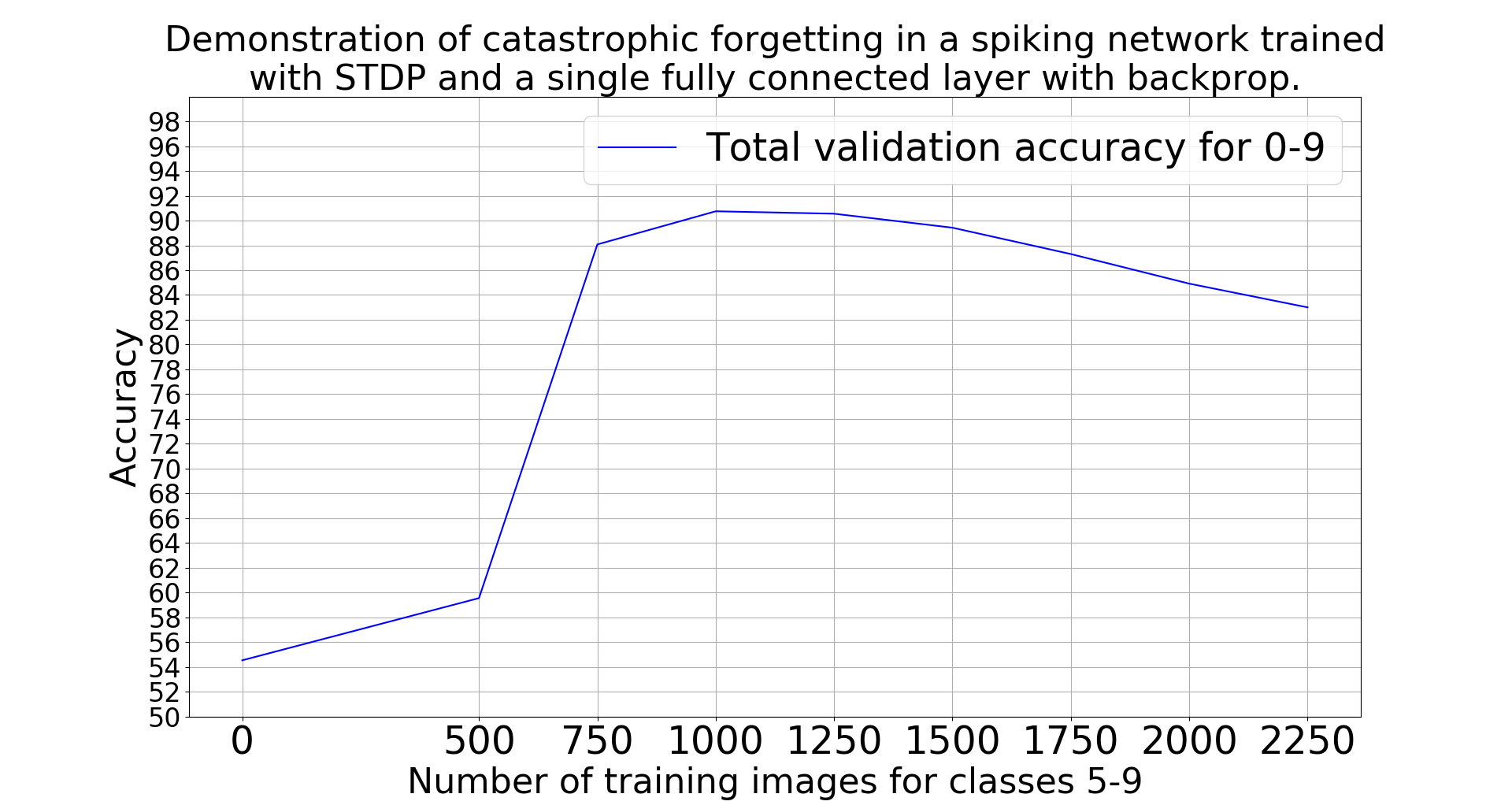}%
\caption{Note that as the number of training images for the classes 5-9
increases the total accuracy drops.}%
\label{dscnn_cata_forgetting_plots2}%
\end{figure}

As another approach we first trained on the set \{0,1,2,3,4\} exactly as just
describe above. However, we then took a different approach to training on the
set \{5,6,7,8,9\}. Specifically we trained on 500 random digits chosen from
\{5,6,7,8,9\} (approximately 50 from each class) and then compute the
validation accuracy on all ten digits. We repeated this for every additional
250 images with the results shown in Figure \ref{dscnn_cata_forgetting_plots2}%
. Interestingly this shows that if we stop after training on 1000 digits from
\{5,6,7,8,9\} we retain a validation accuracy of 91.1\% \ and 90.71\% test
accuracy on all 10 digits.

\begin{table}[th]
\centering
\begin{tabular}
[c]{|c|c|c|c|c|}\hline
\textbf{\# images (classes 5-9)} & \textbf{\# images (classes 0-4)} &
\textbf{Validation} & \textbf{Test} & \textbf{Epochs}\\\hline
10,000 & 1000(10\%) & \ \ \ 95.235\% & 95.1\% & 75\\\hline
10,000 & 1500(15\%) & \ 95.95\% & 95.9\% & 75\\\hline
10,000 & 2500(25\%) & \ 96.83\% & \ 96.81\% & 75\\\hline
10,000 & \ \ \ 2750(27.5\%) & \ 96.98\% & \ \ 96.92\% & 75\\\hline
10,000 & 3000(30\%) & 97.1\% & \ \ \ \ 97.043\% & 75\\\hline
\end{tabular}
\caption{Demonstration of forgetting in a spiking convolution network.}%
\end{table}Jason et al reported an accuracy of 93.88\% for completely disjoint
data sets\cite{allred2019}.

\section{Feature Reconstruction}

We have already presented in Figure \ref{l1l2evolvedfilters} which is a
reconstruction of the convolution kernels (weights) from Layer L1 to Layer 2
into features. Each of the 30 maps of L2 has a convolution kernel in $%
\mathbb{R}
^{2\times5\times5}$ associated with it which maps L1 to L2.

We now want to reconstruct (visualize) the features learned by the second
convolution layer. Each of the 500 maps of L4 (see Figure
\ref{2c2p_mnistnetwork}) has a convolutional kernel in associated with it
which maps L3 to L4, i.e., for $w=0,1,2,...,499$ we have
\[
W_{C2}(w,k,i,j)\in%
\mathbb{R}
^{30\times5\times5}\text{ \ for \ }0\leq k\leq29\text{ \ and }(0,0)\leq
(i,j)\leq(4,4).
\]
\bigskip

A $5\times5$ area of pooled layer L3 receives spikes from $10\times10$ area of
neurons in L2. Thus, for $w=0,1,2,...,499$, the kernels $W_{C2}(w,k,i,j)\in%
\mathbb{R}
^{30\times5\times5}$ are reconstructed to be features
\[
F_{P1}(w,k,i,j)\in%
\mathbb{R}
^{30\times10\times10}\text{\ \ for \ }0\leq k\leq29\text{ \ and }%
(0,0)\leq(i,j)\leq(9,9)
\]
connecting L2 to L4. How is this done? Consider the $0^{th}$ kernel
$W_{C2}(0,k,i,j)$ and for the $k^{th}$ $5\times5$ slice of $W_{C2}(0,k,i,j)\in%
\mathbb{R}
^{30\times5\times5}$ the value of the $(i,j)$ element is mapped to the
$(2i,2j)$ element of the $k^{th}$ $10\times10$ slice of $F_{P1}(0,k,i,j)\in%
\mathbb{R}
^{30\times10\times10}.$ All other values of the $k^{th}$ $10\times10$ slice
are set to zero. This is done for $w=0,1,...,499.$

Now recall that there are 30 kernels in $W_{C1}$. Specifically, for
$z=0,1,2,...,29,$
\[
W_{C1}(z,k,i,j)\in%
\mathbb{R}
^{2\times5\times5}\text{ \ for \ }0\leq k\leq1\text{ \ and }(0,0)\leq
(i,j)\leq(4,4).
\]
$k=0$ is for ON center kernels and $k=1$ is for off center kernels. These
$W_{C1}$ kernels maps spikes from $14\times14$ area of neurons in L1 to a
$10\times10$ area of layer of L2. Thus the feature $F_{P1}(0,k,i,j)\in%
\mathbb{R}
^{30\times10\times10}$ must be reconstructed to be a feature in $%
\mathbb{R}
^{2\times14\times14}$. That is, for $k=0,1$%
\[
F_{L1}(w,k,i,j)\in%
\mathbb{R}
^{2\times14\times14}%
\]
(Each neuron in L4 has a field of view of $2\times14\times14$ neurons in L1).
How is this done?

\begin{figure}[H]%
\centering
\includegraphics[
height=3.25in,
width=5.0in
]%
{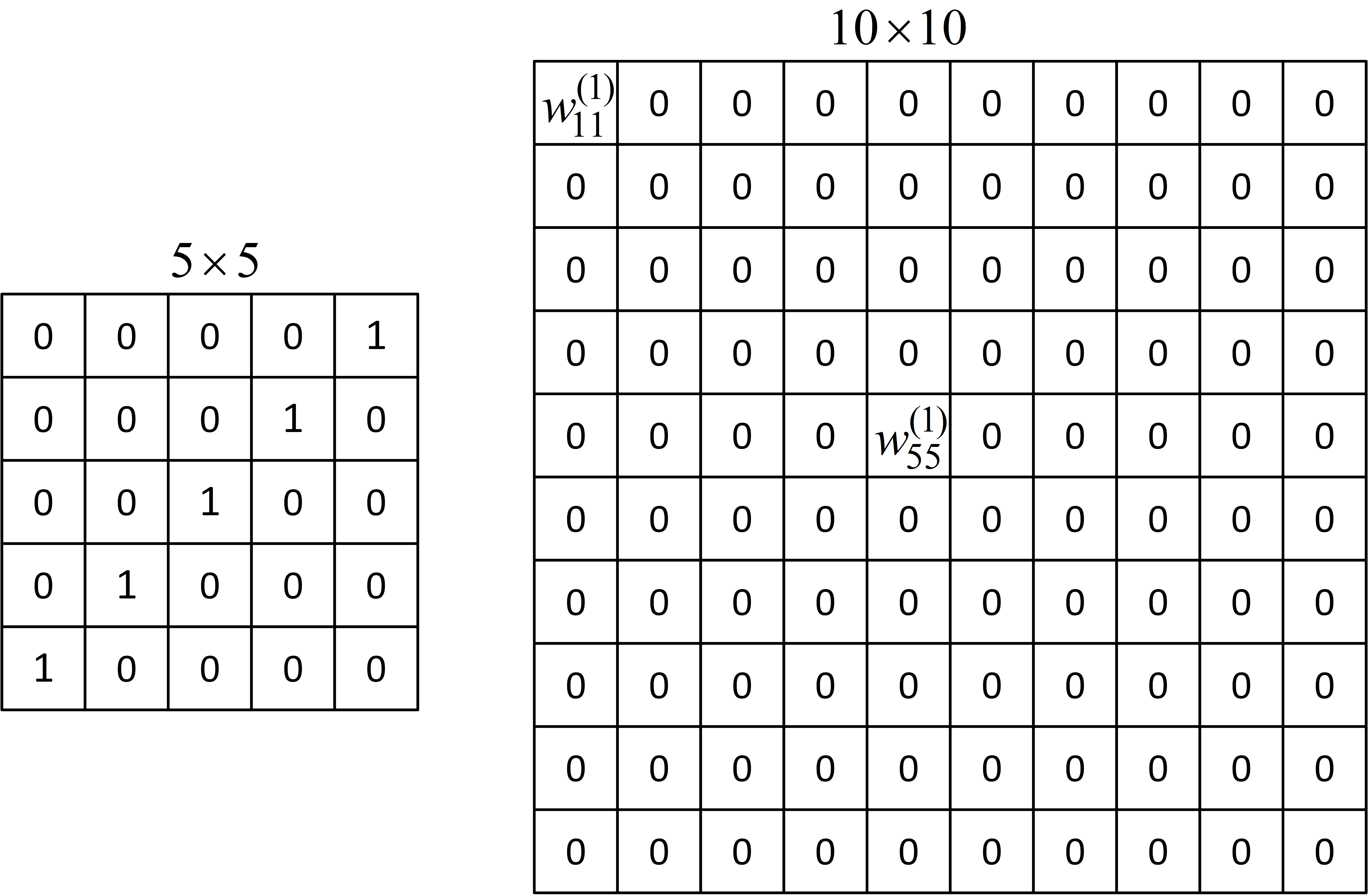}%
\caption{Left: First ON $5\times5$ slice. Right: First $10\times10 $ slice of
pool 1 features.}%
\label{fig_sn12}%
\end{figure}

\begin{figure}[H]%
\centering
\includegraphics[
height=2.3549in,
width=2.3428in
]%
{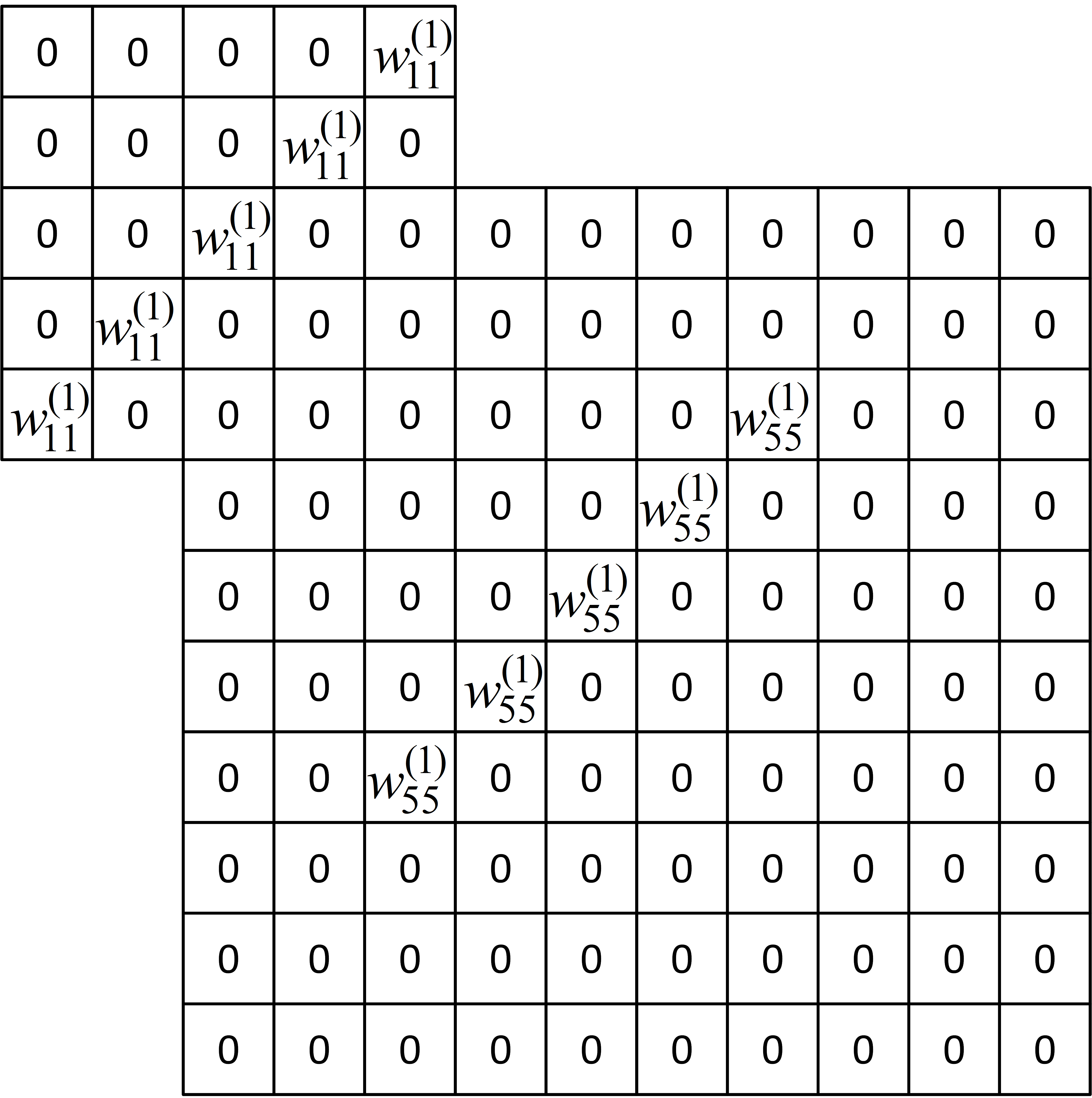}%
\caption{Reconstruction at Conv1 (L2)}%
\label{fig_sn13}%
\end{figure}

\begin{figure}[H]%
\centering
\includegraphics[
height=3.0606in,
width=2.7268in
]%
{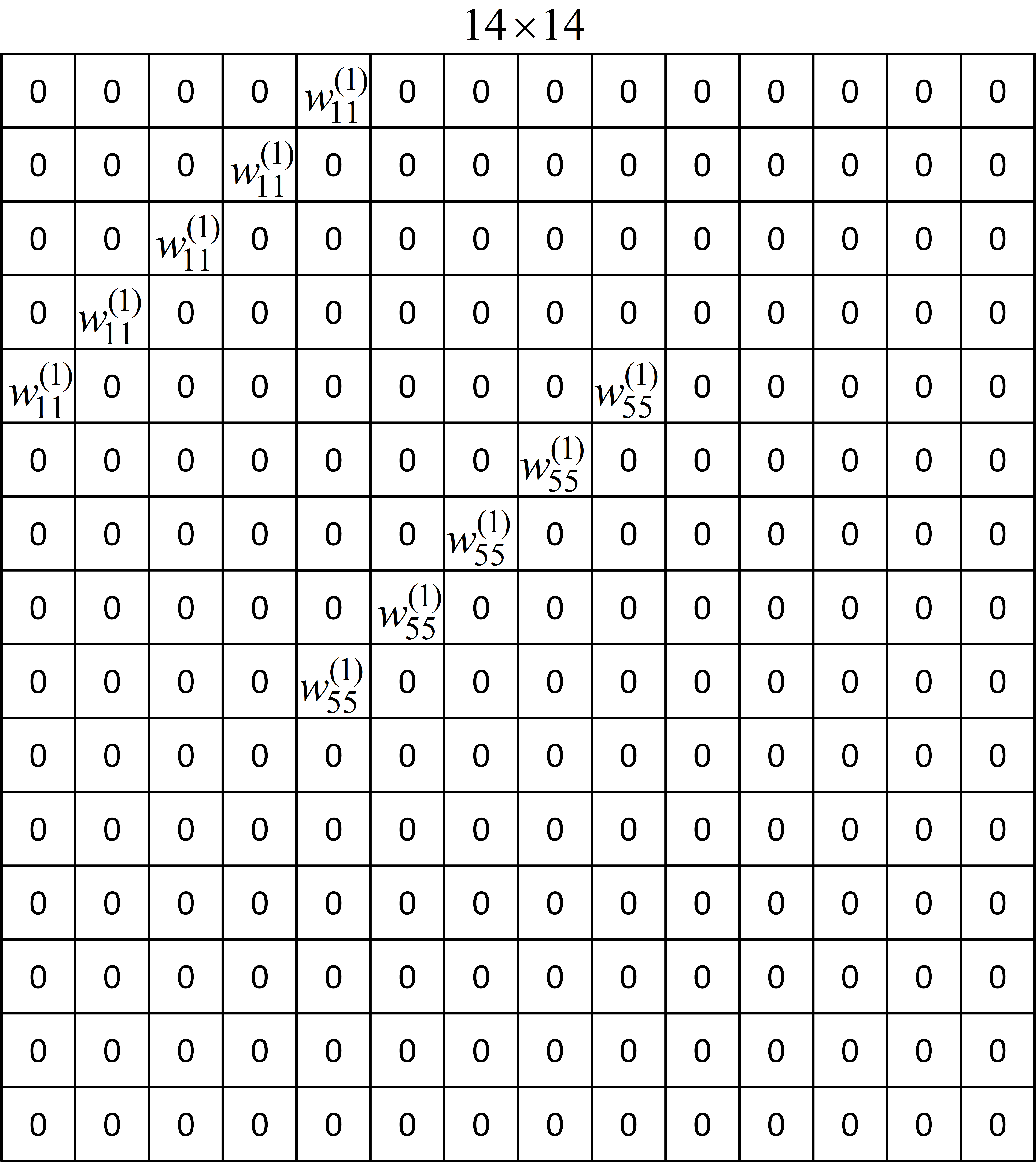}%
\caption{Reconstruction at Conv1 (L2)}%
\label{fig_sn14}%
\end{figure}

\begin{figure}[H]%
\centering
\includegraphics[
height=2.9603in,
width=5.5486in
]%
{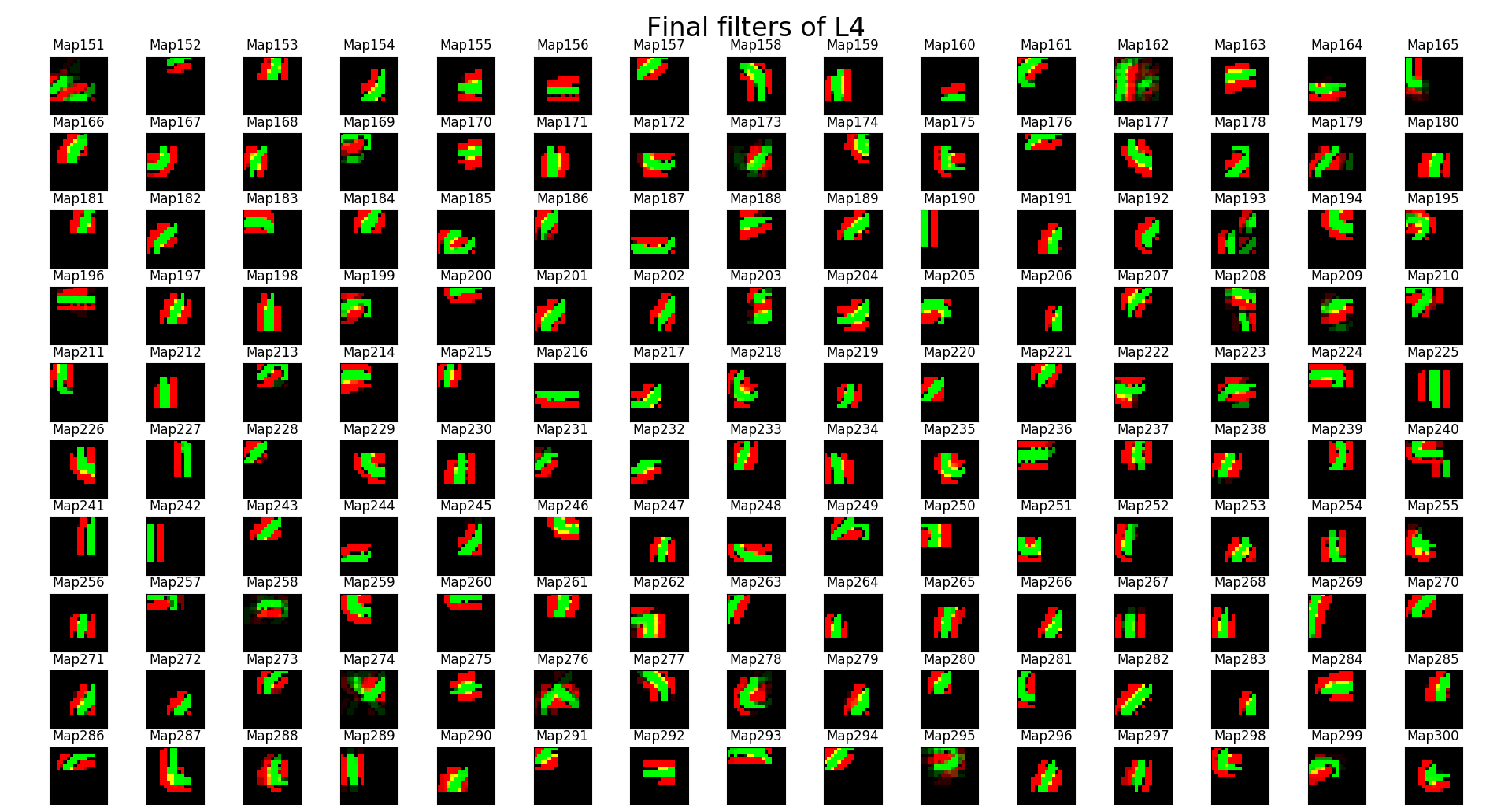}%
\caption{Weights of 150-300 maps of L4 that is trained by in coming spikes
without lateral inhibition in L3, STDP competition region in L4 set to
$\mathbb{R} ^{500\times3\times3}$ and with homeostasis signal applied in L4,
notice that the reconstructed features are quite complex and they could well
represent a digit or a major section of a digit, note that all neurons of a
map in a layer will have shared weights. In this experiment number of maps is
L4 was set to 500. Notice that the reconstructed features are not as complex
looking as in Figure \ref{l3l4finalweightsmnist2inh3trial4ltrl1}}%
\label{l3l4finalweightsmnist2inh3trial3noltrl1}%
\end{figure}

Let the $5\times5$ matrix on the left-hand side of Figure \ref{fig_sn12}
denote an ON center kernel $W_{C1}(z,0,i,j)\in%
\mathbb{R}
^{5\times5}$ for some $z=0,1,...,29.$ In particular, let it be the second
kernel so $z=1.$ Now the $0^{th}$ feature denoted by $F_{P1}(0,k,i,j)\in%
\mathbb{R}
^{30\times10\times10}$ can be visualized as being made up of $10\times10$
slices for $k=0,1,...,29.$ To go with the second kernel $W_{C1}(1,0,i,j)\in%
\mathbb{R}
^{5\times5}$ we take the second slice (k=1) of the feature $F_{P1}(0,k,i,j)\in%
\mathbb{R}
^{30\times10\times10}$ denoted as $F_{P1}(0,1,i,j)\in%
\mathbb{R}
^{10\times10}$ which we take to be the $10\times10$ matrix on the right-hand
side of Figure \ref{fig_sn12}. In practice these slices are sparse and we show
the particular slice Figure \ref{fig_sn12} to have only two non zero elements,
the $(1,1)$ and the $(5,5)$ elements.
To carry out the reconstruction we compute $w_{11}^{(1)}W_{C1}(1,0,i,j)\in%
\mathbb{R}
^{5\times5}$ and center it on $w_{11}^{(1)}$ of $F_{P1}(0,1,i,j)\in%
\mathbb{R}
^{10\times10}$ as indicated in Figure \ref{fig_sn13}. We then repeat this
process for all non zero elements of $F_{P1}(0,1,i,j)\in%
\mathbb{R}
^{10\times10}$ which in this example is just $w_{55}^{(1)}$.
Filling in with zeros we end up with the $14\times14$ matrix shown in Figure
\ref{fig_sn14}.
Each of the 500 reconstructed features of which 150 are shown in Figure
\ref{l3l4finalweightsmnist2inh3trial3noltrl1} is the sum of 30 $14\times14$
matrices of the type shown in Figure \ref{fig_sn14}.

\begin{figure}[H]%
\centering
\includegraphics[
height=3.0in,
width=4.75in
]%
{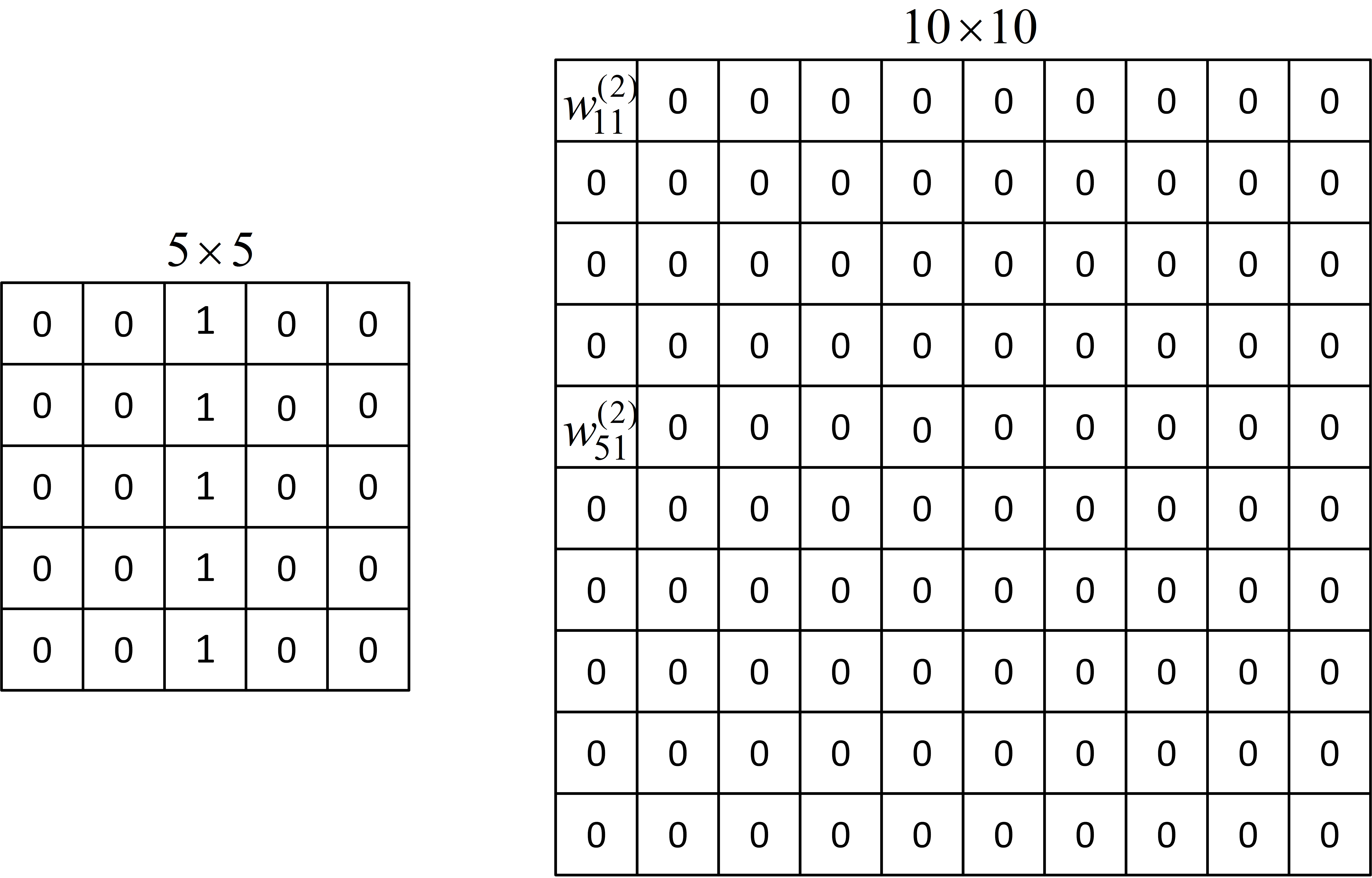}%
\caption{Left: Second ON $5\times5$ slice. Right: Second $10\times10$ slice of
pool 1 features.}%
\label{fig_sn15}%
\end{figure}

To reconstruct the third $14\times14$ matrix we use the third kernel
$W_{C1}(2,0,i,j)\in%
\mathbb{R}
^{5\times5}$ ($z=2$) taken to be the $5\times5$ matrix on the left-side of
Figure \ref{fig_sn15} and the third slice ($k=2$) of the feature
$F_{P1}(0,k,i,j)\in%
\mathbb{R}
^{30\times10\times10}$ denoted as $F_{P1}(0,2,i,j)\in%
\mathbb{R}
^{10\times10}$ which we take to be the $10\times10$ matrix on the right-hand
side of Figure \ref{fig_sn15}.
\textbf{\ }

\begin{figure}[H]%
\centering
\includegraphics[
height=3.0in,
width=4.25in
]%
{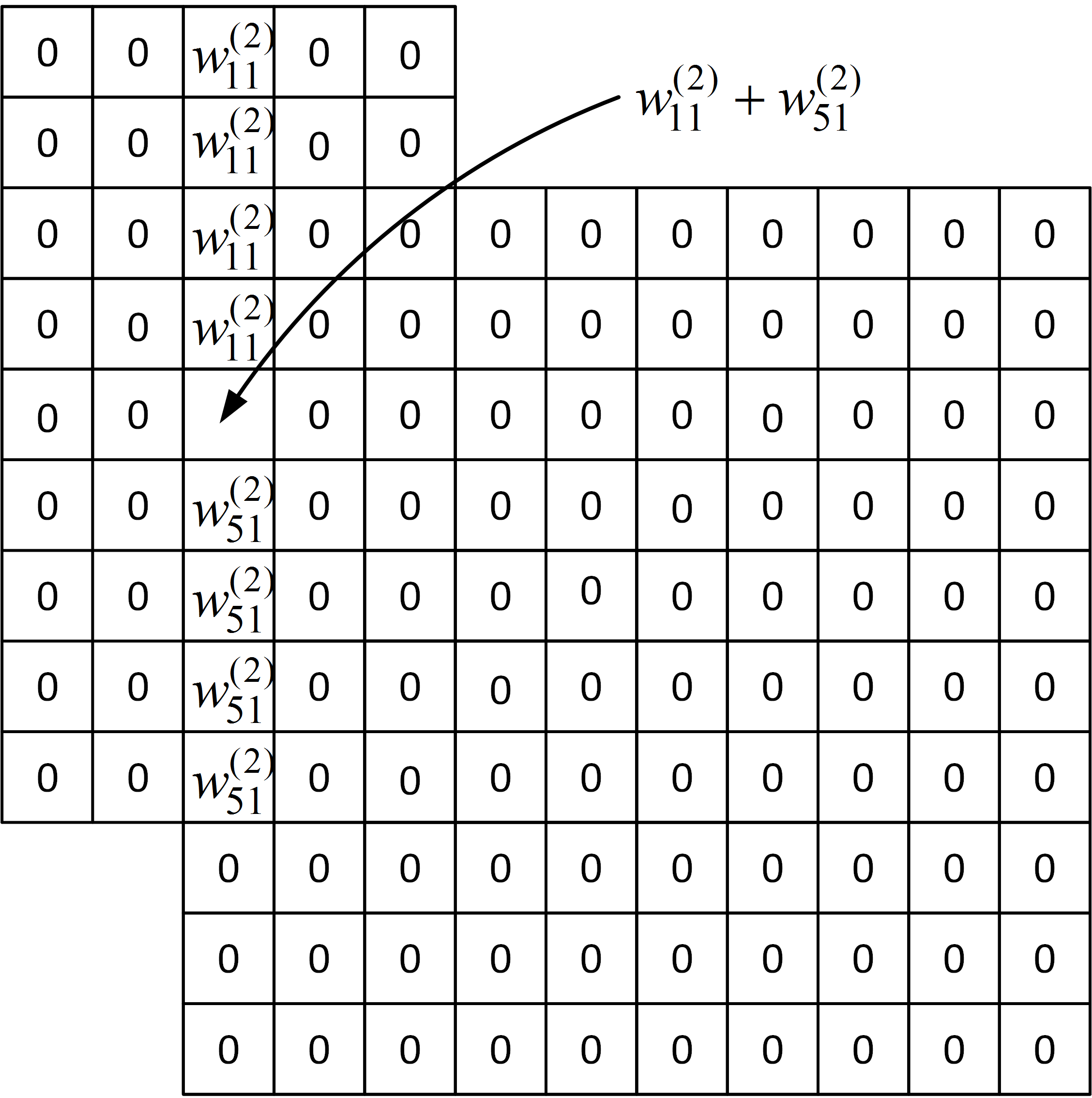}%
\caption{Reconstruction at Conv1 (L2)}%
\label{fig_sn16}%
\end{figure}

\begin{figure}[H]%
\centering
\includegraphics[
height=3.1133in,
width=3.4558in
]%
{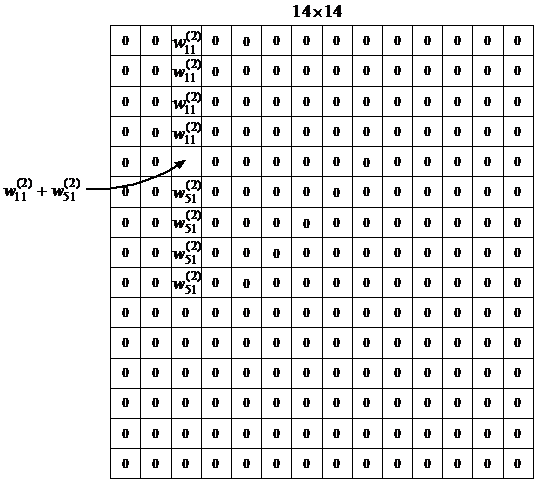}%
\caption{Reconstruction at Conv1 (L2)}%
\label{fig_sn17}%
\end{figure}

Here the only non zero components are $w_{11}^{(2)}$ and $w_{51}^{(2)}$. We
compute $w_{11}^{(2)}W_{C1}(2,0,i,j)\in%
\mathbb{R}
^{5\times5}$ and center it on $w_{11}^{(1)}$ of $F_{P1}(0,2,i,j)\in%
\mathbb{R}
^{10\times10}$ as indicated in Figure \ref{fig_sn16}. We then compute
$w_{51}^{(2)}W_{C1}(2,0,i,j)\in%
\mathbb{R}
^{5\times5}$ and center it on $w_{51}^{(2)}$ of $F_{P1}(0,2,i,j)\in%
\mathbb{R}
.$ In non zero overlapping elements of the $14\times14$ matrix the components
are just added together as shown in Figures \ref{fig_sn16} and \ref{fig_sn17}.%
\textbf{\ }
Finally, 30 of these $14\times14$ matrices are added up to make up one of the
500 features learned by neurons of L4. In other words, a particular neuron of
L4 spikes when it detects its particular\ ($14\times14$) feature in the
original image.

Figure \ref{l3l4finalweightsmnist2inh3trial3noltrl1} shows 150 of the 500
reconstructed features from the 500 convolution kernels of the second
convolution from L3 to L4. Each feature is $14\times14$ neurons (pixels) of
the original spiking image with ON (green) and OFF (red) features superimposed
on top of each other.

\subsubsection{Effect of over training the Convolution Kernels}

\begin{figure}[H]%
\centering
\includegraphics[
height=3.25in,
width=4.4in
]%
{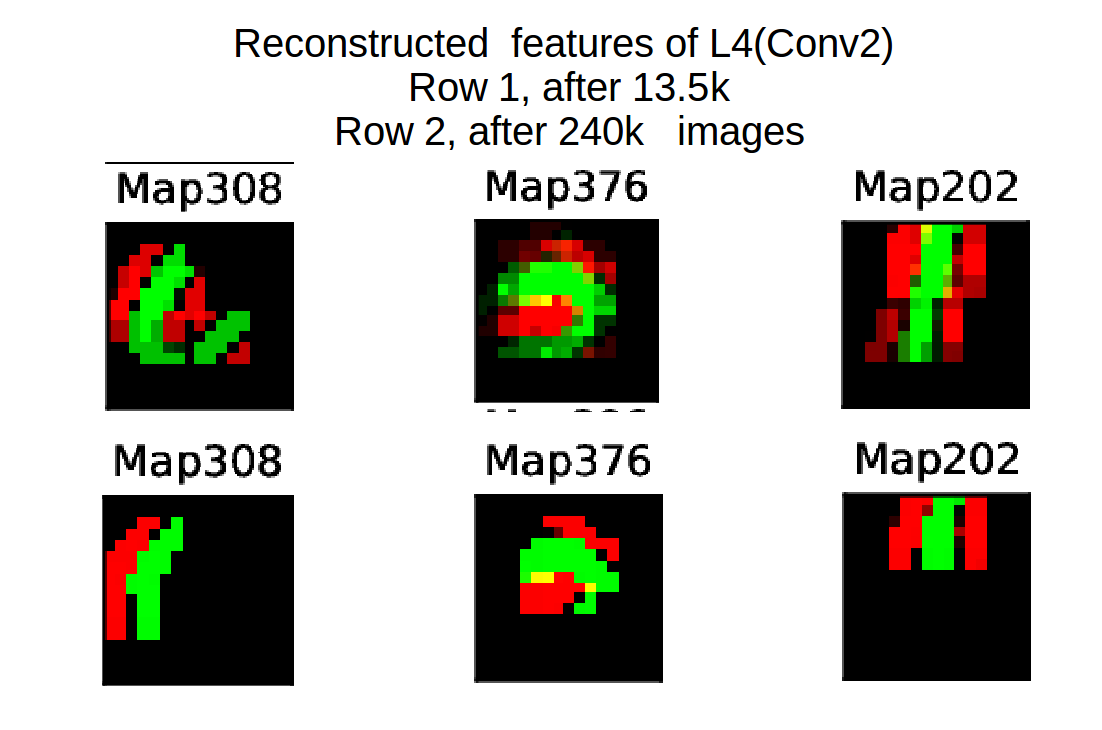}%
\caption{Reduction in the complexity of learnt features because of over
training. First row of this Figure shows reconstruction of L3$\rightarrow$L4
synapses after training for 15.5k images and second row shows the
reconstruction of L3$\rightarrow$L4 synapses after training for 240k images (4
epochs)}%
\label{overfitting}%
\end{figure}

\begin{figure}[H]%
\centering
\includegraphics[
height=2.5in,
width=5.0in
]%
{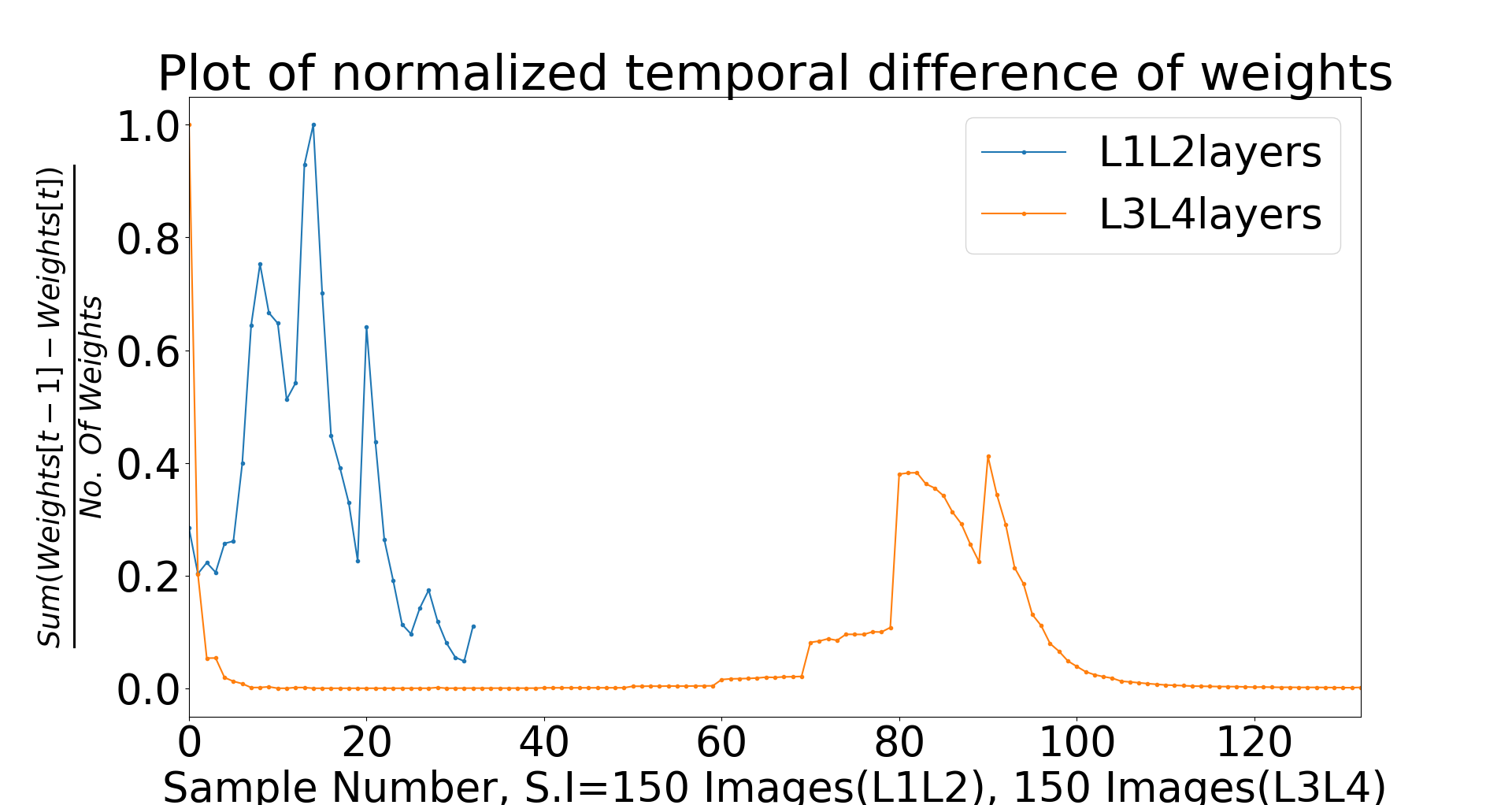}%
\caption{Plot shows the difference of successive samples of synapses,
$\frac{Sum(Weights_{t}-Weights_{t-1})}{no.Of.Synapses}.$ If the difference
approaches zero it means that weights are not changing hence features learnt
by a neuron also remain the same. Notice the sudden jump in difference between
80-100 samples.}%
\label{differenceofweights}%
\end{figure}

The first row of Figure \ref{overfitting} shows the reconstruction of the
features from the convolution kernels of the L3 to L4 layer after training
with just 20,000 images. In contrast, the second row of the Figure
\ref{overfitting} shows the reconstruction of the features from the
convolution kernels of the L3 to L4 layer after training with 60,000 MNIST
images for 4 epochs. This shows that more training results in individual
kernel weights ($w_{ij}$) saturating to 1 or 0 (i.e., the reconstructions in
the second row are sharper), but the features become less complex.
Figure \ref{overfitting} shows that we need a mechanism to stop training. To
this end, we looked at the difference in weights during training.
\[
W_{C2}^{(n)}=\{w^{(n)}(z,i,j,k)\}\in%
\mathbb{R}
^{500\times30\times5\times5}%
\]
where $W_{C2}^{(n)}$ is kernel $W_{C2}$ after the $n^{th}$ training is image
has passed. The L3L4 (red) plot of Figure \ref{differenceofweights} is a plot
of
\[
\frac{\sum_{z=0}^{499}\sum_{i=0}^{29}\sum_{j=0}^{4}\sum_{k=0}^{4}\left(
w^{(n\ast150)}(z,i,j,k)-w^{((n+1)\ast150)}(z,i,j,k)\right)  }{375000}\text{
\ for \ }n=0,1,...,130
\]
where $375000=500\times30\times5\times5$. Similarly the L1L2 (blue) plot was
done for $W_{C1}^{(n)}=\{w^{(n)}(z,i,j,k)\}\in%
\mathbb{R}
^{30\times2\times5\times5}$.

For the L3L4 the weights dramatically change between $n=80$ and $n=100.$
Multiple experiments indicated that over training of $W_{C2}$ kernels starts
after $n=100$. If the network was trained further, we found that the final
classification accuracy drops by by $\sim$2\%.%
Kheradpisheh et al \cite{Kheradpisheh_2016} proposed a convergence factor
given by
\[
\frac{\sum_{z=0}^{499}\sum_{i=0}^{29}\sum_{j=0}^{4}\sum_{k=0}^{4}\left(
w^{(n\ast150)}(z,i,j,k)(1\frac{{}}{{}}-w^{(n\ast150)}(z,i,j,k))\right)
}{375000}\text{ \ for \ }n=0,1,...,130.
\]
The training was stopped when the convergence factor is between 0.01 and 0.02.
We found that using this criteria there was a bit of over training resulting
in 1\%-2\% decrease in testing accuracy.
\begin{figure}[H]%
\centering
\includegraphics[
height=2.5in,
width=5.0in
]%
{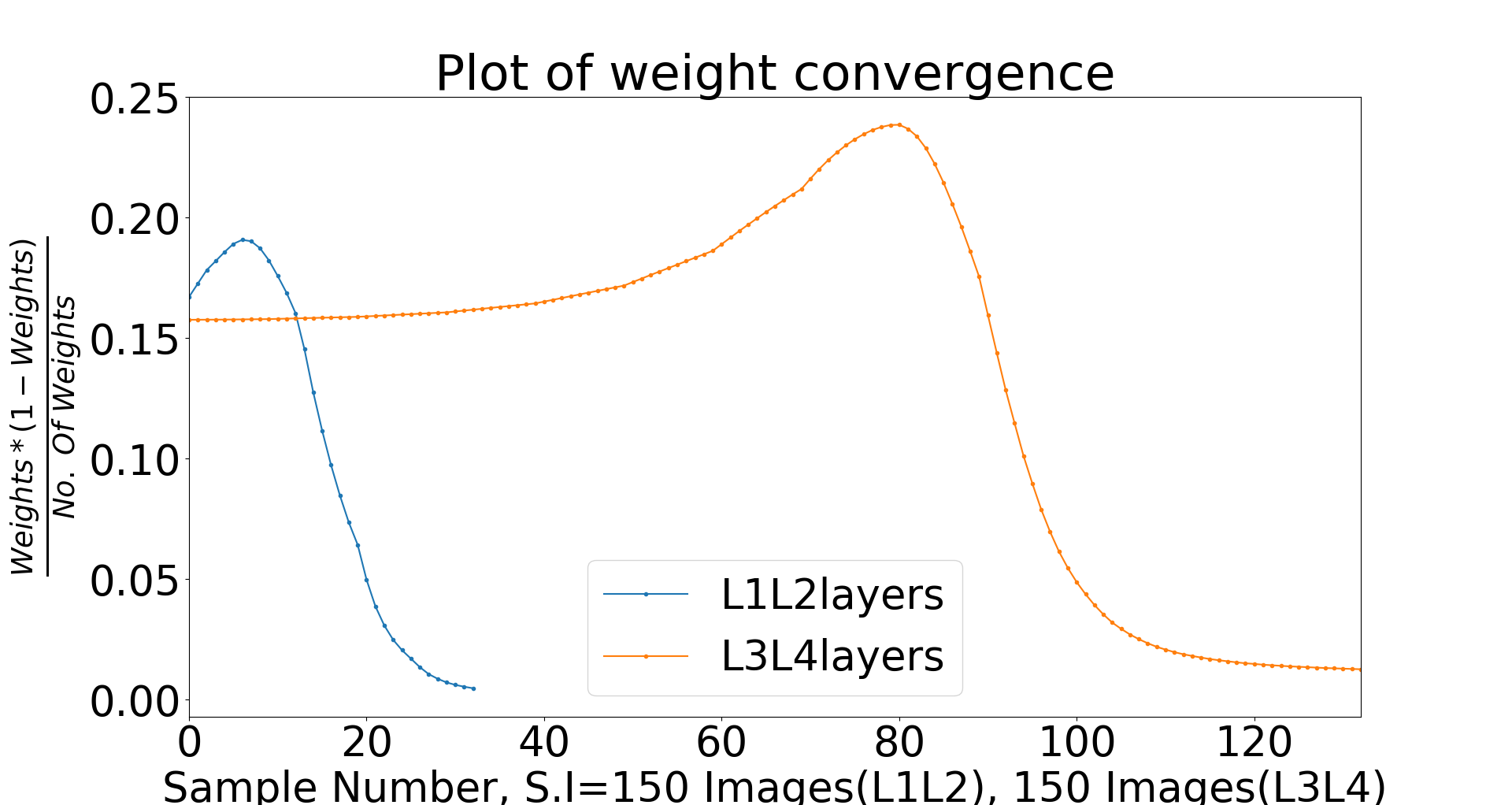}%
\caption{Plot shows the fashion of convergence for the synapses. Note that the
convergence factor dips sharply between the samples 80-100. }%
\label{weightconvergence}%
\end{figure}

\newpage

\section{Conclusion}

We have studied the effects of lateral inhibition and over training in spiking
convolutional networks. We reported above that using a single convolution/pool
layer gives 98.4\% accuracy on the MNIST data set using a two layer backprop
neural network as a classifier. An accuracy of 98.8\% accuracy on the MNIST
data set was obtained when an SVM was used to classify the extracted features.
The same experiments with the same network were carried out on the N-MNIST
data set giving a 97.45\% accuracy with a two layer backprop and a 98.32\%
accuracy using an SVM. We have demonstrated that R-STDP is sensitive to the
weight initialization and a simple two layer error back propagation (avoids
weight transport problem) showed better performance compared to the R-STDP
classifier. We have also shown that catastrophic forgetting is not a severe
problem in spiking convolutional neural networks compared to standard (non
spiking) convolution networks (The spiking network still forgets, but not
catastrophically!). Our spiking CNNs retained a total classification accuracy
of 90.71\% when trained on two disjoint sets and up to 95.1\% when retrained
using 10\% of data from the previously trained data set.

\section{Acknowledgements}

We would like to express our deep gratitude to Professor Timothe\'{e}
Masquelier and Dr. Saeed Reza Kheradpisheh for answering our many questions
about their work \cite{Kheradpisheh_2016}\cite{Kheradpisheh_2016b}. We are
also grateful to Dr. Milad Mozafari for clarifying our questions about his
paper \cite{mozafari1}.

\newpage
\bibliographystyle{abbrv}
\bibliography{Archive_4}
\newpage

\section{APPENDIX}

\subsection{Effect of lateral inhibition in pooling layers on subsequent
convolution layers}

We studied the effects of lateral inhibition \cite{Kheradpisheh_2016}%
\cite{Kheradpisheh_2016b} in convolution and pooling layers in terms
classification accuracy and features learned. Not having lateral inhibition in
pool 1 layer results in better classification provided overtrain in L4 is prevented.

\subsection{With lateral inhibition in pooling layer}

Features learnt in the subsequent layers tend to be more complex looking if
there is lateral inhibition in this layer and less complex looking if lateral
inhibition is not applied. When lateral inhibition is applied, neurons in
pooling layers have no more than one spike per image thereby allowing only the
most dominant neuron at a location $(u,v)$ and across all the maps to spike.
So, out of all the neurons that could have spiked, the synapses of the neuron
that spiked first (dominant) correlate the most with the receptive field.
Hence the features that are learned in the subsequent convolution layers are
more complex looking.%

\begin{figure}[H]%
\centering
\includegraphics[
height=2.8833in,
width=5.4025in
]%
{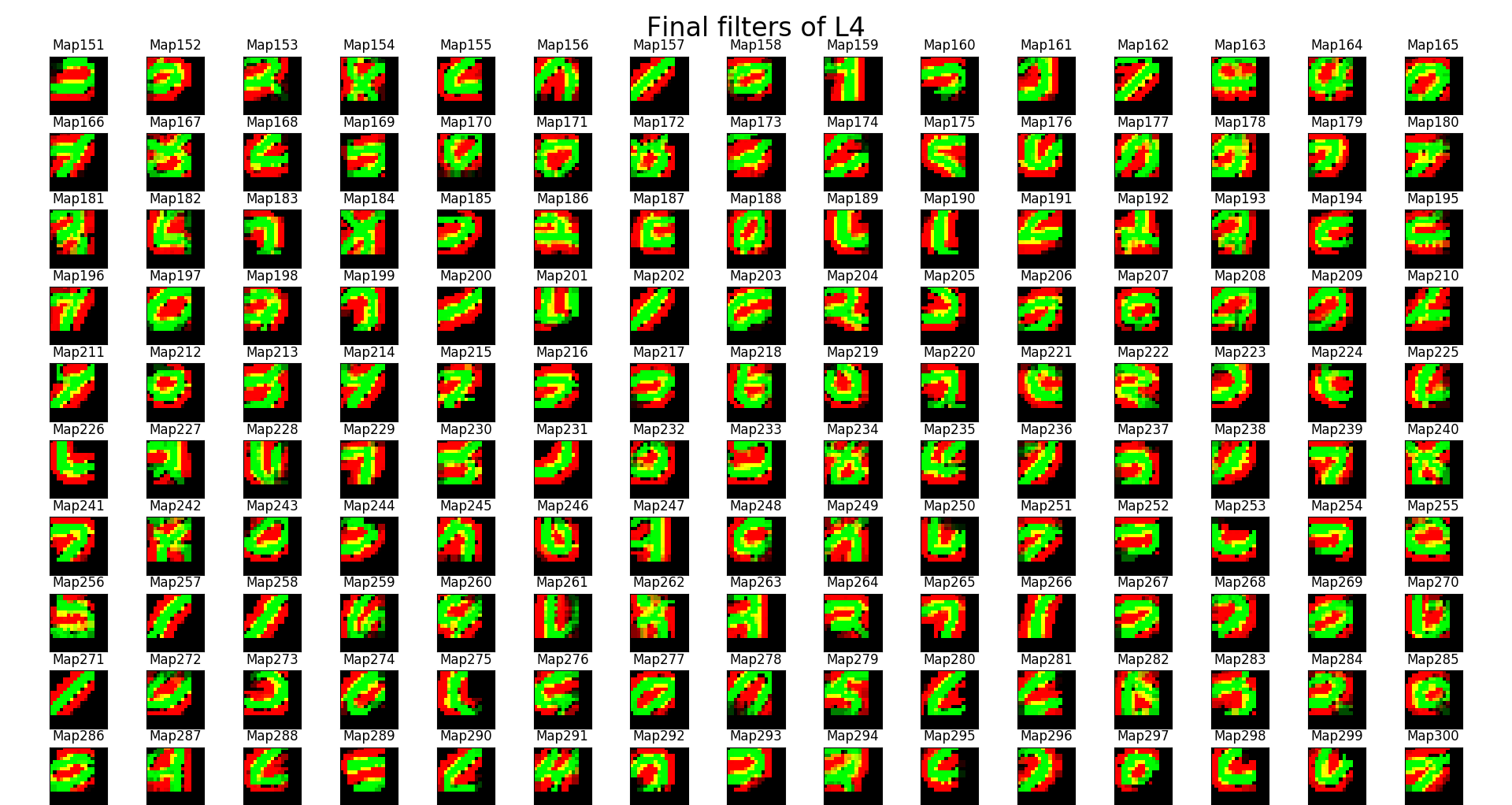}%
\caption{Weights of first 150 maps of L4 that is trained by in coming spikes
with lateral inhibition in L3, STDP competition region in L4 set to
$\mathbb{R} ^{500\times3\times3}$ and with homeostasis signal applied in L4,
notice that the reconstructed features are quite complex and they could well
represent a digit or a major section of a digit, note that all neurons of a
map in a layer will have shared weights. In this experiment number of maps is
L4 was set to 500.}%
\label{l3l4finalweightsmnist2inh3trial4ltrl1}%
\end{figure}

\subsection{Scarcity of the spikes}

With lateral inhibition in pooling layer (L3), number of spikes available at
L4 is reduced drastically. This prevents the build up of the max pooled
potentials of the L4 layer thus it gets harder for a classifier to classify
these vectors.

\end{document}